\documentclass[twoside,11pt]{article}

\usepackage{blindtext}

 \usepackage[abbrvbib,preprint]{jmlr2e}

\usepackage[dvipsnames]{xcolor}
\usepackage{natbib}
\usepackage{amsmath}
\usepackage{array}
\usepackage{multirow}
\usepackage{booktabs} 
\usepackage{bm}
\usepackage{mathtools}
\usepackage{amssymb}

\usepackage{soul}
\usepackage{enumitem}

\usepackage{multirow}
\usepackage{algorithm}
\usepackage[noend]{algpseudocode}
\usepackage{etoolbox}
\algblock{ParFor}{EndParFor}
\algnewcommand\algorithmicparfor{\textbf{for}}
\algnewcommand\algorithmicpardo{}
\algnewcommand\algorithmicendparfor{}
\algrenewtext{ParFor}[1]{\algorithmicparfor\ #1\ }
\algrenewtext{EndParFor}{\algorithmicendparfor}
\newcolumntype{C}[1]{>{\centering\let\newline\\\arraybackslash\hspace{0pt}}m{#1}}

\DeclareMathOperator{\pr}{\mathbb P}
\DeclareMathOperator{\E}{\mathbb E}
\DeclareMathOperator{\Var}{\mathrm{Var}}

\newcommand{\Real}{\mathbb R}

\newcommand{\transpose}{\mathsf T}

\newcommand{\CalF}{\mathcal F}

\newcommand{\CalD}{\mathcal D}
\newcommand{\CalS}{\mathcal S}
\newcommand{\CalH}{\mathcal H}

\newcommand{\CalL}{\mathcal L}
\newcommand{\CalN}{\mathcal N}
\newcommand{\CalW}{\mathcal W}

\newcommand{\CalO}{\mathcal O}
\newcommand{\CalG}{\mathcal G}

\newcommand{\CalT}{\mathcal T}
\newcommand{\CalX}{\mathcal X}

\newcommand{\rmI}{\boldsymbol{\mathrm I}}
\newcommand{\BFx}{\boldsymbol{x}}
\newcommand{\BFX}{\boldsymbol{X}}
\newcommand{\BFy}{\boldsymbol{y}}
\newcommand{\BFP}{\boldsymbol{P}}

\newcommand{\BFc}{\boldsymbol{c}}
\newcommand{\BFD}{\boldsymbol{D}}

\newcommand{\BFY}{\boldsymbol{Y}}
\newcommand{\BFe}{\boldsymbol{e}}
\newcommand{\BFE}{\boldsymbol{E}}
\newcommand{\BFK}{\boldsymbol{K}}
\newcommand{\BFA}{\boldsymbol{A}}
\newcommand{\BFG}{\boldsymbol{G}}
\newcommand{\BFL}{\boldsymbol{L}}
\newcommand{\BFU}{\boldsymbol{U}}
\newcommand{\BFu}{\boldsymbol{u}}
\newcommand{\BFr}{\boldsymbol{r}}
\newcommand{\BFw}{\boldsymbol{w}}
\newcommand{\BFv}{\boldsymbol{v}}
\newcommand{\BFV}{\boldsymbol{V}}

\newcommand{\BFM}{\boldsymbol{M}}

\newcommand{\BFPhi}{\boldsymbol{\Phi}}
\newcommand{\BFalpha}{\boldsymbol{\alpha}}
\newcommand{\BFphi}{\boldsymbol{\phi}}

\newcommand{\BFvarepsilon}{\boldsymbol{\varepsilon}}
\newcommand{\BFSigma}{\boldsymbol{\Sigma}}
\newcommand{\BFdelta}{\boldsymbol{\delta}}
\newcommand{\BFzero}{\boldsymbol{0}}
\newcommand{\BFS}{\boldsymbol{S}}

\newcommand{\argmin}{\operatorname*{argmin}}

\newtheorem{assumption}{Assumption}[section]

\usepackage{lastpage}
\jmlrheading{23}{2024}{1-\pageref{LastPage}}{1/21; Revised 5/22}{9/22}{21-0000}{Lu Zou and Liang Ding}

\ShortHeadings{Kernel Multigrid: Accelerate Back-fitting via Sparse
Gaussian Process Regression}{Zou and Ding}
\firstpageno{1}

\begin{document}

\title{Kernel Multigrid: Accelerate Back-fitting via Sparse
Gaussian Process Regression}

\author{\name Lu Zou  \email lzou@connect.ust.hk \\
       \addr Information Hub, \\
       The Hong Kong University of Science and Technology (Guangzhou)\\
       Guangzhou, Guangdong, China
       \AND
       \name Liang Ding  \thanks{ Corresponding author: Liang Ding}\email liang\_ding@fudan.edu.cn \\
       \addr School of Data Science, \\
       Fudan University\\
       Shanghai, China}

\editor{}

\maketitle

\begin{abstract}
Additive Gaussian Processes (GPs) are popular approaches for nonparametric feature selection. The common training method for these models is Bayesian Back-fitting. However, the convergence rate of Back-fitting in training additive GPs is still an open problem. By utilizing a technique called Kernel Packets (KP), we prove that the convergence rate of Back-fitting is no faster than $(1-\mathcal{O}(\frac{1}{n}))^t$, where $n$ and $t$ denote the data size and the iteration number, respectively. Consequently, Back-fitting requires a minimum of $\mathcal{O}(n\log n)$ iterations to achieve convergence. Based on KPs, we further propose an algorithm called Kernel Multigrid (KMG). This algorithm enhances Back-fitting by incorporating a sparse Gaussian Process Regression (GPR) to process the residuals after each Back-fitting iteration. It is applicable to additive GPs with both structured and scattered data. Theoretically, we prove that KMG reduces the required iterations to $\mathcal{O}(\log n)$ while preserving the time and space complexities at $\mathcal{O}(n\log n)$ and $\mathcal{O}(n)$ per iteration, respectively. Numerically, by employing a sparse GPR  with merely 10 inducing points, KMG can produce accurate approximations  of high-dimensional targets within  5 iterations.
\end{abstract}


\section{Introduction}


Additive Gaussian Processes (GPs) are popular approaches for nonparametric feature selection, effectively addressing high-dimensional generalized additive models \citep{hastie2017generalized} from a Bayesian perspective. This method posits that the high-dimensional observed data $\CalG$ can be decomposed into an additive form of one-dimensional GPs, expressed as $\CalG = \sum_d\CalG_d$. It then leverages Bayes' rule to infer the posterior distribution of the contributions $\CalG_d$ from each dimension individually. Thus, additive GPs offer not only  high interpretability but also uncertainty quantification of their outcomes by providing distributions for the posteriors.

Despite their considerable benefits, additive GPs come with significant drawbacks, in brief their computational complexity. Training an additive GP model necessitates performing matrix multiplications and inversions. For $n$ training points in $D$-dimension,  matrix operations demand $\mathcal{O}((Dn)^3)$ time and $\CalO(Dn^2)$ space. These  complexities can be prohibitive. Given that additive GPs are typically employed in high-dimensional problems with a substantial number of observations, direct computations become impractical. This computational burden severely limits the practicality of additive GPs in addressing many real-world problems.

Considerable research has been dedicated to enhancing the efficiency of computing the posteriors $\CalG_d$ of additive GPs. Among these efforts, Bayesian Back-fitting \citep{breiman1985estimating,hastie2000bayesian,saatcci2012scalable} is one of the most commonly used algorithms for training additive GPs. Back-fitting, which allows a scalable fit over a single dimension of the input, can be used to fit an additive GP
over a $D$-dimensional space with the same overall asymptotic complexity. Specifically, Back-fitting iteratively conducts Gaussian Process Regression (GPR) for each dimension individually while holding the others fixed.  Thanks to efficient algorithms  for training one-dimensional GPR, such as the  state-space model \citep{hartikainen2010kalman} or kernel packets (KP) \citep{chen2022kernel,ding2024general}, each Back-fitting iteration can be executed with efficient time and space complexities of $\CalO(Dn\log n)$ and $\CalO(Dn)$, respectively.

However, the convergence rate of Back-fitting for additive GPs remains an open problem. Despite its remarkable performance in prediction and classification tasks, as shown in \citep{gilboa2013scaling,delbridge2020randomly}, leading to speculation that Back-fitting may exhibit exponentially fast convergence for additive GPs, our analysis reveals a different picture. By employing a technique called Kernel Packet (KP), which provides an explicit formula for the inverse kernel matrix, we can prove that Back-fitting converges no faster than $(1-\mathcal{O}(\frac{1}{n}))^t$ when focusing on convergence towards the posterior of each individual dimension. Here $n$ is the data size and $t$ is the iteration number. This finding implies that a minimum of $\mathcal{O}(n\log n)$ iterations is necessary for Back-fitting to reach convergence. The intuition is that Back-fitting struggles to accurately distribute global features across the correct dimensions, although the composite structure of additive models for prediction or classification might compensate for errors resulting from such misallocations. Inspired by the Algebraic Multigrid method in numerical analysis, we introduce an algorithm named Kernel Multigrid (KMG). This approach refines Back-fitting by integrating a sparse GPR to handle residuals after each iteration. The inclusion of sparse GPR aids in efficiently reconstructing global features during each step of the iterative process, thereby effectively addressing the aforementioned limitation associated with Back-fitting.  KMG is versatile, applicable to additive GPs across both structured and scattered data. If we impose a weak condition on the associated sparse GPR, we can prove that the algorithm significantly reduces the necessary iterations to $\mathcal{O}(\log n)$ while maintaining the time and space complexities unchanged for each iteration. Furthermore, through numerical experiments comparing Back-fitting and KMG, we find that with merely 10 inducing points in the sparse GPR, KMG is capable of accurately approximating high-dimensional targets in 5 iterations.

In summary, we have established a lower bound for the convergence rate of Back-fitting, demonstrating that a minimum of $\mathcal{O}(n\log n)$ iterations is required for convergence. To address this limitation, we introduce the KMG algorithm, a slight modification of Back-fitting that overcomes its shortcomings. Consequently, KMG achieves exponential convergence towards each one-dimensional target posterior $\CalG_d$.

\subsection{Literature Review}
Scalable GPs aim to improve the scalability of full GP models without compromising the quality of predictions. Two prominent approaches in this domain are sparse approximation and iterative methods. Sparse approximation \citep{seeger2003fast, snelson2005sparse,quinonero2005unifying,titsias2009variational,eleftheriadis2023sparse} rely on selecting a set of $m$ inducing points from a large pool of $n$ observation to effectively approximate the posterior at a time complexity of $\mathcal{O}(nm^2)$. The main challenge lies in optimally selecting inducing points to strike a balance between computational efficiency and prediction accuracy.  Recently, several work including \cite{wilson2015kernel,hensman2018variational,dutordoir2020sparse,burt2020variational} have attempted to investigate this issue. Typically, these approaches enable the utilization of a considerable number of inducing points with minimal cost; however, these methods require imposing certain restrictions to the subspaces used for approximation.



Iterative methods, particularly the Conjugate Gradients(CG) method \citep{lanczos1950iteration, hestenes1952methods, saad2003iterative}, provide an alternative strategy for efficient approximation in GPs. Recently, there has been a notable increase in the popularity of iterative methods \citep{cutajar2016preconditioning,cockayne2019bayesian,bartels2020conjugate}, largely due to their precision and  effective utilization of GPU parallelization \citep{gardner2018gpytorch}.  Numerous enhancements  to CG method have been proposed to improve numerical stability and accelerate convergence speed. For example,  \cite{potapczynski2021bias} propose random truncations method to reduce the bias of CG. Similarly, \cite{wenger2021reducing} propose variance reduction techniques to reduce the bias of CG.  Additionally, \cite{maddox2021iterative} provided comprehensive implementation guidance for the CG method, assisting practitioners in consistently attaining robust performance. For literature that delves into theoretical analysis of CG methods in GPs, please refer to \citep{wenger2022preconditioning,wenger2022posterior}. These iterative approaches  have demonstrated effectiveness on large scale datasets  \citep{wang2019exact}.  The performance of the CG is critically dependent on the initial solution and the preconditioner.  An ill-chosen preconditioner can significantly impede convergence \citep{cutajar2016preconditioning,wu2023large}.  

Multigrid methods \citep{saad2003iterative,borzi2009multigrid} are classic computational techniques designed to solve partial differential equations (PDEs), serving as a complement to the CG methods in numerical computation. A key difference between Multigrid methods and the CG methods is that Multigrid methods cycle through various grid resolutions, transferring solutions and residuals between finer and coarser grids, whereas CG methods iteratively refines the solution by moving along conjugate directions. However, the application of Multigrid methods is constrained by their requirement for a grid-based domain and noiseless observations, which limits their suitability for  GPR. Our method generalizes the idea of Multigrid methods for scatter datasets with noise.

Additive GP models \citep{duvenaud2011additive, kaufman2010bayesian,lopez2022high,lu2022additive} have recently attracted significant interest due to their broad applicability in various fields. The  structure of additive GPs allows for implementation of Back-fitting.  A key advantage of Back-fitting is its ability to update the posterior for each dimension at one time, thereby avoiding the costly joint updates \citep{luo2022sparse}. Furthermore,  Back-fitting is also a  popular methodology  for training bayesian tree structure type model \citep{chipman2012bart, linero2018bayesian,lu2023gaussian}, demonstrating its flexibility and effectiveness in various scenarios. Recently, the idea of projection pursuit \cite{friedman1981projection} has been integrated into additive GPs, introducing greater flexibility and enhancing computational efficiency. For further reading on projection pursuit in additive GPs, please see  \cite{gilboa2013scaling, delbridge2020randomly,chen2022projection}.




Back-fitting has also been extensively embraced in the context of additive models within the frequentist setting \citep{luo1998backfitting,sadhanala2019additive,el2019fast}. The consistency of the Back-fitting has been extensively studied in  \cite{buja1989linear, ansley1994convergence,mammen1999existence}.   Recently, \cite{ghosh2022backfitting} developed a Back-fitting algorithm for generalized least squares estimators, demonstrating its operation at a complexity of  $\mathcal{O}(n)$  under certain asymptotic conditions. This methodology has been extended to generalized linear mixed models for logistic regression \citep{ghosh2022scalable}  and to generalized linear mixed models that include random slopes \citep{ghandwani2023scalable}.  The existing literature on nonparametric  additive  models \citep{mammen2006simple, yu2008smooth, fan2005nonparametric, opsomer2000asymptotic} primarily focuses on the asymptotic properties of Back-fitting estimators, rather than its convergence rate.  To the best of our knowledge, research on the convergence rate of Back-fitting for nonparametric additive models is scarce. 

The paper is organized as follows. In Section \ref{secGP} we provide an overview of the background knowledge for GPs, the KP technique, and Bayesian Back-fitting. In Section \ref{sec:lower_bound}, we establish the lower bound of convergence for Bayesian Back-fitting. Section \ref{secKMG} introduces a novel algorithm named KMG, along with a discussion of its convergence properties. In Section \ref{secNumeric}, the performance of KMG is evaluated on both synthetic and real-world datasets. Finally, we conclude our findings and propose directions for future research in Section \ref{sec:Conclude}.

\section{Preliminaries}\label{secGP}
GPs have emerged as a popular Bayesian method for nonparametric regression, offering the flexibility to establish prior distributions over continuous functions. In this section, we will  introduce some foundational concepts in GPs relevant to our study.

\subsection{General Gaussian Processes}

A GP is a distribution on function $\CalG(\cdot)$ over an input space $\CalX$ such
that the joint distribution of $\CalG$ on any size-$n$ set of  input points $\BFX=\{\BFx_i\}_{i=1}^n\subset \CalX$  is described by a multivariate Gaussian density over the associated targets, i.e.,
\[\pr\left(\CalG(\BFx_1),\cdots,\CalG(\BFx_n)\right)=\CalN(m(\BFX),k(\BFX,\BFX))\]
where $m(\BFX)\in\Real^n$ is an $n$-vector whose $i$-entry equals the value of a mean function $m$ on point $\BFx_i$ and $k(\BFX,\BFX)=[k(\BFx_i,\BFx_j)]_{i,j=1}^n\in\Real^{n\times n}$ is 
an $n$-by-$n$ kernel covariance matrix whose $(i,j)$-entry equals the value of a positive definite kernel $k$ on $\BFx_i$ and $\BFx_j$ \citep{wendland2004scattered}. Accordingly, a GP can be characterized by the mean function $m:\CalX\to \Real$ and the kernel function $k :\CalX\times\CalX\to \Real$. Generally , the mean function is set as $0$ when we have no prior knowledge of the true function. Therefore, we can use only the kernel $k$ to determine a GP.

In this study, we will assume that the underlying GP is \textit{stationary}, i.e., its covariance only depends on the difference: $k(x',x)=k(x'-x)$. GPs with stationary kernels have been widely used in time series (see for example \cite{brockwell1991time}) and spatial statistics (see for example \cite{cressie2015statistics}). For a one-dimensional stationary $k$, we can define its Fourier transform as follows:
\[\CalF[k](\omega)=\frac{1}{\sqrt{2\pi}}\int_\Real k(x)e^{-i \omega x}dx.\]
Now we can impose the following specific assumption on the kernel function. This assumption is weak and includes a large number of kernel classes, such as the Mat\'ern kernels and the compactly supported Wendland kernels \cite[Chapter 9]{wendland2004scattered}.
\begin{assumption}
\label{assump:kernel}
    The kernel $k$ is stationary and its Fourier transform $\CalF[k(\cdot)](\omega)$ satisfies:
    \[C_1(1+\omega^2)^{-s}\leq \CalF[k(\cdot)](\omega)\leq C_2(1+\omega^2)^{-s},\quad \omega\in\Real\]
    for some $s>1/2$ where $C_1$ and $C_2$ are positive constants independent of s, $\omega$ and kernel function k. 
\end{assumption}
We consider the case that observation is noisy $\BFy_i=\CalG(\BFx_i)+\varepsilon_i$, where $\varepsilon_i\sim\CalN(0,\sigma_y^2)$ is an i.i.d. Gaussian distributed error. Then, we can use standard identities of the multivariate Gaussian
distribution to show that, conditioned on data $(\BFX,\BFY)=\{(\BFx_i,\BFy_i)\}_{i=1}^n$, the posterior distribution at any point $\BFx^*$ also follows a Gaussian distribution: $\CalG(\BFx^*) \vert  \{\BFX,\BFY\}\sim\CalN(\mu_n(\BFx^*),{s}_n(\BFx^*))$, where
\begin{align}
&\mu_n(\BFx^*)=k(\BFx^*,\BFX)\left[k(\BFX,\BFX)+\sigma_y^2\rmI_n\right]^{-1}\BFY\nonumber\\
    &s_n(\BFx^*)=k(\BFx^*,\BFx^*)-k(\BFx^*,\BFX)\left[k(\BFX,\BFX)+\sigma_y^2\rmI_n\right]^{-1}k(\BFX,\BFx^*)\label{eq:GP_posterior}
\end{align}
where $k(\BFx^*,\BFX)=(k(\BFx^*,\BFx_1),\cdots,k(\BFx^*,\BFx_n)^{T}$, and $ k(\BFX,\BFX)=[k(\BFx_k, \BFx_l)]^n
_{k,l=1}$.

The computation of \eqref{eq:GP_posterior}  all require $\CalO(n^3)$ time complexcity and $\CalO(n^2)$ space complexity due to the need for inverting dense matrices $k(\BFX,\BFX)+\sigma_y^2\rmI_n$. When data size $n$ is large, GPs are computational inefficient and unstable because the kernel covariance matrix is dense and nearly singular.

\subsection{Additive GPs}
A $D$-dimensional additive GP can be view as summation of $D$ one-dimensional GPs. Specifically,
\[\CalG(\BFx_i)=\sum_{d=1}^D\CalG_d(x_{i,d}),\]
where $\CalG_d$ is a one-dimensional zero-mean GP characterized by kernel $k_d$ and $x_{i,d}$ is the $d$-th entry of the $D$-dimensional training point $\BFx_i$. So an additive GP has kernel function $k=\sum_{d=1}^Dk_d$. We use the following model to describe the generation of observations $\BFY$:
\begin{equation}
    \label{eq:additive_GP}
    \BFy_i=\sum_{d=1}^D\CalG_d(x_{i,d})+\varepsilon_i.
\end{equation}
An additive prior occupies a middle ground between a linear model, which is additive, and a nonlinear regression prior, allowing for the modeling of arbitrary interactions between inputs.  While it does not model interactions between input dimensions, an additive model still provides interpretable results. For instance, one can easily plot the posterior mean of the individual $\CalG_d$ to see how each predictor is related to the target.

The objective of additive GP is to estimate the posterior distribution given the observation data \((\BFX, \BFY)\). Let $\BFX_d=\{x_{i,d}\}_{i=1}^n$ denote the $d$-th dimension of all data points $\{\BFx_i\}_{i=1}^n$. By applying Bayes' rule, we can first derive the distribution of \(\mathcal{G}_d(\BFX_d) | \BFX, \BFY\), which is also Gaussian:
\begin{align}
  \pr\left(\{\mathcal{G}_d(\BFX_d)\}_{d=1}^D | \BFX, \BFY\right) 
    \propto &\pr\left(\BFY|\BFX,\{\mathcal{G}_d(\BFX_d)\}_{d=1}^D\right)\pr\left(\{\mathcal{G}_d(\BFX_d)\}_{d=1}^D|\BFX\right)\nonumber\\
     =&\CalN\left(\sum_{d=1}^D\CalG_d(\BFX_d),\sigma_y^2\rmI_n\right)\CalN\left(\boldsymbol{0},\BFK\right),\label{eq:bayes_rule_1}
\end{align}
where $\BFK=\text{diag}[\BFK_1,\BFK_2,\cdots,\BFK_D]\in\Real^{Dn\times Dn}$ is a block diagonal matrix with  one-dimensional kernel covariance matrices $\BFK_d=k_d(\BFX_d,\BFX_d)$ on its diagonal. A direct result of \eqref{eq:bayes_rule_1} is that the posterior \(\mathcal{G}_d(\BFX_d) | \BFX, \BFY\) is Gaussian with mean and variance as follows:
\begin{align}
    &\E\left[[\mathcal{G}_1(\BFX_1);\cdots;\mathcal{G}_D(\BFX_D)] | \BFX, \BFY\right]= \left[\BFK^{-1}+\sigma_y^{-2}\BFS\BFS^\transpose\right]^{-1}\sigma_y^{-2}\BFS\BFY,\label{eq:posterior_block_mean}\\
    &\Var \left[[\mathcal{G}_1(\BFX_1);\cdots;\mathcal{G}_D(\BFX_D)] | \BFX, \BFY\right]= \left[\BFK^{-1}+\sigma_y^{-2}\BFS\BFS^\transpose\right]^{-1}\label{eq:posterior_block_var},
\end{align}
where  $\BFS=[\rmI_n;\rmI_n;\cdots;\rmI_n]\in\Real^{Dn\times n}$.  

Based on \eqref{eq:posterior_block_mean} and \eqref{eq:posterior_block_var}, we then can compute the posterior of each \(\mathcal{G}_d\) at a one-dimensional input point \(x^*_d\), we can use the following the marginal distribution:
\begin{align}
    \pr\left(\mathcal{G}_d(x^*_d) | \BFX, \BFY\right)=\int \pr\left(\mathcal{G}_d(x^*_d) | \mathcal{G}_d(\BFX_d)\right) \pr\left(\mathcal{G}_d(\BFX_d) | \BFX,\BFY\right) d \mathcal{G}_d(\BFX_d).
\end{align}
 From direct calculations, $\mathcal{G}_d(x^*_d) | \BFX, \BFY$ is also Gaussian with mean and variance as follows:
 \begin{align}
     \E\left[\mathcal{G}_d(x^*_d) | \BFX, \BFY\right]=&k_d(x^*_d,\BFX_d)\BFK_d^{-1}\E\left[\mathcal{G}_d(\BFX_d)| \BFX, \BFY\right]\nonumber\\
     =&k_d(x^*_d,\BFX_d)\BFK_d^{-1}\left[\BFe_d^\transpose \left[\sigma_y^{2}\BFK^{-1}+\BFS\BFS^\transpose\right]^{-1}\BFS\BFY \right] \label{eq:posterior_block_mean_single}\\
     \Var\left[\mathcal{G}_d(x^*_d) | \BFX, \BFY\right]&=k_d(x_d^*,x_d^*)-k_d(x_d^*,\BFX_d)\BFK_d^{-1}k_d(\BFX_d,x^*_d)\nonumber\\
     &\quad +k_d(x^*_d,\BFX_d)\BFK_d^{-1}\left[\BFe_d^\transpose \left[\BFK^{-1}+\sigma_y^{-2}\BFS\BFS^\transpose\right]^{-1}\BFe_d\right]\BFK_d^{-1}k_d(\BFX_d,x^*_d)\label{eq:posterior_block_var_single}
 \end{align}
 where 
 \[\BFe_d^\transpose=[\underbrace{\boldsymbol{0}, \boldsymbol{0},\cdots,\boldsymbol{0},\boldsymbol{0}}_{\substack{(d-1)\ \boldsymbol{0}'{\rm s} \\ \text{where $\boldsymbol{0}\in\Real^{n\times n}$}}},\rmI_n,\boldsymbol{0},\cdots,\boldsymbol{0}]\in\Real^{n\times Dn}\]
 is a block matrix for querying the $d$-th block of a $Dn\times n$ block matrix. For notation simplicity, we write $\lambda:=\sigma_y^{-2}$  in subsequent discussions.

To calculate \eqref{eq:posterior_block_mean_single} and \eqref{eq:posterior_block_var_single}, we will demonstrate in next subsection how the KP technique facilitates matrix operations of the form $k_d(x^*_d,\BFX_d)\BFK^{-1}_d\BFv$ to be executed in $\CalO(n\log n)$ time and $\CalO(n)$ space for any vector $\BFv$. However, it's important to note that we must also perform matrix operations of the form $[\BFK^{-1}+\sigma_y\BFS\BFS^\transpose]^{-1}\BFv$ for certain vectors $\BFv$, which demand significantly higher time and space complexities of $\CalO((Dn)^3)$ and $\CalO(Dn^2)$, respectively. This presents considerable challenges in managing large datasets and addressing high-dimensional problems with additive GPs.


 \subsection{Kernel Packet}\label{sec:KP}
KP yields a sparse formulation for one-dimensional kernel function. Here, we will use the Mat\'ern kernel with half-integer smoothness parameter as an example and kernels in the Mat\'ern class are also our tool to prove the lower bound.  Notably, the half-integer Matérn kernel can be expressed in a closed form as the product of an exponential function and a polynomial of order $s=\nu+1/2$:
\begin{equation}
\label{eq:Matern_half-_int}
    k_{\nu}(x,x')= \sigma^2\exp({-w{\vert x-x'\vert}})\frac{s!}{2s!}\left(\sum_{l=0}^{s}\frac{(s+l)!}{l!(s-l)!}({2w}{\vert x-x'\vert})^{s-l}\right),
\end{equation}
where $w>0$ is called the scaled parameter. Any Mat\'ern-$\nu$ kernel satisfies Assumption \ref{assump:kernel} with $s=\nu+1/2$ for $\nu=1/2,3/2,\cdots$, thereby indicating smoothness of $s$-th order. For general kernel,  \cite{ding2024general} has shown that KP factorization exists as long as the kernel satisfied some specific differential equation, which is a weak condition. 

\begin{figure}[ht]
    \centering
    \includegraphics[width=.45\linewidth]{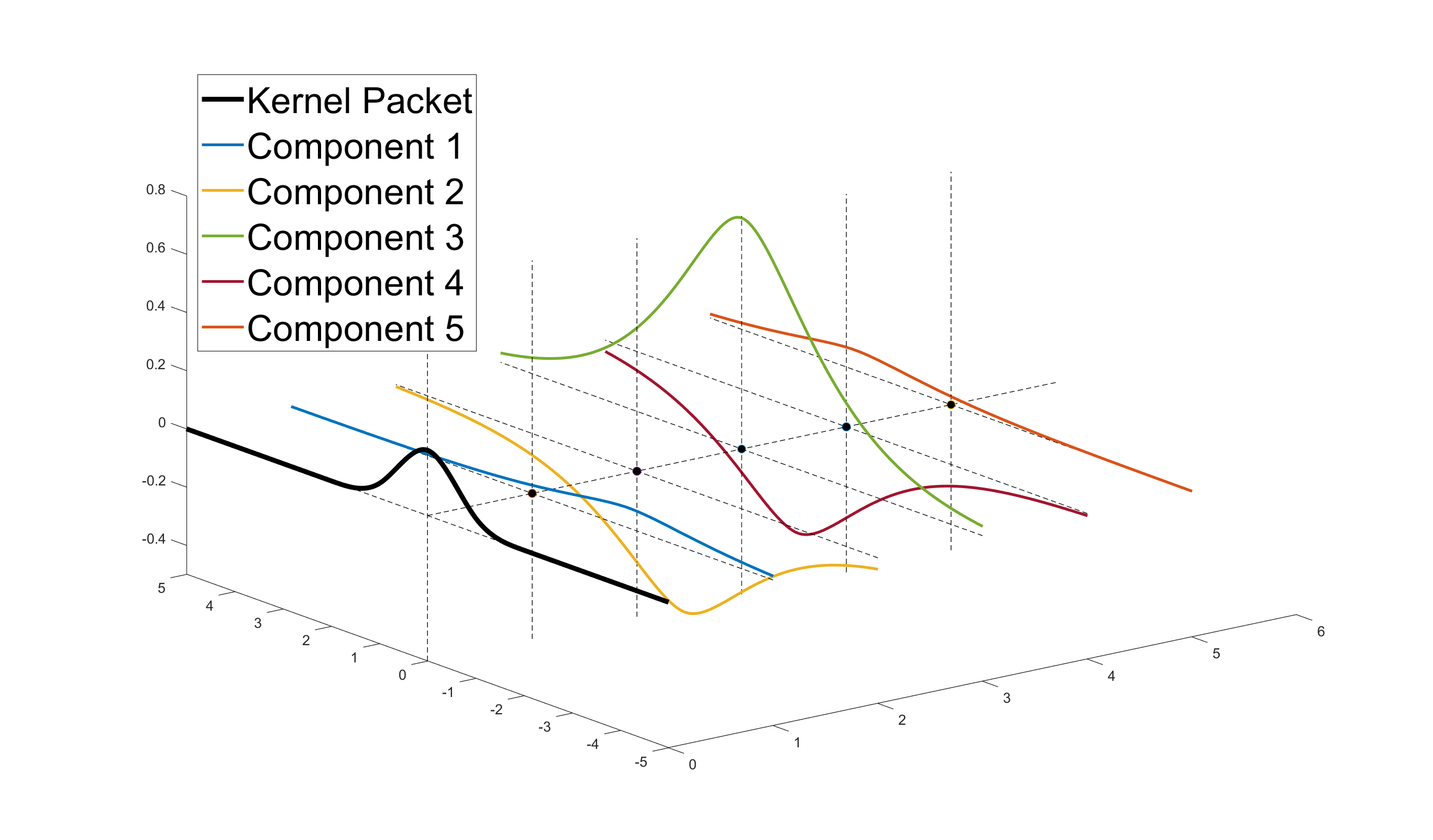}
    \includegraphics[width=.35\linewidth]{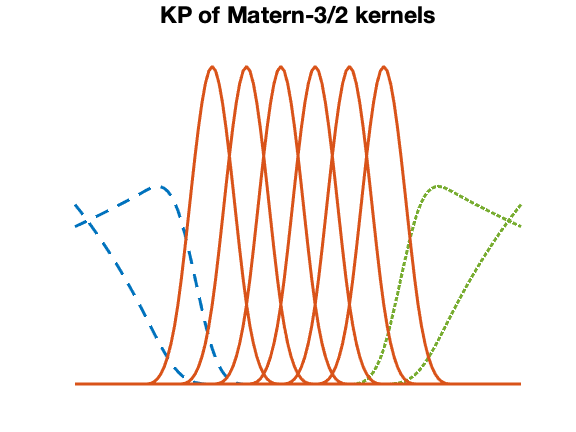}
    \caption{Left: the addition of five Mat\'ern-${3}/{2}$ kernels $a_j k(\cdot,x_j)$ (colored lines, without compact supports) leads to 
    a KP (black line, with a compact support); Right: converting 10 Mat\'ern-${3}/{2}$ kernel functions $\{k(\cdot,x_i)\}_{i=1}^{10}$ to 10 KPs, where each KP is non-zeron on at most three points in $\{x_i\}_{i=1}^{10}$.}
    \label{fig:KP}
\end{figure}

The essential idea of KP fatorization is that for any half-integer Mat\'ern-$\nu$ kernel $k$  and any $2\nu+2=2s+1$ consecutive points $x_1<x_2<\cdots<x_{2\nu+2}$ , let  $a_1,\cdots,a_{2\nu+2}$ be the solution of the following $(2s+1)\times (2s)$ KP system with:
\begin{equation}
    \label{eq:KP_system}
    \sum_{i=1}^{2\nu+2}a_i[e^{-x_i},\ e^{x_i},\ x_ie^{-x_i},\ x_ie^{x_i},\ \cdots, \ x_i^{s-1}e^{-x_i},\ x_i^{s-1}e^{x_i}\ ]^\transpose =\boldsymbol{0},
\end{equation}
then  the following function
\[\phi_{(x_1,\cdots,x_{2\nu+2})}(\cdot)=\sum_{i=1}^{2\nu+2}a_ik(\cdot,x_i)\]
is non-zero only on the open interval $(x_1,x_{2\nu+2})$ and we call function $\phi_{(x_1,\cdots,x_{2\nu+2})}$ KP. Similarly, for any $\nu+3/2=s+1$ consecutive points $x_1<\cdots<x_{\nu+3/2}$, let  $a_1,\cdots,a_{\nu+3/2}$ be the solution of the following $(s+1)\times s$ KP system:
\begin{equation}
    \label{eq:KP_one_sided_system}
    \sum_{i=1}^{\nu+3/2}a_i[e^{\pm x_i},\ x_ie^{\pm x_i},\  \cdots, \ x_i^{s-1}e^{\pm x_i}\  ]^\transpose =\boldsymbol{0},
\end{equation}
then the following function
\[\phi_{(x_1,\cdots,x_{\nu+3/2})}(\cdot)=\sum_{i=1}^{\nu+3/2}a_ik(\cdot,x_i)\]
is non-zero only on the half interval $(-\infty, x_{\nu+3/2})$ (correspond to “$-$” sign in \eqref{eq:KP_one_sided_system}) or $(x_{1},\infty)$ (correspond to “$+$” sign in \eqref{eq:KP_one_sided_system}).  We call these function $\phi_{(x_1,\cdots,x_{2\nu+2})}$ one-sided KP. It is obvious that the solutions in \eqref{eq:KP_system} and \eqref{eq:KP_one_sided_system} can be solved in $\CalO(\nu^3)$ time and they are both unique up to  multiplicative constants.

\begin{algorithm}
{
\caption{Computing KP factorization $\BFP^T\BFK\BFP=\BFA^{-1}[\phi_i(x_j)]_{i,j}$}\label{alg:banded_factorization}
\hspace*{\algorithmicindent} \textbf{Input}: one-dimension Mat\'ern-$\nu$ kernel $k$,  scattered points $\{x_{i}\}_{i=1}^n$ \\
 \hspace*{\algorithmicindent} \textbf{Output}: banded matrices $\BFA$ and $\BFPhi$, and permutation matrix $\BFP$ 
\begin{algorithmic}
\Ensure $\nu$ is a half-integer, $s=\nu+1/2$, $n \geq 2\nu+2$
\State Initialize $\BFA \leftarrow \BFzero\in\Real^{n\times n}$
\State search permutation $\BFP$ to sort $\{x_{i}\}_{i=1}^n$ in increasing order

\For{$i= 1$ to $s$}
\State solve for $\{a_l\}_{l=0}^{s}$: $ \sum_{l=0}^{s}a_l[e^{ x_{i+l}},\ x_{i+l}e^{ x_{i+l}},\  \cdots, \ x_{i+l}^{s-1}e^{ x_{i+l}}\  ]^\transpose =\boldsymbol{0}$
 \State $[\BFA]_{i,i:i+s} \leftarrow (a_0,\cdots,a_{s})$\;
\EndFor
\For{$i= s+1$ to $n-s$}
\State solve $\{a_l\}_{l=-s}^{s}$: $\sum_{l=-s}^{s}a_l[e^{-x_{i+l}},\ e^{x_{i+l}},\  \cdots \ x_{i+l}^{s-1}e^{-x_{i+l}},\ x_{i+l}^{s-1}e^{x_{i+l}}\ ]^\transpose =\boldsymbol{0},$ 
\State $[\BFA]_{i,(i-s):(i+s)} \leftarrow (a_{-s},\cdots,a_{s})$\;
\EndFor
\For{$i= n-s+1$ to $n$}
\State
Solve for $\{a_l\}_{l=-s}^{0}$: $ \sum_{l=-s}^{0}a_l[e^{ -x_{i+l}},\ x_{i+l}e^{ -x_{i+l}},\  \cdots ,\ x_{i+l}^{s-1}e^{- x_{i+l}}\  ]^\transpose =\boldsymbol{0}$
\State $[\BFA]_{i,(i-s):i} \leftarrow (a_{-s},\cdots,a_{0})$\;
\EndFor
\State $[\phi_i(x_j)]_{i,j}=\BFA\BFP^\transpose k(\BFX,\BFX)\BFP$
\end{algorithmic}}
\end{algorithm}

Algorithm \ref{alg:banded_factorization}, which has time complexity of $\CalO(n\log n)$ and space complexity $\CalO(n)$, can generate the following KP factorization based on \eqref{eq:KP_system} and \eqref{eq:KP_one_sided_system} for any one-dimensional Mat\'ern kernel $k(\BFX,\BFX)$:
\begin{equation}
    \label{eq:banded_factorize}
    \BFP^\transpose k(\BFX,\BFX)\BFP=\BFA^{-1}[\phi_i(x^*_j)]_{i,j}.
\end{equation}
Here, $\BFX$ represents any one-dimensional collection of points and $\BFP$ is the permutation matrix that sorts $\BFX$ into ascending order. It is evident that both  $[\phi_i(x^j)]{i,j}$ and $\BFA$ are banded matrices with bandwidths of $\nu-1/2$ and $\nu+1/2$, respectively. The banded nature of these matrices stems from the construction of the KPs $\phi_i$, where each KP is designed to be nonzero only within a specific range: a compact interval for central KPs or a half-interval for one-sided KPs. This localized support ensures that $\BFA$ and the KP evaluation matrix are sparse, with nonzero elements concentrated around the diagonal up to the specified bandwidths. Because both $\BFA$ and $[\phi_i(x^*_j)]_{i,j}$ are banded matrices, and $\BFP$ is permutation matrix, KP can lower the time and space complexities for one-dimensional GPs to $\CalO(n \log n)$ and $\CalO(n)$, respectively.  For KP factorization as formulated in \eqref{eq:banded_factorize} for general kernels , please refer to \cite{ding2024general}.

\subsection{Bayesian Back-fitting}
To address the computational challenges presented in computing the posterior mean \eqref{eq:posterior_block_mean}, one widely adopted solution is the Bayesian Back-fitting algorithm, as introduced by \citet{hastie2000bayesian}. This algorithm utilizes an iterative approach, making it capable of solving target equation $[\BFK^{-1}+\lambda\BFS\BFS^\transpose]^{-1}\lambda\BFS\BFY$ when data size and dimension is large. The pseudocode for this procedure is outlined in Algorithm \ref{alg:bayes_backfit}.

\begin{algorithm}
\caption{Bayesian Back-fitting}\label{alg:bayes_backfit}
\hspace*{\algorithmicindent} \textbf{Input}: Training data $(\BFX,\BFY)$\\
 \hspace*{\algorithmicindent} \textbf{Output}:  estimation of $[\BFK^{-1}+\lambda\BFS\BFS^\transpose]^{-1}\lambda\BFS\BFY$ 
\begin{algorithmic}
\State Initialize $\BFu_d^{(0)}=0$, $d=1,\cdots,D$
\For{$t= 1$ to $T$}
\For{$d= 1$ to $D$}
\State \begin{equation}
    \label{eq:bayes_backfit}
    \BFu_d^{(t)}=[\BFK_d^{-1}+\lambda \rmI_n]^{-1}\lambda\left[\BFY-\sum_{d'<d}\BFu_{d'}^{(t)}-\sum_{d'>d}\BFu_{d'}^{(t-1)}\right]
\end{equation}
\EndFor
\State\textbf{end for}
\EndFor
\State\textbf{end for}
\State return $\BFu=[\BFu_1^{(T)},\cdots,\BFu_D^{(T)}]^\transpose$
\end{algorithmic}
\end{algorithm}

Many one-dimensional GP $\mathcal{G}_d$ can be reformulated as a stochastic differential equation, for example, kernel $k$ is a half-integer Matérn. In this case, the time and space complexities for the computation of \eqref{eq:bayes_backfit} can be lowered to $\mathcal{O}(n\log n)$ and $\CalO(n)$, respectively. This is accomplished by first reformulating $\mathcal{G}_d$ as a state-space equation and then applying the Kalman filter or KP factorization. For more details on the efficient computation of \eqref{eq:bayes_backfit} using GP's state space, please refer to \cite{hartikainen2010kalman,saatcci2012scalable,loper2021general}.

In this paper, we focus on the optimal number of iterations \(T\) necessary for convergence in Bayesian Back-fitting. While \cite{saatcci2012scalable} conjectured that \(T=\mathcal{O}(\log n)\) iterations might suffice, our analysis presents a counterexample showing that, in numerous cases, the minimum number of iterations required is \(\mathcal{O}(n\log n)\).

\section{ Back-fitting Convergence Lower Bound}
\label{sec:lower_bound}
In this section, we outline our strategy for establishing the lower bound of convergence for Bayesian Back-fitting, introduce the essential technique - a global feature that  Bayesian Back-fitting fails to reconstruct efficiently, and conclude with a two-dimensional example to demonstrate the concept.

\subsection{ Proof Overview}
\label{sec:proof_overview}
As described by \cite{hastie2000bayesian,saatcci2012scalable}, the Bayesian Back-fitting technique is analogous to the Gauss-Seidel method, an iterative algorithm used for addressing finite difference problems. In the domain of numerical analysis, it is well known that solving a partial differential equation discretized on a grid of size $\mathcal{O}(n)$ via the Gauss-Seidel method requires $\mathcal{O}(n^2)$ iterations  (see \citealt[ Chapter 7.3]{strang2007computational} or \citealt[Chapter 13.2.3]{saad2003iterative}). This is due to Gauss-Seidel's nature as a filter that rapidly reduce high-frequency noise while exhibiting a marked insensitivity to the solution's low-frequency Fourier components \citep{strang2007computational,saad2003iterative}.

Unlike  the Gauss-Seidel method, where observations are collected from a regular grid, the data points in Bayesian Back-fitting are scattered. Additionally,  Gauss-Seidel aims to invert a finite difference matrix induced by noiseless observations on a regular grid while Bayesian Back-fitting aims to invert a kernel matrix induced by scattered, noisy data. As a result, the rationale behind the low convergence rate observed in the Gauss-Seidel approach does not directly applicable to Bayesian Back-fitting. To establish the lower bound for the convergence rate of Bayesian Back-fitting, we employ the KPs discussed in Section \ref{sec:KP}. Through these, we construct global features to which Bayesian Back-fitting is insensitively adjusted. 

In our counterexample, the covariates $\BFX$ are sampled from a \textit{Latin Hypercube} design (LHD) \citep{mckay2000comparison}, an experimental design frequently used for generalized linear models with regularization. More specifically, for each dimension of $\BFX$, the $n$ elements are chosen such that one element comes from each number  \(\frac{1}{n}, \frac{2}{n}, \ldots, 1\), and these elements are then randomly permuted. Let $\BFX^*=[1/n,2/n,\cdots,1]$ be a set consisting of sorted points. Then, in a LHD, the following identity holds for any $d$: $\BFX_d=\BFP_d\BFX^*$
where $\BFP_d$ is a permutation matrix uniform distributed on the permutation group. In each outer iteration $t$ of Bayesian Back-fitting, the total $D$ inner iterations can be summarized as
\begin{equation}
    \label{eq:back_fit_update}
    \BFu^{(t)}=[\BFK^{-1}+\lambda \rmI_{Dn}+\lambda \BFL]^{-1}\lambda\left(\BFS\BFY-\BFL^\transpose\BFu^{(t-1)}\right)
\end{equation}
where $\BFL\in\Real^{Dn\times Dn}$ is a lower triangular matrix with zero diagonal as follows: 
\[\BFL=\begin{bmatrix}
    \boldsymbol{0} & \boldsymbol{0} &\cdots& \boldsymbol{0} &\boldsymbol{0} \\
    \rmI_n &  \boldsymbol{0}&\cdots &\boldsymbol{0}&\boldsymbol{0} \\
    \vdots&\vdots &\ddots & \vdots&\vdots\\
    \rmI_n& \rmI_n&\cdots&\rmI_n &\boldsymbol{0}
\end{bmatrix}
\]

Let $\BFu^{\infty}$ denote the solution. It must be a stationary point of the iteration \eqref{eq:back_fit_update}.  Define error in the $t$-th iteration of Back-fitting as $\BFvarepsilon^{(t)}=\BFu^{(t)}-\BFu^{(\infty)}$. We then can rewrite \eqref{eq:back_fit_update} in the error form as follows:
\begin{equation}
    \label{eq:error_form}
     \BFvarepsilon^{(t)}=\left(-[\BFK^{-1}+\lambda \rmI_{Dn}+\lambda \BFL]^{-1}\lambda\BFL^\transpose\right)^t\BFvarepsilon^{(0)}=\CalS_n^t\BFS\BFY.
\end{equation}
for $t=1,\cdots,T$. The analysis of the convergence for Bayesian Back-fitting then involves quantifying the distance between  $\|\CalS_n^T\BFS\BFY\|_2$ and 0, where $\|\cdot\|_2$ represents the vector $l^2$ norm. To achieve this, a global feature is constructed using KPs as follows:
\begin{equation}
\label{eq:blobal_feature}
    \frac{1}{\sqrt{n}}\boldsymbol{1}=\BFV_*=\sum_{i=1}^n\phi_i(\BFX^*)=\BFA k_d(\BFX^*,\BFX^*)[1,1,\cdots,1]^\transpose.
\end{equation}
where $\boldsymbol{1}=[1,\cdots,1]$. By proving that the analysis across each dimension can be broken down into proving the inequality $\BFV_*^\transpose[k_d(\BFX^*,\BFX^*)]^{-1}\BFV_*\leq \CalO(1/n)$ for each dimension, and, consequently,
\[
\lambda \BFV_*^\transpose\left[\left[k_d(\BFX^*,\BFX^*)\right]^{-1}+\lambda \rmI_n\right]^{-1}\BFV_* \geq \BFV_*^\transpose\left[\rmI_n- \lambda^{-1}\left[k_d(\BFX^*,\BFX^*)\right]^{-1}\right]\BFV_* \geq 1-\CalO\left(\frac{1}{\lambda n}\right),
\]
we can derive the lower bound of the maximum eigenvalue of $\CalS_n$. This, in turn, establishes a lower bound on the convergence rate.

\begin{theorem}
\label{thm:back-fit_lower_bound}
    Let $\BFX$ be a LHD and $\BFY$ be generated by  additive GP with kernel $k=\sum_{d=1}^Dk_d$ where each $k_d$ satisfies Assumption \ref{assump:kernel}. Let $\BFu$ be the outputs by Algorithm \ref{alg:bayes_backfit} with input $(\BFX,\BFY)$ and iteration number $t$. Then we have the following lower bound:
    \[\E\|\BFu-[\BFK^{-1}+\lambda\BFS\BFS^\transpose]^{-1}\lambda\BFS\BFY\|_2\geq \CalO\left((1-\frac{1}{\lambda n})^{t}\right).\]
\end{theorem}
Theorem \ref{thm:back-fit_lower_bound} shows the requisite number of iterations for Algorithm \ref{alg:bayes_backfit} to yield accurate results. Specifically,  it demonstrates that at least $T=\mathcal{O}(n\log n)$ iterations are needed for the discrepancy between the true posterior mean and the estimate produced by Algorithm \ref{alg:bayes_backfit} to diminish to $o(1)$. 

\subsection{Global Feature for  \texorpdfstring{Mat\'ern-$1/2$}{Lg} Kernels}

To understand the rationale behind constructing the target global feature, let's begin with the case for Mat\'ern-$1/2$ kernel. Utilizing a proposition from \cite{ding2022sample}, we can directly derive the desired result.

\begin{proposition}[Proposition 1 \cite{ding2022sample}]
\label{prop:ding_inverseK}
    Let $k$ be a Mat\'ern-$1/2$ kernel. Let $\BFX^*=\{i/n=ih\}_{i=1}^n$.   $\BFA$ is  then a tridiagonal matrix 
    \begin{align*}
    &\BFA^*_{i,i}=\left\{\begin{array}{lc}
            \frac{e^{\omega h}}{2\sinh(\omega h)} & \text{if} \quad i=1,n \\
             \frac{\sinh(2\omega h)}{2\sinh(\omega h)^2} & \text{otherwise}
        \end{array}\right. \ \ \BFA^*_{i,i+1}=\BFA^*_{i,i-1}=\frac{-1}{2\sinh(\omega h)},
\end{align*}
and $\BFA k(\BFX^*,\BFX^*)=\rmI_n=[\phi_i(jh)]_{i,j}$.
\end{proposition}
For the base case $\nu=1/2$, it is obvious that $\sum_i\phi_i(\BFX^*)=\rmI_n\boldsymbol{1}=\boldsymbol{1}$. If we normalized it by letting $\BFV_*=\frac{1}{\sqrt{n}}\boldsymbol{1}$, then obviously we can have
\begin{align*}
    \BFV_*^\transpose[k(\BFX^*,\BFX^*)]^{-1}\BFV_*=&\frac{1}{n}\left(\BFA^*_{1,1}+\BFA^*_{1,2}+[\sum_{i=2}^{n-1}\BFA^*_{i,i-1}+\BFA^*_{i,i}+\BFA^*_{i,i+1}]+\BFA^*_{n,n-1}+\BFA^*_{n,n}\right)\\
    =&\frac{1}{n}\left(\frac{e^{\omega h}-1}{\sinh(\omega h)}+(n-2)\frac{\sinh(2\omega h)/\sinh(\omega h) -2}{2\sinh{\omega h}}\right)=\CalO(\frac{1}{n})
\end{align*}
where the last line is from Taylor expansions of $e^{\omega x}$ and $\sinh(x)$ around 0. Also, $\BFV_*$ is invariant under any permutation $\BFP$. So we can directly show that, for Mat\'ern-$1/2$ kernel, the smoothing operator $\CalS_n$ cannot adjust the normalized global feature $\BFV_*$ efficiently for any random permuted point set $\BFX=\BFP \BFX^*$:
\[\lambda \BFV_*^\transpose\left[[k(\BFX,\BFX)]^{-1}+\lambda \rmI_n\right]^{-1}\BFV_*\geq \BFV_*^\transpose\left[\rmI_n- \lambda^{-1}[k(\BFX,\BFX)]^{-1}\right]\BFV_*\geq 1-\CalO(\frac{1}{\lambda n}). \]
So we can show that the largest eigenvalue of $\lambda \left[[k(\BFX,\BFX)]^{-1}+\lambda \rmI_n\right]^{-1}$ is close to $1$.

We initially introduce the Mat\'ern-\({1}/{2}\) kernel as it is foundational for constructing global features for the general Mat\'ern-\(\nu\) kernel. In the following subsection, we will demonstrate that, due to the convolutional nature of Mat\'ern kernels, \(\BFV_*\) continues to satisfy the above inequality when the Matérn-\({1}/{2}\) kernel is replaced with another Matérn-\(\nu\) kernel.

\subsection{Global Feature for general Mat\'ern Kernels}
Now we can build the global feature by induction on $\nu$. The essential idea relies on the following convolution identities between KPs of Mat\'ern-$\nu$ and Mt\'ern-$1/2$ kernels:
\begin{proposition}
\label{prop:convolution_A}
    Suppose $\BFA_{\nu}$ is a KP factorization matrix associated to Mat\'ern-$\nu$ kernel matrix  $k_{\nu}(\BFX^*,\BFX^*)$. Construct matrix $\BFA$ through the following convolution operations:
    \begin{align*}
    &[\BFA]_{i,i:i+\nu+3/2}=\left[[\BFA_{\nu}]_{i,i:i+\nu+1/2}\ \ 0\right ]\BFA^*_{1,1}+\left[0\ \ [\BFA_{\nu}]_{i,i:i+\nu+1/2} \right]\BFA^*_{1,2},\ i\leq \nu+3/2\\
    &[\BFA]_{i,(i-\nu-3/2):i}=\left[[\BFA_{\nu}]_{i,(i-\nu-1/2):i}\ \ 0\right ]\BFA^*_{n-1,n}+\left[0\ \ [\BFA_{\nu}]_{i,(i-\nu-1/2):i} \right]\BFA^*_{n,n},\ i\geq n-\nu-1/2;
\end{align*}
and 
\begin{align*}
    &[\BFA]_{i,(i-\nu-3/2):(i+(i+\nu+3/2)}=\left[[\BFA_{\nu}]_{i,(i-\nu-1/2):((i+\nu+1/2))}\ \ 0 \ \ 0 \right ]\BFA^*_{i,i-1}\nonumber \\
    &+\left[0\ \ [\BFA_{\nu}]_{i,(i-\nu-1/2):((i+\nu+1/2))} \ \ 0 \right]\BFA^*_{i,i}
    +\left[0\ \   0 \ \ [\BFA_{\nu}]_{i,(i-\nu-1/2):((i+\nu+1/2))}  \right]\BFA^*_{i,i+1}.
\end{align*}
for $i=\nu+5/2,\cdots,n-\nu-3/2$. Then $\BFA$ is a KP factorization matrix of Mat\'ern-$(\nu+1)$ kernel matrix  $k_{\nu+1}(\BFX^*,\BFX^*)$.
\end{proposition}

\begin{proposition}
\label{prop:convolution_Phi}
    Let $\BFA_\nu$ be the KP matrix constructed via convolutions in Proposition \ref{prop:convolution_A}. Let $\phi_0^{l}$, $\phi_0^{c}$, and $\phi_0^{r}$ be the left-sided, central, and right-sided KP of Mat\'ern-$1/2$ kernel, respectively as follows:
    \begin{align*}
        \phi_0^{l}(x)=\frac{e^{-\omega(|x|-h)}-e^{-\omega|x-h|}}{2\sinh(\omega h)}\boldsymbol{1}_{\{x\leq h\}},\quad \phi_0^{r}(x)=\frac{e^{-\omega(|x-1|-h)}-e^{-\omega|x-1+h|}}{2\sinh(\omega h)}\boldsymbol{1}_{\{x\geq n-h\}},
    \end{align*}  
    \begin{equation*}
        \phi_0^{c}(x) =\left\{\begin{array}{lc}
           \frac{\sinh(\omega (x-h)}{\sinh(\omega h))}\  &\text{if} \quad x\in(h,2h)\\
             \frac{\sinh(\omega(3h-x))}{\sinh(\omega h)} \  &\text{if}\quad x\in(2h,3h)\\
             0 \ &\text{otherwise}
        \end{array}\right.  .
    \end{equation*}
    Specifically, $\phi^{l}_0$, $\phi^{c}_0$, and $\phi^{r}_0$ are the Mat\'ern-$1/2$ KPs descried in Proposition \ref{prop:ding_inverseK} induced by points $[ih]_{i=0}^1$, $[ih]_{i=-1}^1$, and $[ih]_{i=n-1}^n$, respectively. Then for Mat\'ern-$\nu$ KP matrix $[\phi_i(x^*_j)]$ associated to $\BFA_\nu$, we have
    \begin{align*}
     \phi_i(x^*_j) =\left\{\begin{array}{lc}
           \underbrace{\phi_0^{l}*\cdots*\phi_0^{l}}_{\nu+1/2\ \text{convolutions}}((j-i)h)\  &\text{if} \quad i\leq \nu+1/2\\
             \underbrace{\phi_0^{c}*\cdots*\phi_0^{c}}_{\nu+1/2\ \text{convolutions}}((j-i)h)\  &\text{if} \quad \nu+3/2\leq i\leq n-\nu-1/2\\
              \underbrace{\phi_0^{r}*\cdots*\phi_0^{r}}_{\nu+1/2\ \text{convolutions}}((j-i)h)\  &\text{if} \quad i\geq n-\nu+1/2
        \end{array}\right. .
\end{align*}
    
\end{proposition}

\begin{figure}[ht] 
    \centering
    \includegraphics[width=.5\linewidth]{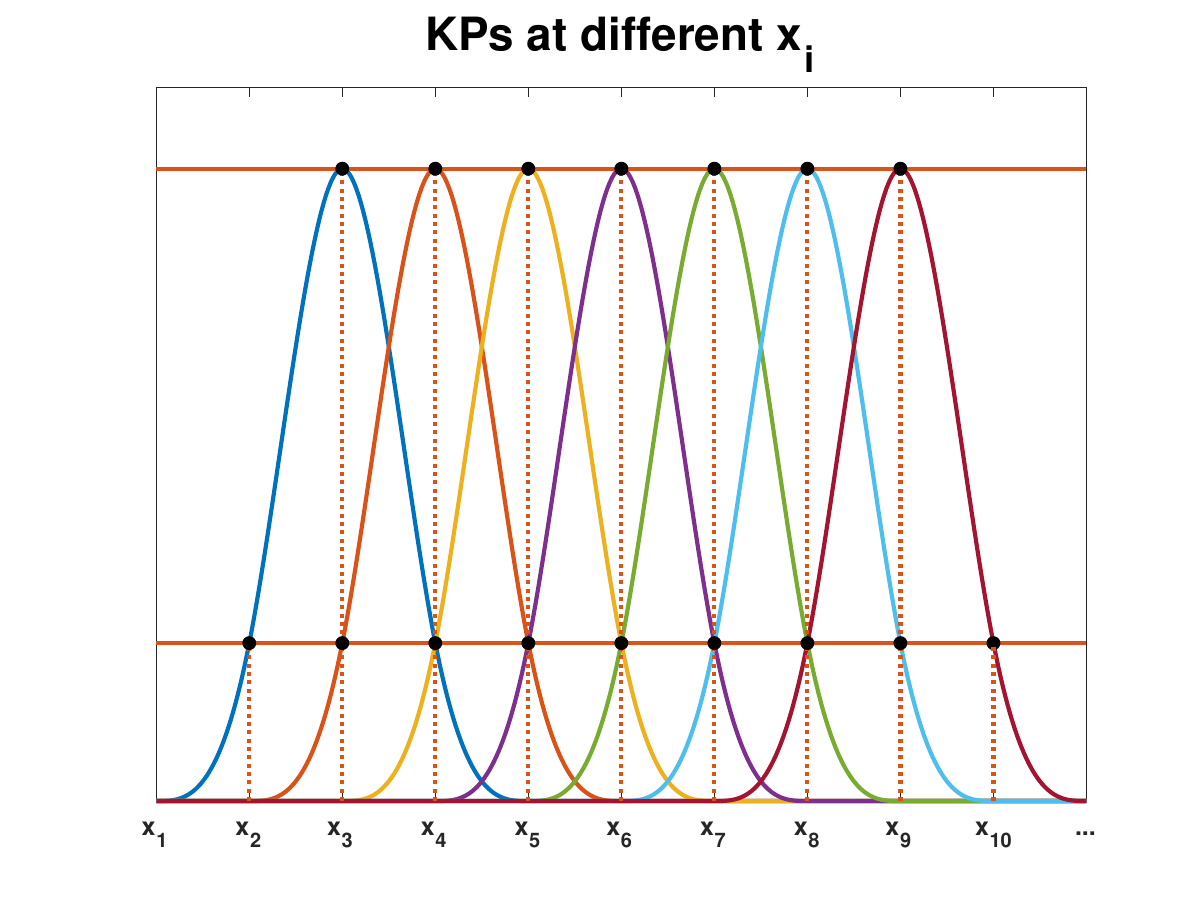}

    \caption{ $\sum_i\phi_i(x^*_j)$ can be normalized to $1$ for any $x^*_j$, as KPs induced by $\BFX^*={ih}$ at different points have identical values.}
    \label{fig:central_KP}
\end{figure}

Propositions \ref{prop:convolution_A} and \ref{prop:convolution_Phi} reveal that for evenly spaced points $\BFX^*$, the KP factorization corresponding to a higher order Matérn kernel can be derived through the convolution of factorization matrices associated with lower order Matérn kernels. Leveraging the convolution identities provided by these propositions and scaling operation to each row of $\BFA_\nu$, we can propose the following regarding the desired KP factorization:

\begin{proposition}
    \label{prop:KP_estimation}
    For half-integer Mat\'ern-$\nu$ kernel $k_\nu$ and $\BFX^*=\{ih\}_{i=1}^n$, there exists KP factorization $\BFA k(\BFX^*,\BFX^*) =[\psi_i(x^*_j)]_{i,j}$ such that
    \begin{align}
    &\sum_{j}\BFA_{i,j}\asymp\left\{\begin{array}{lc}
          1 \  \text{if} \quad i\leq \nu+1/2\ \text{or} \ i\geq n-\nu+1/2 \\
             \frac{1}{n} \  \text{otherwise}
        \end{array}\right.\nonumber\\
        &[\sum_j\psi_j(\BFX^*)]_{i}=\sum_j\psi_j(x^*_i)=[\boldsymbol{1}^\transpose\BFA k_{\nu}(\BFX^*,\BFX^*)]_i = 1\quad\forall i=1,\cdots,n \label{eq:global_feature_5}
\end{align}
where $a_n\asymp b_n$ denotes the asymptotic relation $0<\lim_{n\to \infty}|\frac{a_n}{b_n}|<\infty$.
\end{proposition}
The second equation in \eqref{eq:global_feature_5} results from the sparsity of KPs, the even spacing of $\BFX^*$, and the KPs' translation invariance, as outlined in Theorem 3 and Theorem 10 in \cite{chen2022kernel}. A graphical illustration is provided in Figure \ref{fig:central_KP}. Detailed proofs for Propositions \ref{prop:convolution_A} to \ref{prop:KP_estimation} are provided in the Appendix.

From \eqref{eq:global_feature_5}, we can use the connection between Mat\'ern kernels and kernels satisfying Assumption \ref{assump:kernel} to have the following estimate:
\begin{theorem}
    \label{thm:global_feature}
    For any kernel $k$ satisfying Assumption \ref{assump:kernel} and $\BFX^*$ from a LHD, we have
    \[\frac{\boldsymbol{1}^\transpose}{\sqrt{n}}\left[\lambda \rmI_n+[k(\BFX^*,\BFX^*)]^{-1}\right]^{-1}\lambda \frac{\boldsymbol{1}}{\sqrt{n}}\geq 1-\CalO(\frac{1}{\lambda n}). \]
    Hence, the largest eigenvalue of $\lambda \left[\lambda \rmI_n+[k(\BFX^*,\BFX^*)]^{-1}\right]^{-1}$ is greater than $1-\CalO(\frac{1}{\lambda n}) $.
\end{theorem}

\subsection{A Two-dimensional Counterexample}
\label{sec:counter_example_2d}
We now use a two-dimensional case to illustrate the counterexample. For notation simplicity, let $\lambda=1$.
 The two-dimensional error form can be written explicitly as follows:
\begin{align*}
    &\BFvarepsilon^{(t)}_1=-[\BFK_1^{-1}+\rmI_n]^{-1}[\BFK_2^{-1}+\rmI_n]^{-1}\BFvarepsilon_1^{(t-1)}\\
    &\BFvarepsilon^{(t)}_2=-[\BFK_2^{-1}+\rmI_n]^{-1}[\BFK_1^{-1}+\rmI_n]^{-1}\BFvarepsilon_2^{(t-1)}
\end{align*}
for any $k_1$ and $k_2$ satisfying Assumption \ref{assump:kernel}. Let  $\BFX$ be a LHD, i.e.,  $\BFX_i=\BFP_i
\BFX^*$. We then estimate how the error associated to $\BFV_*$ changes in each iteration. Let $\BFV_*=\boldsymbol{1}/\sqrt{n}$, then 
 \begin{align*}
     & \BFV_*^\transpose\left([\BFK_1^{-1}+\rmI_n]^{-1}[\BFK_2^{-1}+\rmI_n]^{-1}\right)\BFV_*\\
     =&\BFV_*^\transpose\left(\left[ \rmI_n-[\BFK_1+\rmI_n]^{-1}\right]\left[\rmI_n-[\BFK_2+\rmI_n]^{-1}\right]\right)\BFV_*\\
     \geq &\BFV_*^\transpose\left(\rmI_n- [\BFK_1+\rmI_n]^{-1}-[\BFK_2+\rmI_n]^{-1}\right)\BFV_*\\
     \geq & \left|1-\BFV_*^\transpose\BFK_{1}^{-1}\BFV_*-\BFV_*^\transpose\BFK_{2}^{-1}\BFV_*\right|\\
     \geq &1-\CalO(\frac{1}{n})
 \end{align*}
where the last line is because $\BFV_*$ is invariant under any permutation.

Because $\BFV_*$ is a normalized vector, we can conclude that the largest eigenvalue of the Back-fit operator $\CalS_n$ is greater than $1-\CalO(\frac{1}{n})$. Let $\BFE_*$ be the eigenvector associated to the largest eigenvalue $\lambda_*\geq 1-\CalO(\frac{1}{n})$. Then the projection of initial error $\BFY=\sum_{d=1}^2\CalG_d(\BFX_d)+\varepsilon$ onto $\BFE_*$ can be lower bounded simply as follows:

\begin{align*}
    \E\|\BFE_*^\transpose\left(\sum_{d=1}^2\CalG_d(\BFX_d)+\varepsilon\right)\|_2^2\geq \E|\sum_{i=1}^n[\BFE_*]_i\varepsilon_i|^2=\sigma_y^2.
\end{align*}
It can be directly deduced that in each iteration, at most a $1-\CalO(\frac{1}{n})$ fraction of the error projected onto $\BFE_*$ is eliminated. By induction, the fraction of error eliminated after $t$ iterations is at most $(1-\CalO(\frac{1}{n}))^t$:
\[\E\|\CalS_n^t\left(\sum_{d=1}^2\CalG_d(\BFX_d)+\varepsilon\right)\|_2\geq \left(\BFE_*^\transpose\CalS_n\BFE_*\right)^t\sigma_y\geq (1-\CalO(\frac{1}{n}))^t.\]

 \section{Kernel Multigrid for Back-fitting} \label{secKMG}
From the posterior mean \eqref{eq:posterior_block_mean_single} and posterior variance \eqref{eq:posterior_block_var_single} of a single dimension, our objective can be summarize as efficiently solving kernel matrix equation of the form 
\begin{equation}
\label{eq:goal_KMG}
   \BFc= [\lambda^{-1}\BFK^{-1}+\BFS\BFS^\transpose]^{-1}\BFv
\end{equation}
for general given $\BFv\in\Real^{Dn}$.   In Section \ref{sec:lower_bound}, we have shown that the suboptimal performance of $\CalS_n$ stems from its inefficiency in eliminating errors associated with global features. Nevertheless, this limitation can be effectively overcome by accurately estimating those global features and this leads to our algorithm. In this section, we first introduce sparse GPR, which is the key component of KMG. We then introduce the detailed KMG algorithm and its convergence property for approximating the solution of \eqref{eq:goal_KMG}.

\subsection{Sparse  Gaussian Process Regression }
Before we introduce sparse GRP, let us first establish some basic notations. We select $m$ inducing points $\BFU=\{(u_{i,1},\cdots,u_{i,D})\}_{i=1}^m$ from the dataset $\BFX$. The notation $\BFU_d = \{u_{i,d}\}_{i=1}^m$ represents the set of values in the $d$-th dimension of $\BFU$. Define block diagonal matrix $\BFK_{n,n} = \text{diag}[k_{1}(\BFX_1,\BFX_1), \cdots, k_{D}(\BFX_D,\BFX_D)]$, which comprises $D$ one-dimensional kernel covariance matrices for the data. Similarly, $\BFK_{m,m}$ is introduced as the block diagonal matrix $\BFK_{m,m} = \text{diag}[k_{1}(\BFU_1,\BFU_1), \cdots, k_{D}(\BFU_D,\BFU_D)]$, pertaining to the inducing points. Finally, $\BFK_{m,n} = \BFK_{n,m}^\transpose$ represents the block diagonal matrix $\BFK_{m,n} = \text{diag}[k_{1}(\BFU_1,\BFX_1), \cdots, k_{D}(\BFU_D,\BFX_D)]$, indicating the covariance between the data points ${\BFX_d}$ and the inducing points ${\BFU_d}$. 

For notation simplicity, we let $\lambda=1$. Our sparse GPR of \eqref{eq:goal_KMG} induced by $\BFU$ is then defined as 
\begin{equation}
        \label{eq:sparse_GPR}
   \hat{\BFc}=\BFK_{n,m}\BFK_{m,m}^{-1}[\BFK_{m,m}^{-1}+\BFSigma_{m,m}]^{-1}\BFK_{m,m}^{-1}\BFK_{m,n}\BFv
\end{equation}
where  \[\BFSigma_{m,m}=\BFK_{m,m}^{-1}\BFK_{m,n}\BFS\BFS^\transpose\BFK_{n,m}\BFK_{m,m}^{-1}.\] In \eqref{eq:sparse_GPR}, the operation $\BFv_m = \BFK_{m,m}^{-1}\BFK_{m,n}\BFv$ represents the projection of $\BFv$ onto the $mD$-dimensional Hilbert space spanned by kernel functions $\CalH_m = \{k_d(\cdot,u_{i,d}) \mid d=1,\ldots,D, i=1,\ldots,n\}$. This operation is analogous to the coarsening step in Algebraic Multigrid. By mapping $\BFv$ to  $\CalH_m$, a similar problem to the original one can be formulated in the coarser space.

The projection $\BFK_{m,m}^{-1}\BFK_{m,n}$ is analogous to the  Galerkin projection \citep{saad2003iterative}. We treat $\BFK_{n,n}^{-1}+\BFS\BFS^\transpose$ and its inverse as  linear operators. By applying the concept of Galerkin projection, an equivalent operator can be defined in the coarser space $\CalH_m$ as:
\[\BFK_{m,m}^{-1}\BFK_{m,n}[\BFK^{-1}_{n,n}+\BFS\BFS^\transpose]\BFK_{n,m}\BFK_{m,m}^{-1}=\BFK^{-1}_{m,m}+\BFK_{m,m}^{-1}\BFK_{m,n}\BFS\BFS^\transpose\BFK_{n,m}\BFK_{m,m}^{-1},\]
which leads to a different penalty term $\BFSigma_{m,m}$. Once the coarser problem is solved, the solution can be mapped back to the original function space by interpolation operator $\BFK_{n,m}\BFK_{m,m}^{-1}$ to have an approximated solution of \eqref{eq:goal_KMG}. 

Obviously, the accuracy of the approximation \eqref{eq:sparse_GPR} relative to the original problem \eqref{eq:goal_KMG} improves as the number of inducing points increases. As $\BFU=\BFX$, meaning all data points are selected as inducing points, \eqref{eq:sparse_GPR} is reduced to the original problem \eqref{eq:goal_KMG}.  Identifying the ideal sparse GPR for our KMG algorithm involves pinpointing the optimal quantity and arrangement of inducing points. The following theorem offers guidance on selecting the necessary inducing points.

\begin{theorem}[Approximation Property]
    \label{thm:sparse_GPR}
    Suppose $\BFc=[\lambda^{-1}\BFK_{n,n}^{-1}+\BFS\BFS^\transpose]^{-1}\BFv$ for some vector $\BFv\in\Real^{Dn}$ and $k_d$ satisfies Assumption \ref{assump:kernel} for all $d=1,\cdots,D$. Suppose inducing points $\BFU$ satisfy 
    \begin{equation}
        \label{eq:fill_distance}
    \max_{d=1,\cdots,D}\max_{x\in[0,1]}\min_{u\in\BFU_d}|x-u|\leq \CalO(h_m)
    \end{equation}
    for some $h_m=o(1)$. 
    Then  the sparse GPR approximation $\hat{\BFc}$ has the following error rate:
    \[\frac{1}{\sqrt{n}}\|\BFc-\hat{\BFc}\|_2\leq 
    C^*[\frac{h_m^{s}}{\sqrt{\lambda \kappa^*_n\kappa_m^*}}+\sqrt{\lambda n}h_m^{2s}]\sqrt{\BFc^\transpose\BFK^{-1}_{n,n}\BFc}\]
    where $\kappa^*_m$ and $\kappa^*_n$ are called \textit{restricted isometry constants} and only depend on the distributions of points $\BFU$ and $\BFX$, respectively, and $C^*$ is come universal constant independent of $n$, $m$, $D$, and $\lambda$.
\end{theorem}
\begin{remark}
    The restricted isometry constant was first proposed in \cite{Dantzig_Selector}. It is an important geometry characteristics for the function spaces $\CalH_m$ under the empirical distributions of data. This constant reflects the "orthogonality"  between the Reproducing Kernel Hilbert Spaces (RKHSs) induced by those one-dimensional kernel functions, such as $k_d$ and $k_{d'}$, for instance. For more details on RKHS, please refer to \cite{wendland2004scattered,paulsen2016introduction,steinwart2008support}.
\end{remark}
\begin{remark}
    Condition \eqref{eq:fill_distance} can be readily met in a variety of scenarios due to its focus on the fill distance in a single dimension, in contrast to the  definition of the original fill distance. For instance, if the input data points $\BFX$ are uniformly distributed within the hypercube $[0,1]^D$, this condition can be naturally satisfied by selecting points uniformly.
\end{remark}

We leave the proof of Theorem \ref{thm:sparse_GPR} in the appendix. Leveraging sparse GPR, we are now prepared to outline our algorithm and discuss its convergence properties.

\subsection{Kernel Multigird}

The basic idea of KMG relies on two facts. Firstly, Back-fitting can be interpreted as a filtering process that reduces high-frequency errors. Secondly, sparse GPRs are particularly effective in capturing low-frequency components by explicitly discarding high-frequency components, albeit at the cost of reduced accuracy compared to full GPRs. Consequently, by combining Back-fitting with sparse GPR, the KMG method effectively mitigates the weaknesses of each approach, resulting in enhanced overall performance. The details of the KMG algorithm are as shown in Algorithm \ref{alg:KMG}.

\begin{algorithm}[ht]
\caption{Kernel Multigrid}\label{alg:KMG}
\hspace*{\algorithmicindent} \textbf{Input}:  data points $\BFX$, vectors $\{\BFv_d\in\Real^{n}\}_{d=1}^D$, number of inducing points $m$\\
 \hspace*{\algorithmicindent} \textbf{Output}: estimation of  $[{\lambda^{-1}}\BFK_{n,n}^{-1}+\BFS\BFS^\transpose]^{-1}[\BFv_1,\BFv_2,\cdots,\BFv_D]^\transpose$ 
\begin{algorithmic}
\State Initialize $\BFu_d^{(0)}$, $d=1,\cdots,D$
\State Uniformly select $m$ inducing points $\BFU_d$ from $\BFX_d$, $d=1,\cdots,D$
\For{$t= 1$ to $T$}
\For{$d= 1$ to $D$}
\State \begin{equation}
    \label{eq:KMG_backfit}
    \BFu_d^{(t)}=[\lambda^{-1}\BFK_d^{-1}+\rmI_n]^{-1}\left[\BFv_d-\sum_{d'<d}\BFu_{d'}^{(t)}-\sum_{d'>d}\BFu_{d'}^{(t-1)}\right]
\end{equation}
\EndFor
\State\textbf{end for}
\State $\BFr_n=[\BFv_1,\BFv_2,\cdots,\BFv_D]^\transpose-[{\lambda^{-1}}\BFK_{n,n}^{-1}+\BFS\BFS^\transpose][\BFu_1^{(t)},\BFu_2^{(t)},\cdots,\BFu_D^{(t)}]^\transpose$\Comment{residual}
\State $\BFr_m=\BFK_{m,m}^{-1}\BFK_{m,n}\BFr_n$ \Comment{projection}
\State $\BFdelta_m=[\lambda^{-1}\BFK_{m,m}^{-1}+\BFSigma_{m,m}]^{-1}\BFr_m$\Comment{coarser  problem}
\State $[\BFu^{(t)}_1,\cdots,\BFu^{(t)}_D]^\transpose\leftarrow[\BFu^{(t)}_1,\cdots,\BFu^{(t)}_D]^\transpose+\BFK_{n,m}\BFK^{-1}_{m,m}\BFdelta_m$\Comment{interpolation}
\EndFor
\State\textbf{end for}
\State return $\BFu=[\BFu^{(T)}_1,\cdots,\BFu_D^{(T)}]^\transpose$
\end{algorithmic}
\end{algorithm}

Efficient computations of the KMG algorithm can be divided into two main parts. The first part encompasses the execution of the Back-fitting step, as described by equation \eqref{eq:KMG_backfit}. This phase requires solving $D$ linear systems in the form $[\lambda^{-1}\BFK_d^{-1}+\rmI_n]^{-1}\BFu$. Utilizing the KP technique \citep{ding2024general} or state-space model \citep{hartikainen2010kalman}, as outlined in Section \ref{sec:KP}, enables the resolution of all $D$ linear systems within a time complexity of $\CalO(n\log n)$ and a space complexity $\CalO(n)$. We focus on the second part, which entails applying a sparse GPR to the residual. The second part can be further decomposed into four steps: computing the residual, performing projection, solving the coarser problem, and performing interpolation.

For computing the residual $\BFr_n$, we use KP factorization \eqref{eq:banded_factorize}: 
\begin{align}
    \BFr_n&=[\BFv_1,\BFv_2,\cdots,\BFv_D]^\transpose-[{\lambda^{-1}}\BFK_{n,n}^{-1}+\BFS\BFS^\transpose][\BFu_1^{(t)},\BFu_2^{(t)},\cdots,\BFu_D^{(t)}]^\transpose\nonumber\\
    &=[\BFv_1,\BFv_2,\cdots,\BFv_D]^\transpose-[\lambda^{-1}\BFP\BFPhi^{-1}\BFA\BFP^\transpose+\BFS\BFS^\transpose][\BFu_1^{(t)},\BFu_2^{(t)},\cdots,\BFu_D^{(t)}]^\transpose\label{eq:residual_KP}
\end{align}
where $\BFA=\text{diag}[\BFA_1,\cdots,\BFA_D]$, $\BFPhi=\text{diag}[\BFPhi_1,\cdots,\BFPhi_D]$, $\BFP=\text{diag}[\BFP_1,\cdots,\BFP_D]$  and   $\BFA_d$ and $\BFPhi=[\phi_i(\BFx_j)]_{i,j}$ are the KP matrices and $\BFP_d$ is the permutation matrix, all induced by points $\BFX_d$.  Here, matrix multiplications involving the sparse matrices $\BFP$, $\BFA$, and $\BFS\BFS^\transpose$ can all be computed in $\CalO(Dn)$ time and space. Similarly, the matrix multiplication involving the inverse banded matrix $\BFPhi^{-1}$ can be solved in $\CalO(Dn)$ time and space complexities using a banded matrix solver.

For computing the projection and interpolation, recall that  inducing point $\BFU_d$   are selected from data points $\BFX_d$ and $\BFK_{m,m}=\text{diag}[k_{1}(\BFU_1,\BFU_1),k_{2}(\BFU_2,\BFU_2),\cdots,k_{D}(\BFU_D,\BFU_D)]$ so, using KP factorization, $\BFK_{m,m}^{-1}\BFK_{m,n}$ can also be factorized as
\begin{equation}
    \BFK_{m,m}^{-1}\BFK_{m,n}=\BFP_{m,m}\BFPhi_{m,m}^{-1}\BFphi_{m}(\BFX) \label{eq:projection_KP}
\end{equation}
where $\BFP_{m,m}$, $\BFPhi_{m,m}$, and $\BFphi_{m}(\BFX)$ correspond to the matrices $\BFP$, $\BFPhi$, and $\BFphi$ introduced in the previous step but induced by the inducing points $\{\BFU_d\}_{d=1}^D$. Notably, $\BFP_{m,m}$ is a permutation matrix, $\BFPhi_{m,m}$ is a banded matrix, and $\BFphi_{m}(\BFX)$ is a sparse matrix with $\CalO(Dn)$ non-zero entries due to the compact support property of KPs. As a result, any matrix multiplication of the form $\BFv_m\BFK_{m,m}^{-1}\BFK_{m,n}$ or $\BFK_{m,m}^{-1}\BFK_{m,n}\BFv_n$ can be computed in $\CalO(Dn)$ time and space for any vector $\BFv_m\in\Real^{Dm}$ and $\BFv_n\in\Real^{Dn}$. This efficiency arises from the sparsity of $\BFP_{m,m}$ and $\BFphi_{m}(\BFX)$, as well as the  banded structure of $\BFPhi_{m,m}$.

Finally, solving the coarser problem is notably more efficient due to its significantly smaller scale compared to the original problem. This step requires only $\CalO((Dm)^3)$ time and $\CalO((Dm)^2)$ space to complete, as it involves matrices of size $Dm$-by-$Dm$ only.

In summary, each iteration of the KMG algorithm can be executed efficiently, with the overall time and space complexities for each iteration being $\CalO(Dn\log n+(Dm)^3)$ and $\CalO(Dn+(Dm)^2)$, respectively.

\subsection{Convergence Analysis}
The KMG algorithm can be summarized as the following steps in each of the $t$-th iteration
\begin{enumerate}
    \item Smoothing: $\BFu_n^{(t+1/2)}=\CalS_n\BFu_n^{(t)}$
    \item Get residual: $\BFr_n=\BFv-[\lambda^{-1}\BFK^{-1}+\BFS\BFS^\transpose]\BFu_n^{(t+1/2)}$
    \item Coarsen: $\BFr_m=\BFK^{-1}_{m,m}\BFK_{m,n}\BFr_n$
    \item Solve in the coarsen scale: $\BFdelta_m=[\lambda^{-1}\BFK^{-1}_{m,m}+\BFSigma_{m,m}]^{-1}\BFr_m$ 
    \item Correct: $\BFu_n^{(t+1)}=\BFu_n^{(t+1/2)}+\BFK_{n,n}\BFK^{-1}_{m,m}\BFdelta_m$.
\end{enumerate}
These steps closely mirror those of the Algebraic Multigrid method, with the key difference being that our objective is  GPR rather than solving a differential equation. It is clear that the result of one iteration of the above algorithm corresponds to
a iteration process of the form
\begin{equation}
\label{eq:KMG_iteration}
    \BFu^{(t+1)}_n=\CalT^{m}_{n}\CalS_n\BFu_n^{(t)}
\end{equation}
where 
\[\CalT^{m}_{n}=\rmI_{Dn}- \BFK_{n,m}\BFK_{m,m}^{-1}[\lambda^{-1}\BFK_{m,m}^{-1}+\BFSigma_{m,m}]^{-1}\BFK_{m,m}^{-1}\BFK_{m,n}[\lambda^{-1}\BFK^{-1}+\BFS\BFS^\transpose]\]
executes sparse GPR on the residual and subsequently incorporates the correction into the solution and $\CalS_n=[\lambda^{-1}\BFK_{n,n}^{-1}+\rmI_{Dn}+\BFL]^{-1}\BFL^\transpose$ is the back-fit operator. Both $\CalS_n$ and $\CalT^m_n$ are linear operators. Thus, the convergence analysis of \eqref{eq:KMG_iteration} simplifies to estimating the distance between $\|(\CalT^m_n\CalS_n)^t\BFS\BFY\|_2$ and $0$ as before.

From the representation theorem, any function  in the $Dn$-dimensional RKHS $\CalH_n=\text{span}\{k_d(\cdot,x_{i,d}):d=1,\cdots,D,i=1,\cdots,n\}$ can be represented by the $Dn$-dimensional vector $\BFv$ equipped with the RKHS norm: $\|\BFv\|_{\CalH_n}^2:=\BFv^\transpose\BFK^{-1}\BFv$. 

From \cite[Theorem 12.1]{wendland2004scattered}, the RKHS norm for any $L^2$ function $f$ within $\CalH_n$ can increase at a rate significantly beyond $n$. Assuming Assumption \ref{assump:kernel}, 
there exists vector $\BFv$ such that $ \BFv^\transpose\BFK^{-1}\BFv\geq \CalO\left({n^{2s-1}}\right)$. Given that our initial error is the noisy observation vector $\BFvarepsilon^{(0)}=\BFS\BFY$, it is reasonable to infer that the RKHS norm of $\BFvarepsilon^{(0)}$ is significantly substantial. The smoothing function of $\CalS_n$ can effectively diminish the error contributing to this substantial RKHS norm:
\begin{lemma}[Smoothing Property]
\label{lem:smoothing_property}
    Given error $\BFvarepsilon^{(t)}$ in the $t$-th iteration for any $t$,  we have \[ \|\CalS_n \BFvarepsilon^{(t)}\|_{\CalH_n}\leq  \sqrt{\lambda}D\|\BFvarepsilon^{(t)}\|_2.\]
\end{lemma}
We refer Lemma \ref{lem:smoothing_property} as the \textit{smoothing property} of $\CalS_n$, signifying that the Back-fit operator $\CalS_n$ effectively smooths out any error with an RKHS norm exceeding $\sqrt{\lambda} D\|\BFvarepsilon^{(t)}\|_2$. 

\begin{remark}
    The smoothing property in our study is different from the one commonly used in solving numerical PDEs. This difference arises because numerical PDEs are typically approached as interpolation problems, whereas additive GPs  are regression problems. In the scenario of interpolation, it becomes evident that we would have $\lambda=\sigma_y^{-2}=\infty$ as per Lemma \ref{lem:smoothing_property}, leading to the conclusion that the smoothing property is not applicable.
\end{remark}

In addition to smoothing property of $\CalS_n$, we have already proved the \textit{approximation property} of $\CalT^m_n$. As a result, by combining Theorem \ref{thm:sparse_GPR} and Lemma \ref{lem:smoothing_property}, we can directly have the following estimate for any error $\BFvarepsilon^{(t)}$ :
\begin{align*}
    \|\CalT_{n}^m\CalS_n \BFvarepsilon^{(t)}\|_2\leq &C^*\sqrt{n}[\frac{h_m^{s}}{\sqrt{\lambda \kappa^*_n\kappa_m^*}}+\sqrt{\lambda n}h_m^{2s}]\|\CalS_n\BFvarepsilon^{(t)}\|_{\CalH_n} \\
    \leq&  C^*\sqrt{\lambda n}D[\frac{h_m^{s}}{\sqrt{\lambda \kappa^*_n\kappa_m^*}}+\sqrt{\lambda n}h_m^{2s}] \|\BFvarepsilon^{(t)} \|_2.
\end{align*}
We then can use induction argument to have the following  convergence rate  of KMG:
\begin{theorem}
\label{thm:KMG_convergence}
    Suppose Assumption \ref{assump:kernel} holds. Let the  inducing points $\BFU$ satisfy
     \begin{equation}
     \label{eq:KMG_condition}
         C^*\sqrt{\lambda n}D[\frac{h_m^{s}}{\sqrt{\lambda \kappa^*_n\kappa_m^*}}+\sqrt{\lambda n}h_m^{2s}]\leq 1-\delta
     \end{equation}
      for some $\delta>0$ independent of data size $n$. Then for any initial guess $\BFvarepsilon^{(0)}$, we have the following error bound for the error of KMG
    \[\|\BFvarepsilon^{(t)}\|_2 \leq \left(1-\delta\right)^t \|\BFvarepsilon^{(0)}\|_2.\]
\end{theorem}
\begin{remark}
    According to our selection of inducting points, the error term $1-\delta$ must be independent of $n$.  For instance, we can choose inducing points in a manner that ensures \(1-\delta \leq e^{-1}\). Under this scenario, for any \(\epsilon > 0\), merely \(\epsilon \log n\) iterations are required to attain an accuracy of the order \(\mathcal{O}(n^{-\epsilon})\).
\end{remark}
\begin{remark}
Condition \eqref{eq:KMG_condition} is readily achievable for a wide class of data distributions. For instance, if $\BFX$ follows a uniform distribution or represents a LHD over the space $[0,1]^D$, then selecting $m$ points at random from $\BFX$ will yield an inducing point set $\BFU$ whose one-dimensional fill distance, as defined in \eqref{eq:fill_distance}, is approximately $h_m=\CalO\left(\frac{1}{m}\right)$ (ignoring the potential log term). Under these conditions, it suffices to randomly select $m\approx n^{\frac{1}{2s}}$  inducing points.
\end{remark}

\section{Numerical Experiments}\label{secNumeric}
To illustrate the theories in the previous sections, we first test the convergence rate of Back-fitting on the global feature $\BFV_*$. We then evaluate the performance of KMG on both synthetic and real datasets, showcasing its effectiveness in practical scenarios.

All the experiments are implemented in Matlab (version 2023a) on a laptop computer with macOS 13.0, Apple M2 Max CPU, and 32 GB of Memory. Reproducible codes are available at \url{https://github.com/ldingaa}.

\subsection{Lower Bound of Back-fitting}
To demonstrate that Back-fitting struggles with efficiently reconstructing global features, we measure the reduction in error during each iteration of the error form \eqref{eq:error_form}, starting with a randomly selected initial error $\BFvarepsilon^{(0)}\sim\mathcal{N}(0,\rmI_{n})$.

\begin{figure}[ht]
    \centering
    \includegraphics[width=.32\linewidth]{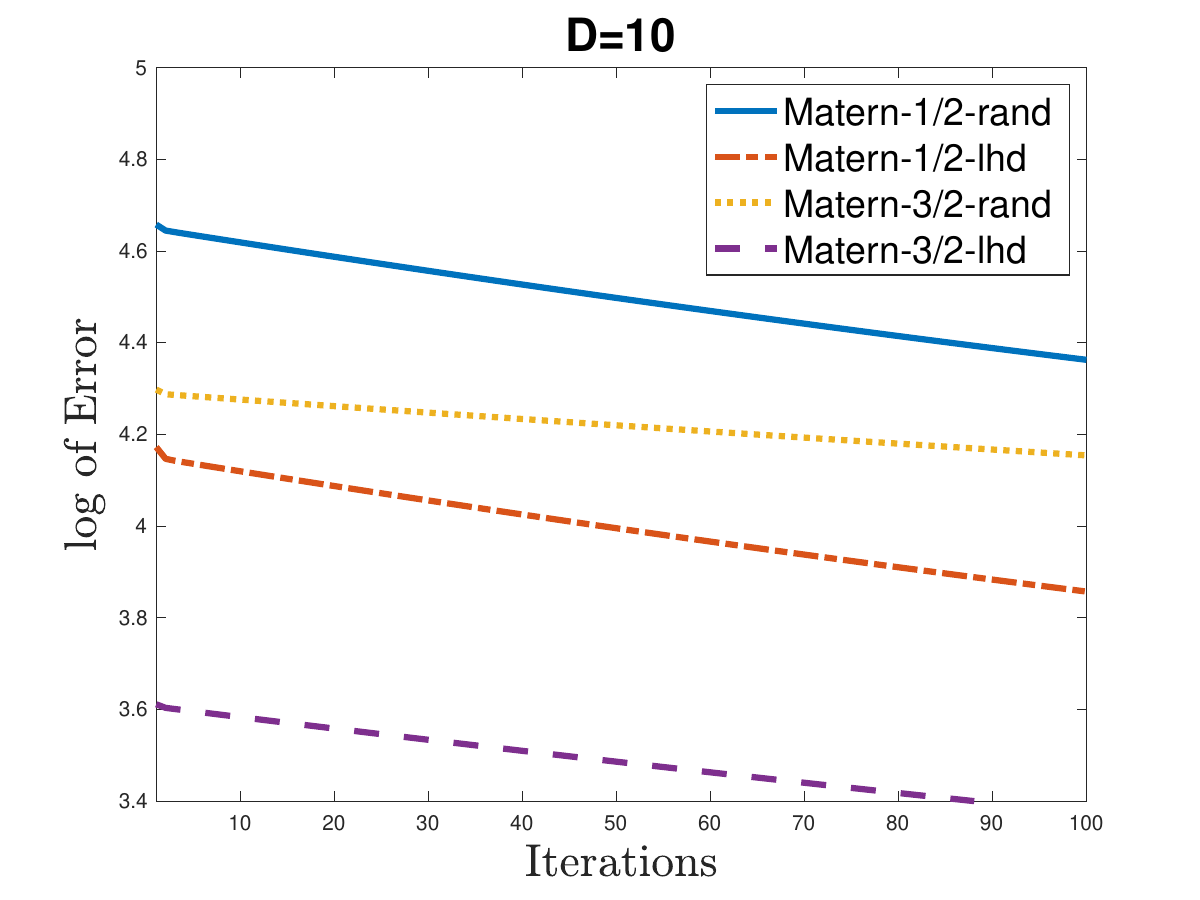}
    \includegraphics[width=.32\linewidth]{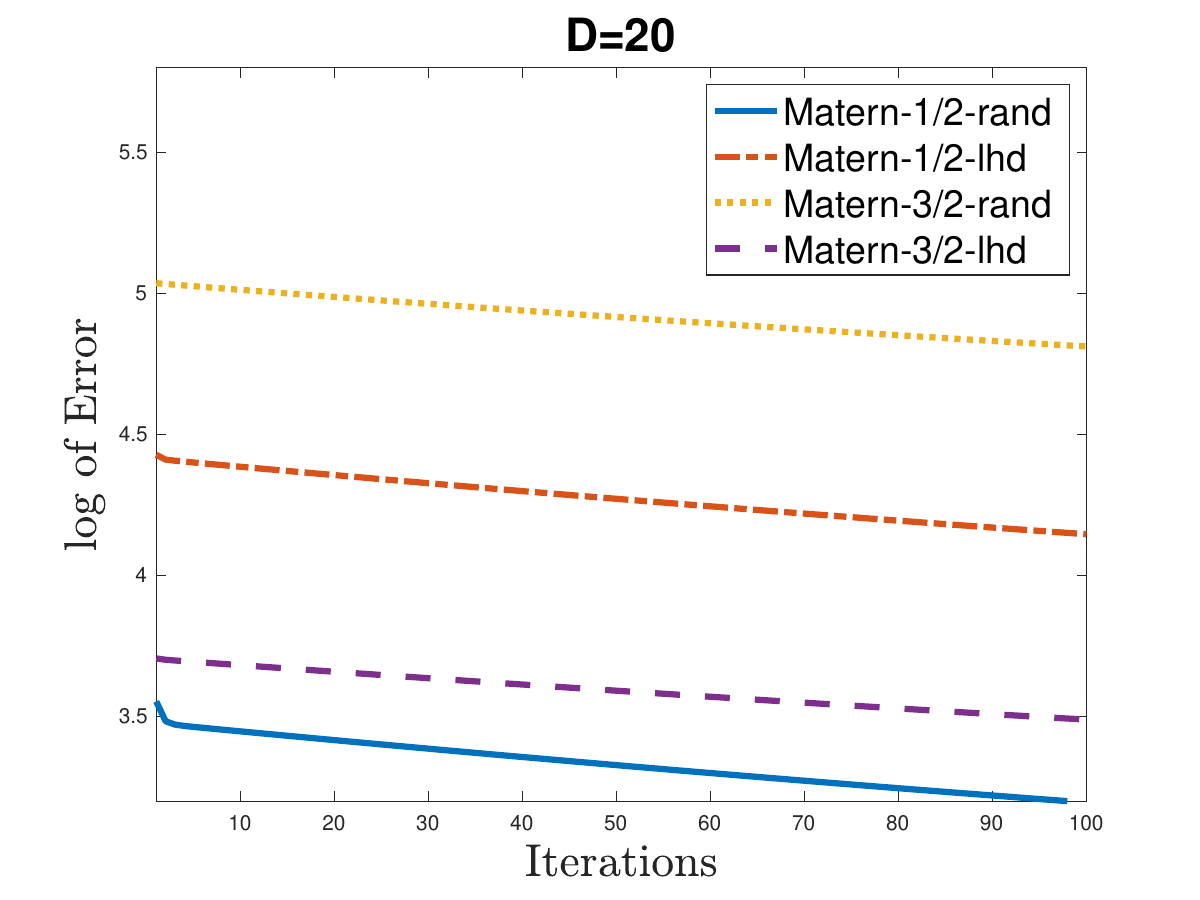}
    \includegraphics[width=.32\linewidth]{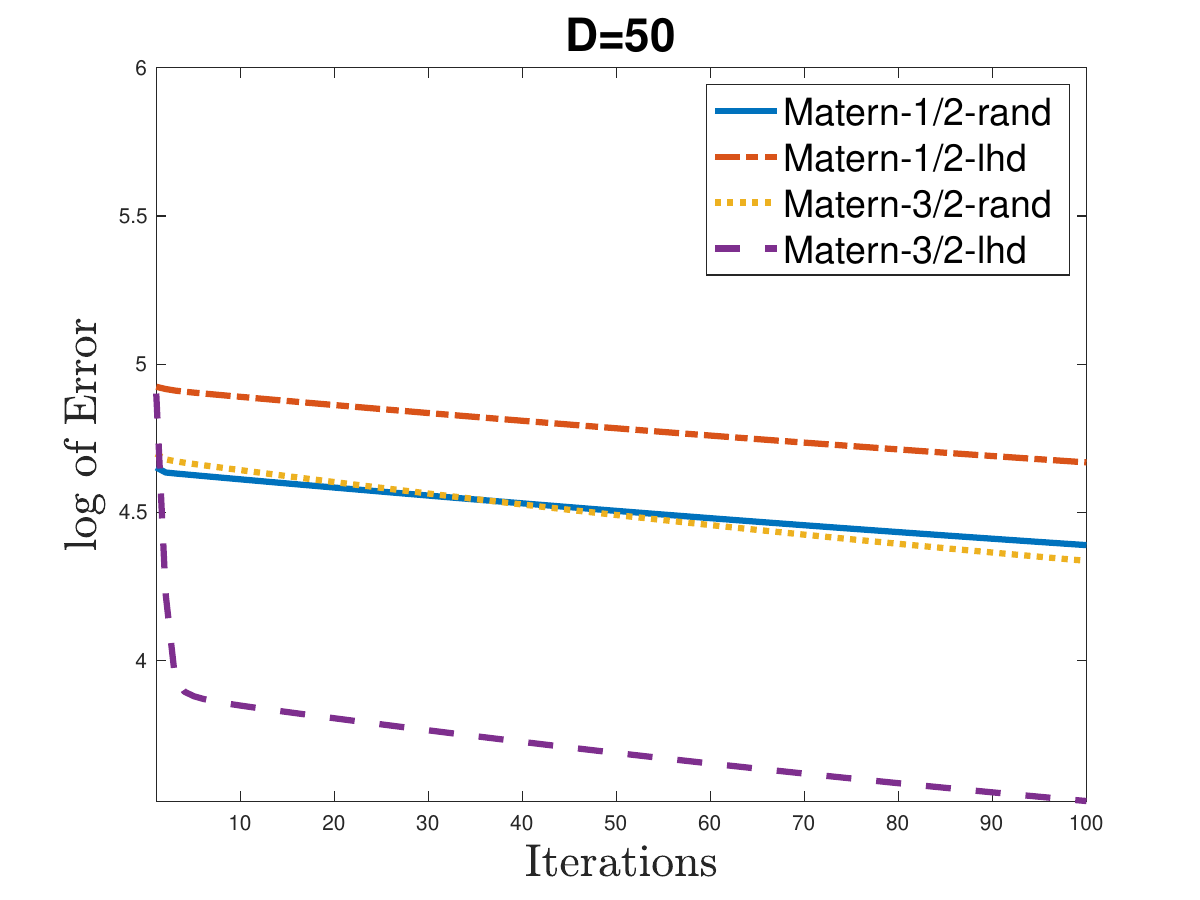}
     \includegraphics[width=.32\linewidth]{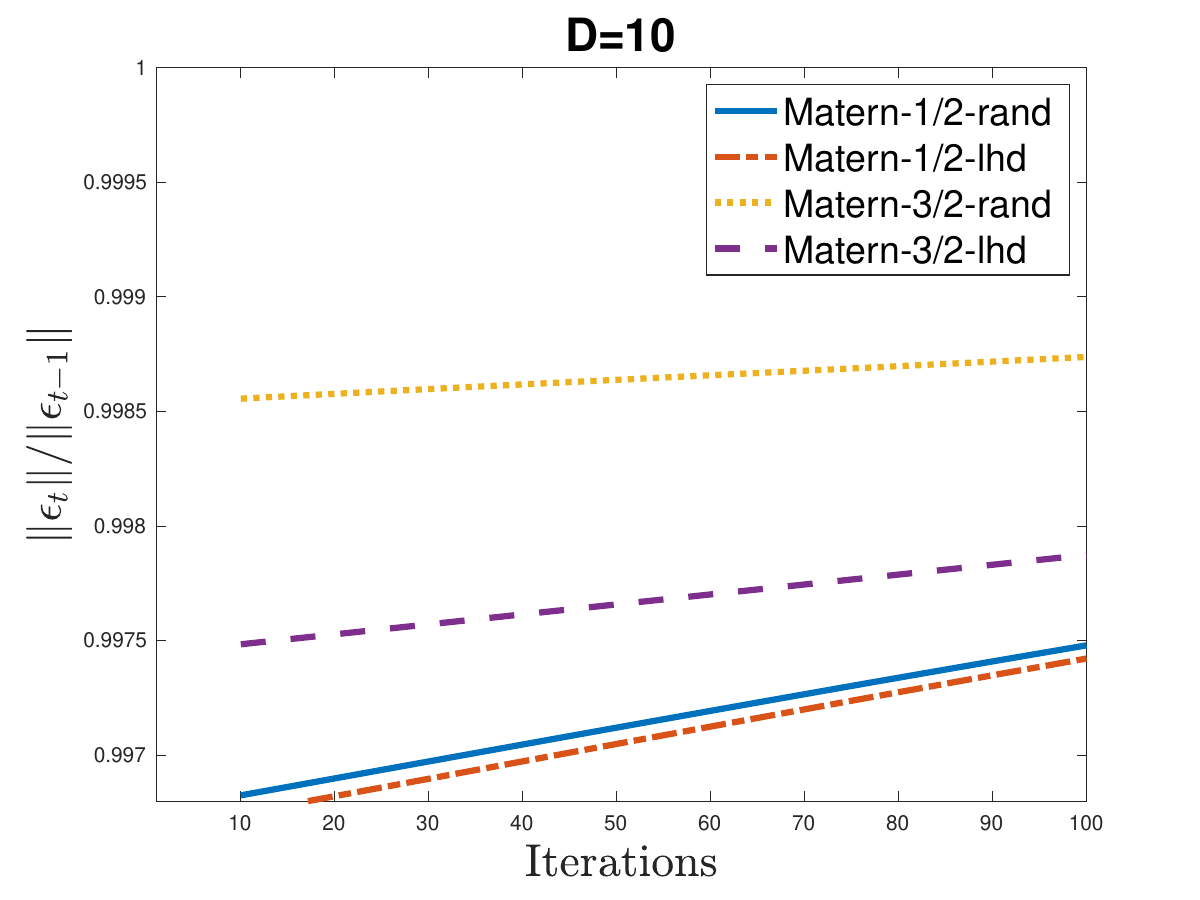}
    \includegraphics[width=.32\linewidth]{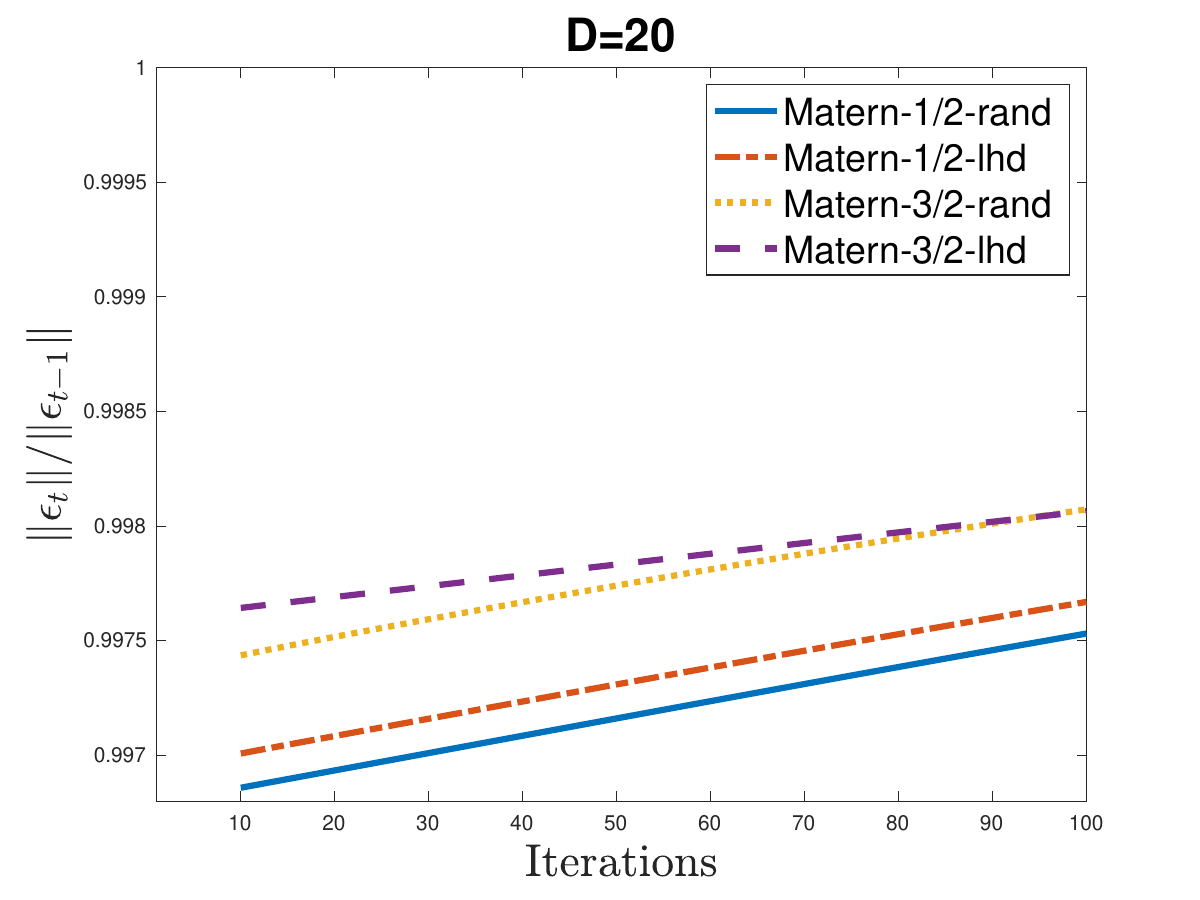}
    \includegraphics[width=.32\linewidth]{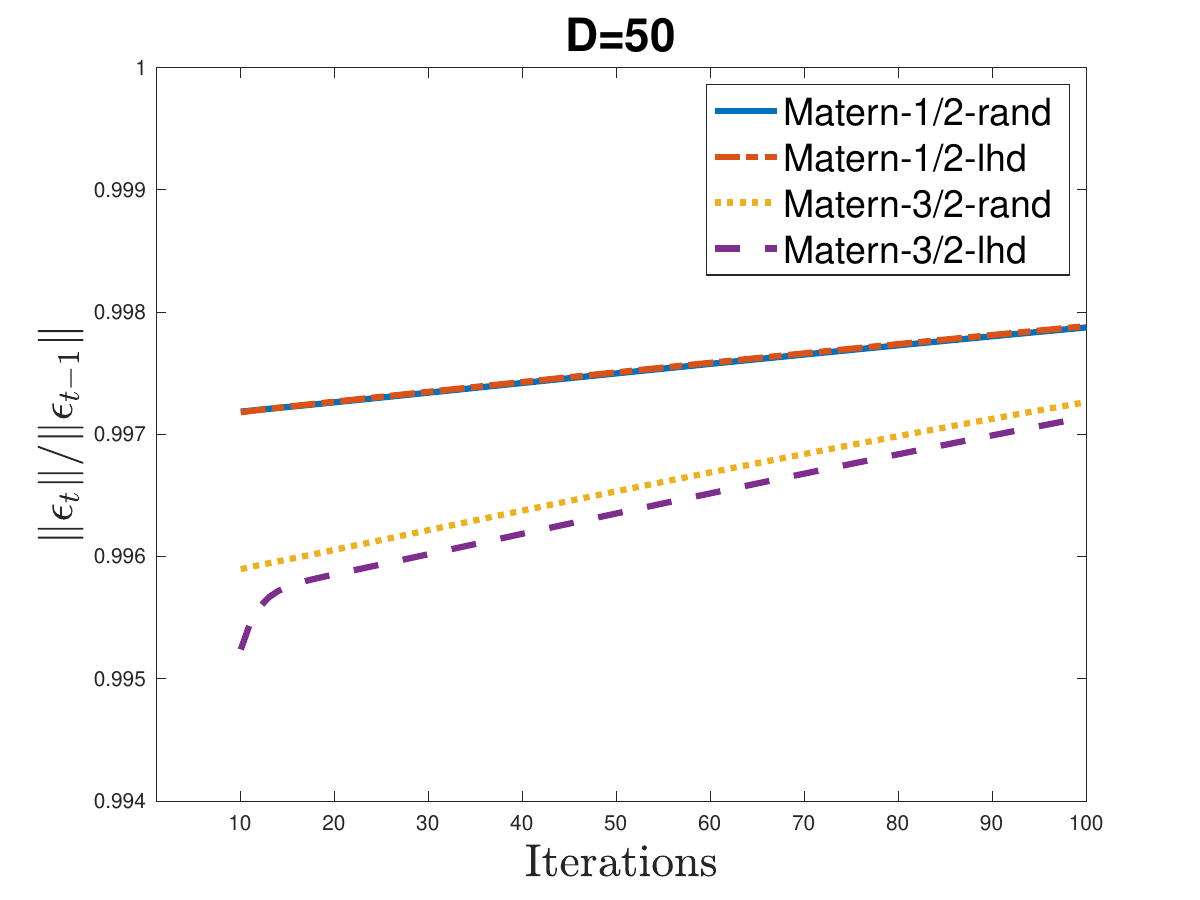}
    \caption{Upper row: log of error decreases with number of iterations; lower row: error ratio $\|\BFvarepsilon^{(t)}\| /\|\BFvarepsilon^{(t)-1}\|$ is close to our lower bound}
    \label{fig:lower_bound}
\end{figure}

Our experiments employ additive GPs with additive-Matérn-$\frac{3}{2}$ and additive-Matérn-$\frac{1}{2}$ kernels, conducted under two distinct scenarios: one where the sampling points $\BFX_n$ are uniformly and randomly positioned within the hypercube $[0,1]^D$, termed \textit{random design}, and another scenario where $\BFX_n$ are arranged according to a LHD in $[0,1]^D$, similar to our approach in the counterexample. Therefore, each comparison involves four unique configurations, represented by the various combinations of kernels and sampling designs. The algorithms under examination are named as follows:
\begin{enumerate}
    \item \textbf{Mat\'ern-$\nu$-rand}: Back-fitting using a Mat\'ern-$\nu$ kernel, where the sample points $\BFX_n$ are distributed uniformly at random within $[0,1]^D$ with  size $n=500$;
    \item \textbf{Mat\'ern-$\nu$-lhd}: Back-fitting using a Mat\'ern-$\nu$ kernel, where the sample points $\BFX_n$ are  arranged according to a LHD in $[0,1]^D$ with size $n=500$;.

\end{enumerate}
We assess each algorithm across dimensions $D=10, 20$, and $50$. For every algorithm, we conduct 100 iterations. At the conclusion of each iteration, we record the norm $\|\BFvarepsilon^{(t)}\|_2$ for each algorithm. Subsequently, we generate plots for both the logarithm of the error norm, $\log |\BFvarepsilon^{(t)}|_2$, and the improvement ratio, $\|\BFvarepsilon^{(t)}\|_2/\|\BFvarepsilon^{(t-1)}\|_2$. The outcomes are displayed in Figure \ref{fig:lower_bound}.

In alignment with Theorem \ref{thm:back-fit_lower_bound}, it is established that:
\[\log \|\BFvarepsilon^{(t)}\|_2\geq Ct\log[1-\CalO(\frac{1}{n})],\quad \frac{\|\BFvarepsilon^{(t)}\|_2}{\|\BFvarepsilon^{(t-1)}\|_2}\geq 1-\CalO(\frac{1}{n}).\]
Examination of Figure \ref{fig:lower_bound} permits us to ascertain that our empirical findings are in precise agreement with these theoretical lower bounds. In the first row of Figure \ref{fig:lower_bound}, the slope of the log error curves becomes significantly close to 0 after the initial few iterations. This indicates that once the high-frequency errors are mitigated, the reduction of error attributable to global features becomes remarkably small in each subsequent iteration. In the second row of Figure \ref{fig:lower_bound}, it is evident that the error ratio $\frac{\|\BFvarepsilon^{(t)}\|}{\|\BFvarepsilon^{(t-1)}\|}$ starts from a minimum of $.997$. Considering our theoretical lower bound for this ratio is $1-\CalO(1/n)$ and given $n=500$ in our experiments, the observed error ratio of approximately $1-.003$ in the experiments aligns closely with our theoretical prediction of $1-\CalO(1/500)$.

To summarize, our numerical experiments corroborate the validity of the lower bound established in Theorem \ref{thm:back-fit_lower_bound}.

\subsection{Comparison on Synthetic Data}
In this subsection, we utilize synthetic data to evaluate the performance of our KMG algorithm in comparison to Back-fitting. Following the approach of the previous section, our experiments implement additive Gaussian Processes with both additive-Matérn-$1/2$ and additive-Matérn-$3/2$ kernels, under two distinct scenarios: one involving sampling points $\BFX_n$ distributed randomly within $[0,1]^D$, and another with $\BFX_n$ derived from a Latin Hypercube Design (LHD) within the same domain. Experiments are conducted across dimensions $D=10$, $20$, and $50$. For each dimension $D$,  we let data size $n=10D$. Specifically, data size $n=100$ for dimension $D=10$, $n=200$ for dimension $D=20$, and $n=500$ for $n=50$. Data are generated considering $\hat{D}$ as the effective dimension:
\[\BFY_n = \sum_{d=1}^{\hat{D}}\CalG_d(\BFX_d) + \BFvarepsilon\]
where $\BFvarepsilon \sim \CalN(0,1)$ represents observation noise following a standard normal distribution,  effective dimensions set to $\hat{D}=3$ for $D=10$, $\hat{D}=5$ for $D=20$, and $\hat{D}=10$ for $D=50$ and each hidden underlying one-dimensional GP $\CalG_d$ employs a kernel function identical to either Mat'ern-$1/2$ or Mat'ern-$3/2$, which are identical to the kernels utilized by the competing algorithms. Therefore, data $(\BFX_n,\BFY_n)$ are summation of \textit{hidden data} $(\BFX_d,\CalG(\BFX_d))$ plus observation noise. The objective for the competing algorithms is to accurately reconstruct the hidden target function $\CalG_d(\BFX_d)$ from these observations.

The target functions, which serve as the approximation objectives for both KMG and Back-fitting, are defined as
\begin{equation}
\label{eq:experiment_goal}
    [{f}^*_1,\cdots,{f}^*_D]^\transpose = [\BFK^{-1} + \BFS\BFS^\transpose]^{-1}\BFS\BFY,
\end{equation}
as specified in \eqref{eq:goal_KMG}. The competing algorithms are outlined as follows:
\begin{enumerate}
    \item \textbf{KMG-rand}: Kernel Multigrid with $m=10$ inducing points $\BFU_m$ and sample points $\BFX_n$  distributed uniformly at random within $[0,1]^D$;
    \item \textbf{KMG-lhd}: Kernel Multigrid  with $m=10$ inducing points $\BFU_m$ and  sample points $\BFX_n$ arranged according to a LHD in  $[0,1]^D$;
    \item \textbf{Backfit-rand}: Back-fitting with sample points $\BFX_n$  distributed uniformly at random within $[0,1]^D$;
    \item \textbf{Backfit-lhd}: Back-fitting with sample points $\BFX_n$  on a LHD in  $[0,1]^D$.
\end{enumerate}

\begin{figure}[ht]
    \centering
    \includegraphics[width=.32\linewidth]{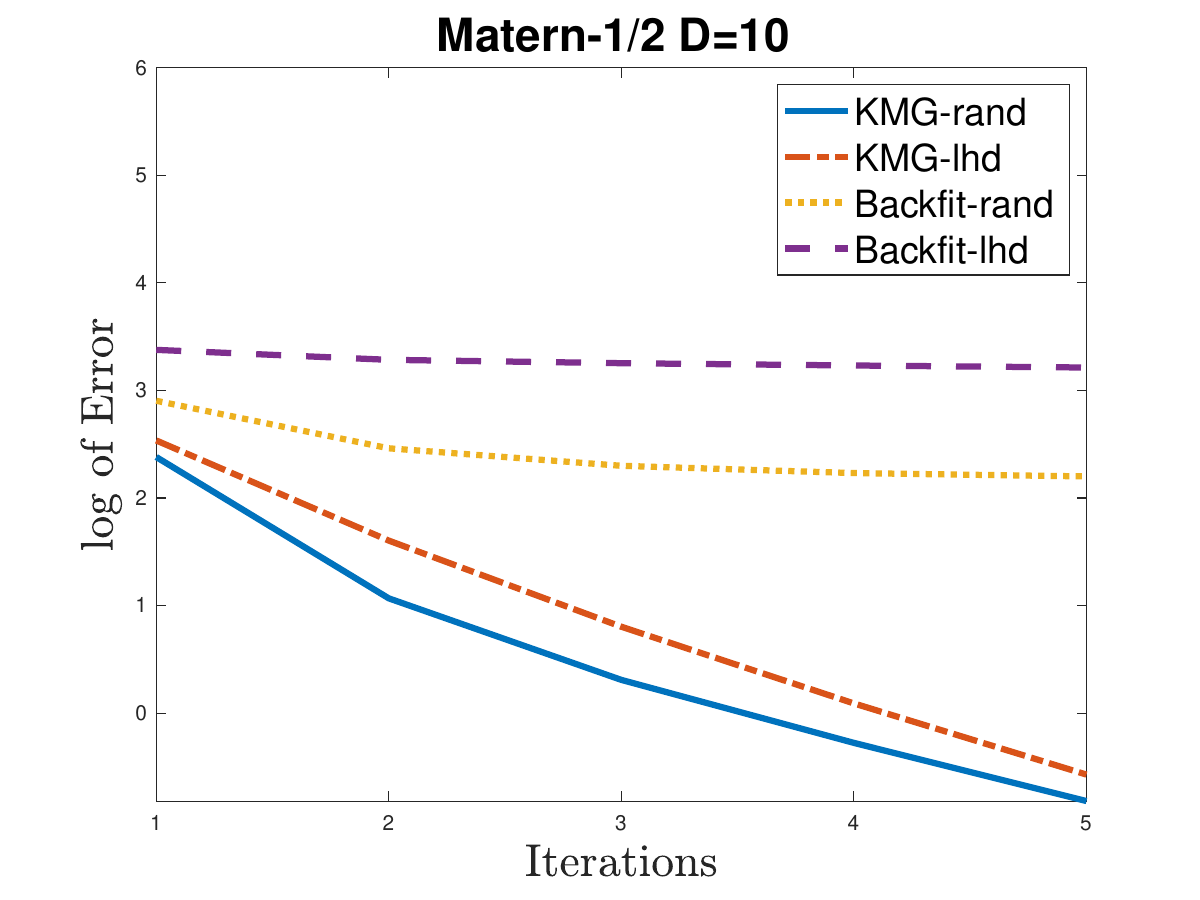}
   \includegraphics[width=.32\linewidth]{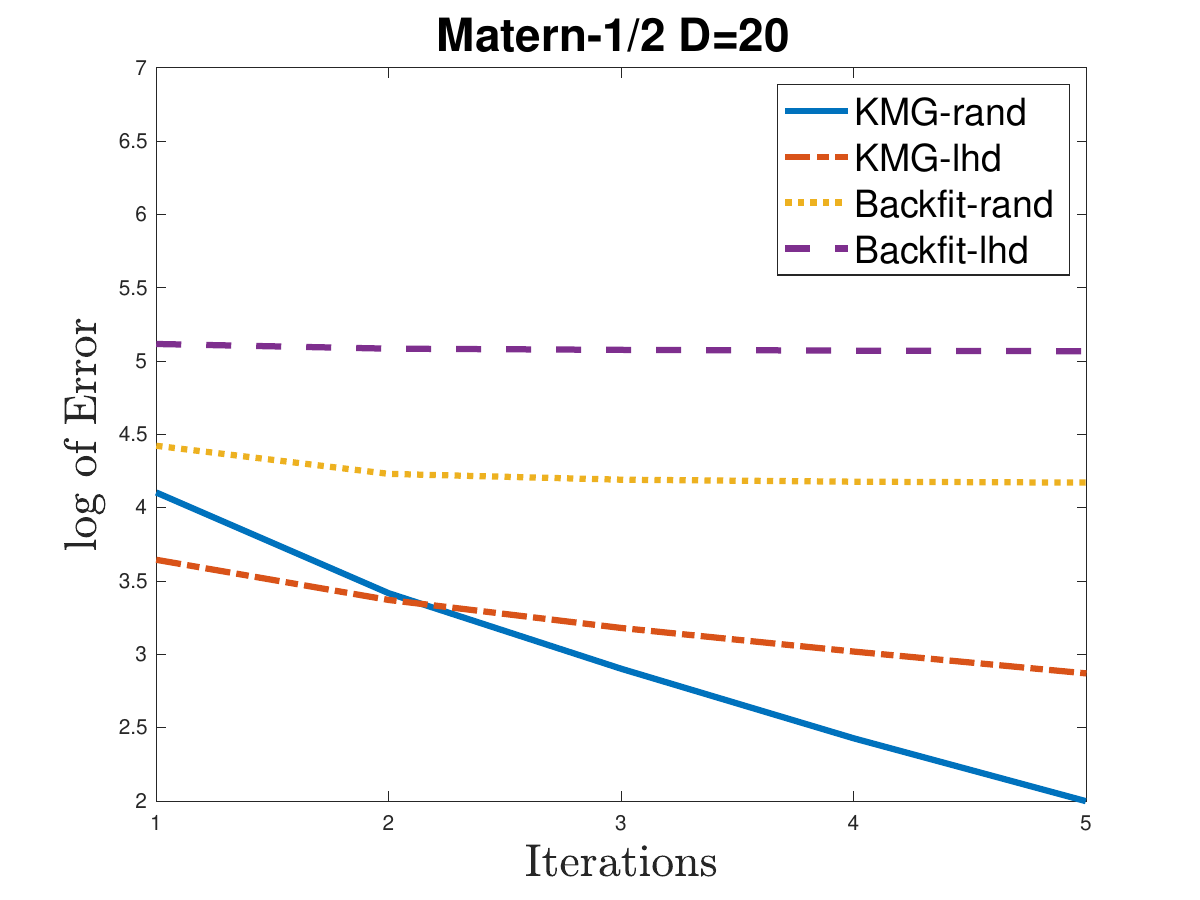}
    \includegraphics[width=.32\linewidth]{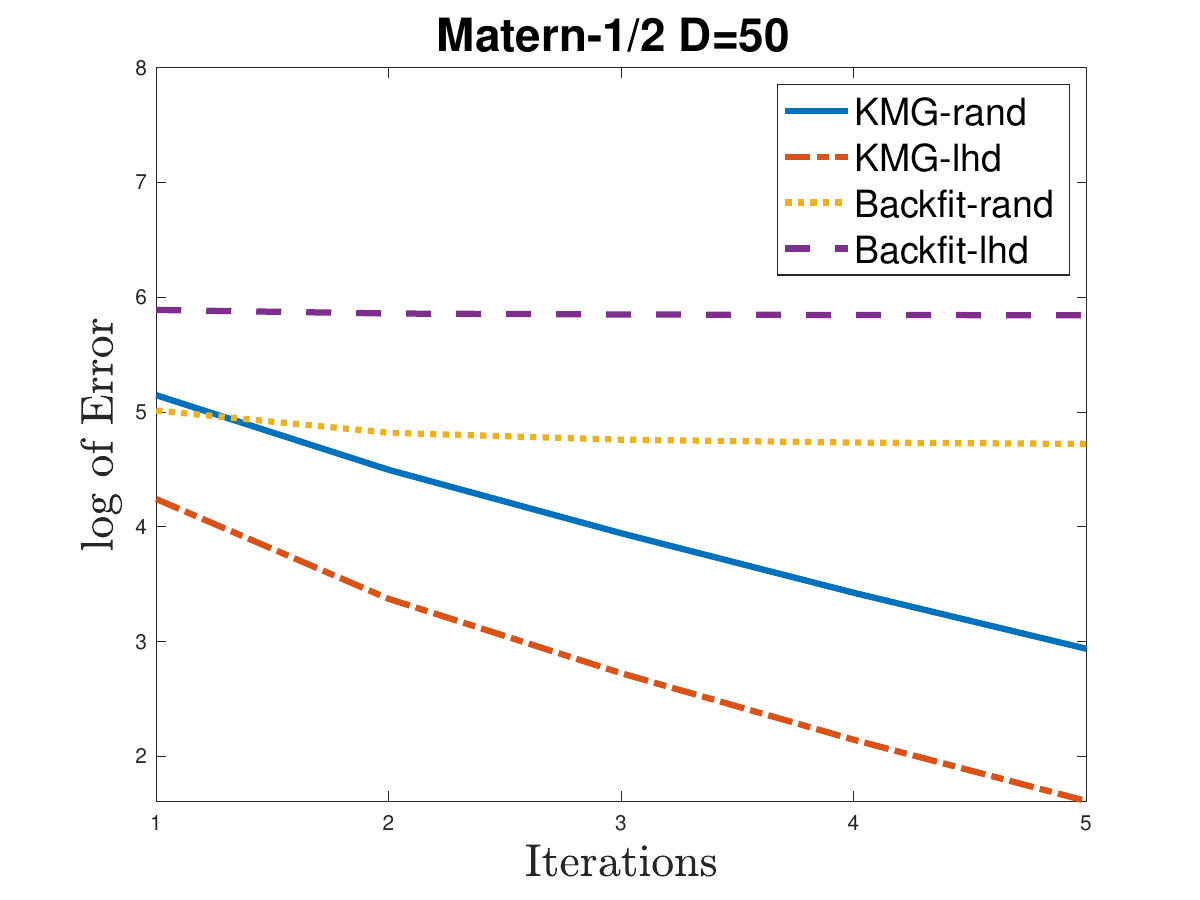}
     \includegraphics[width=.32\linewidth]{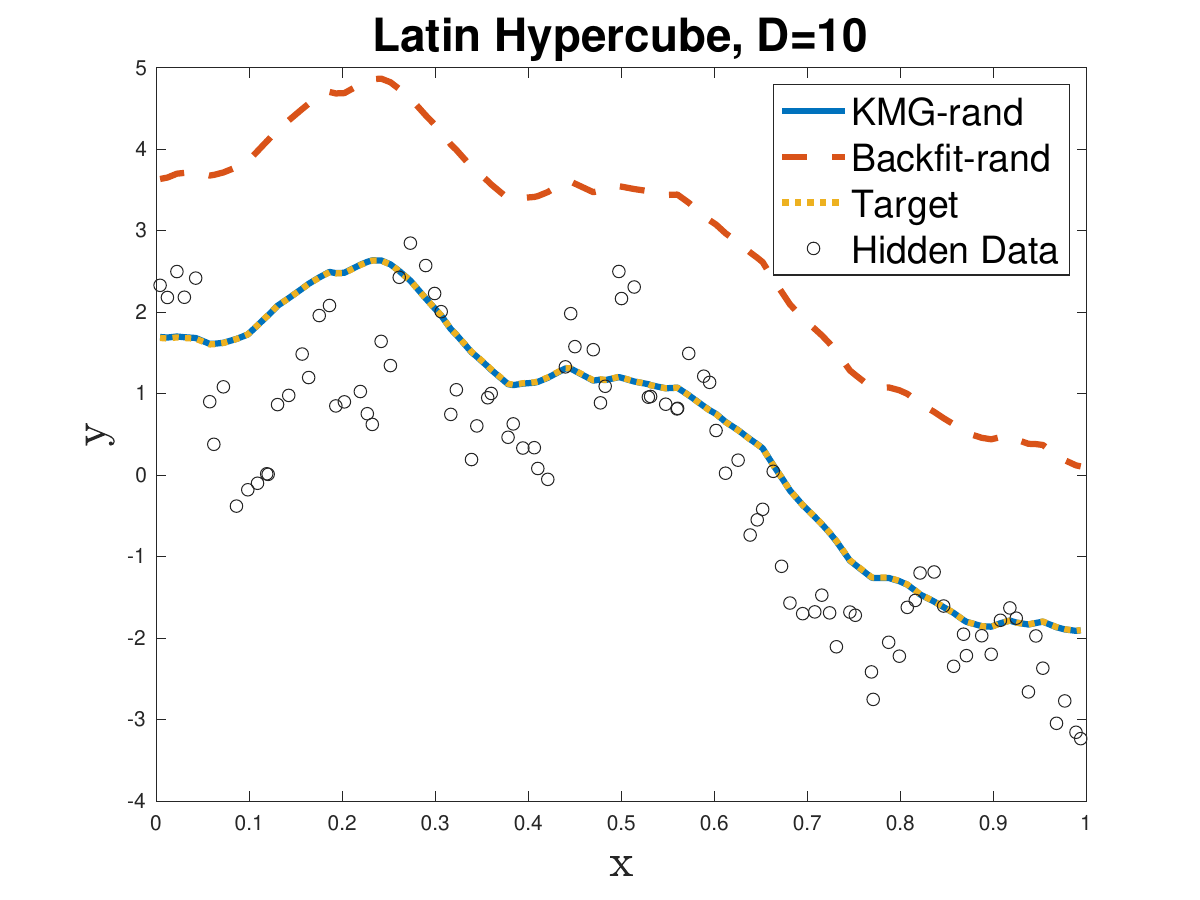}
    \includegraphics[width=.32\linewidth]{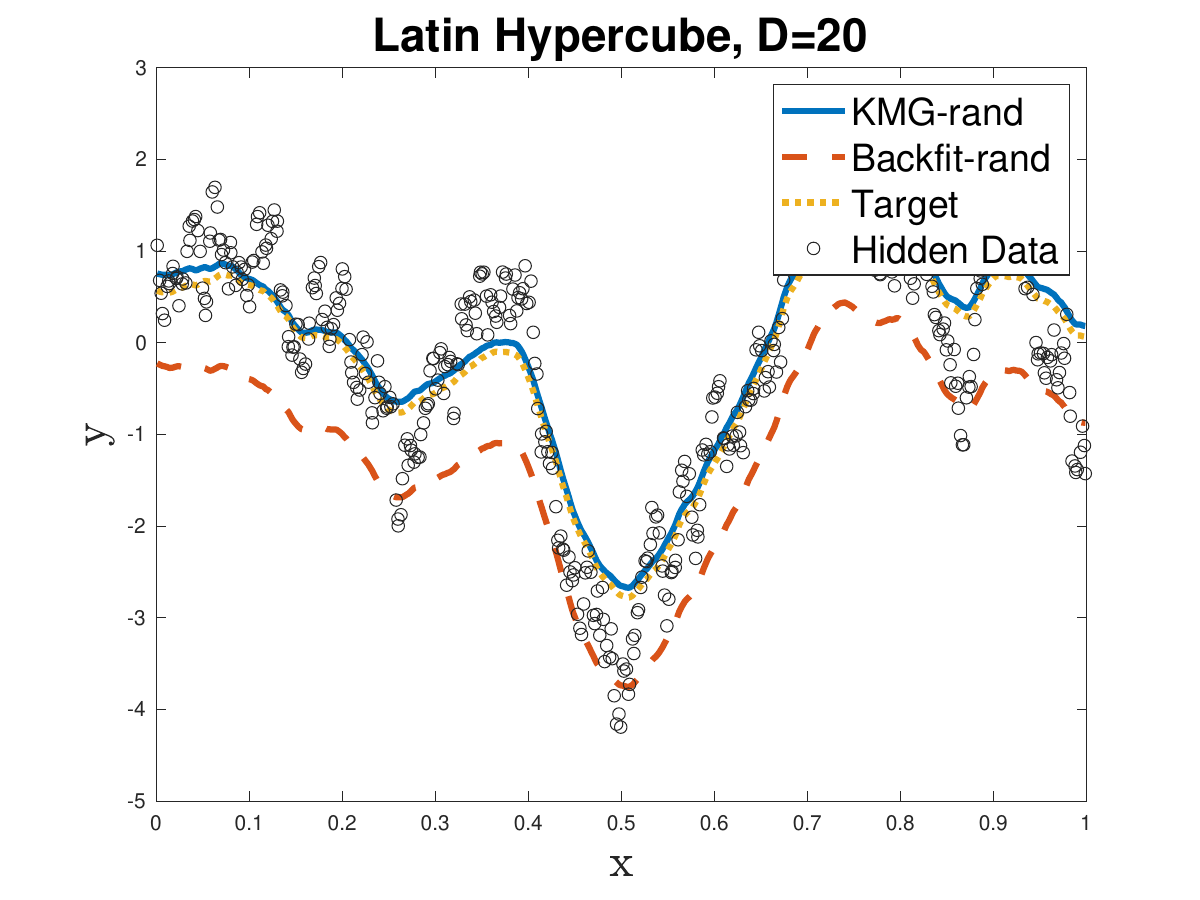}
    \includegraphics[width=.32\linewidth]{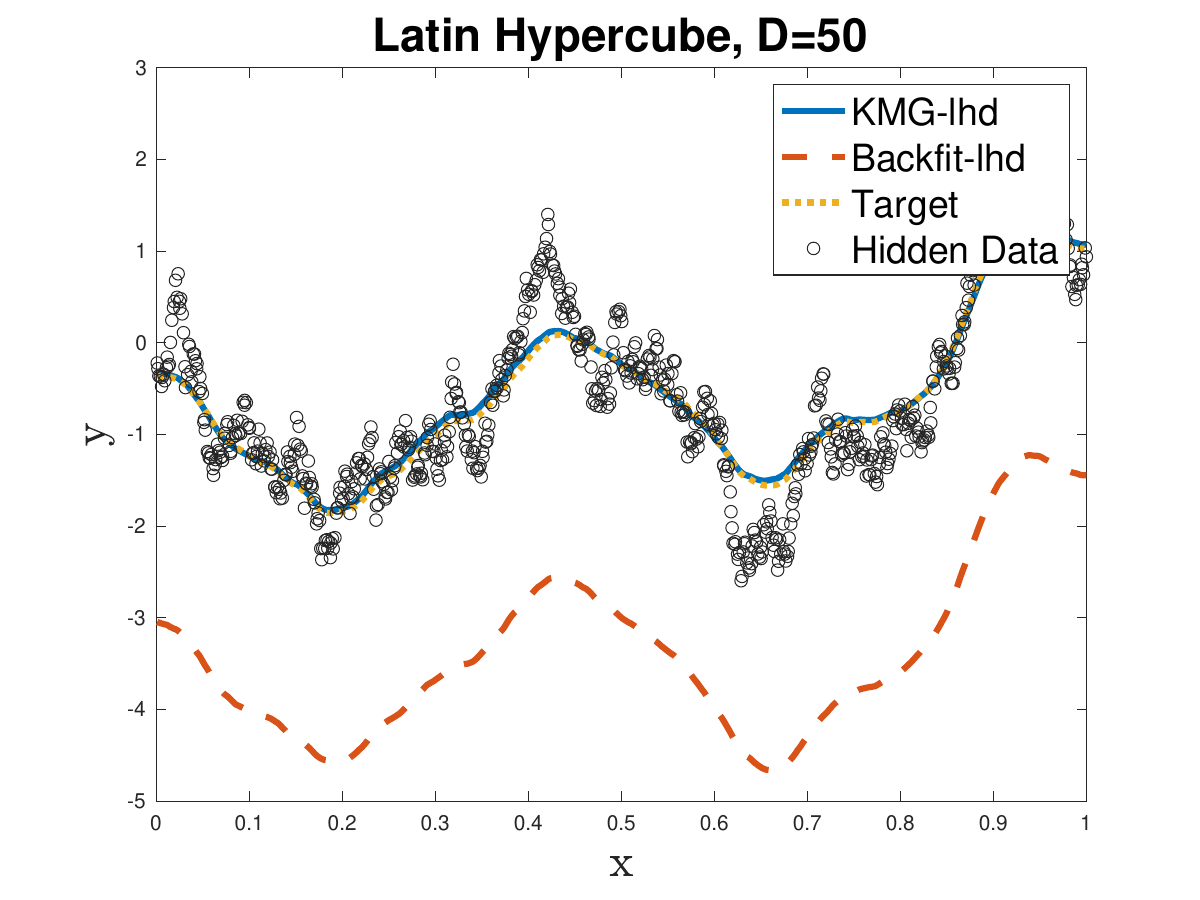}
    \includegraphics[width=.32\linewidth]{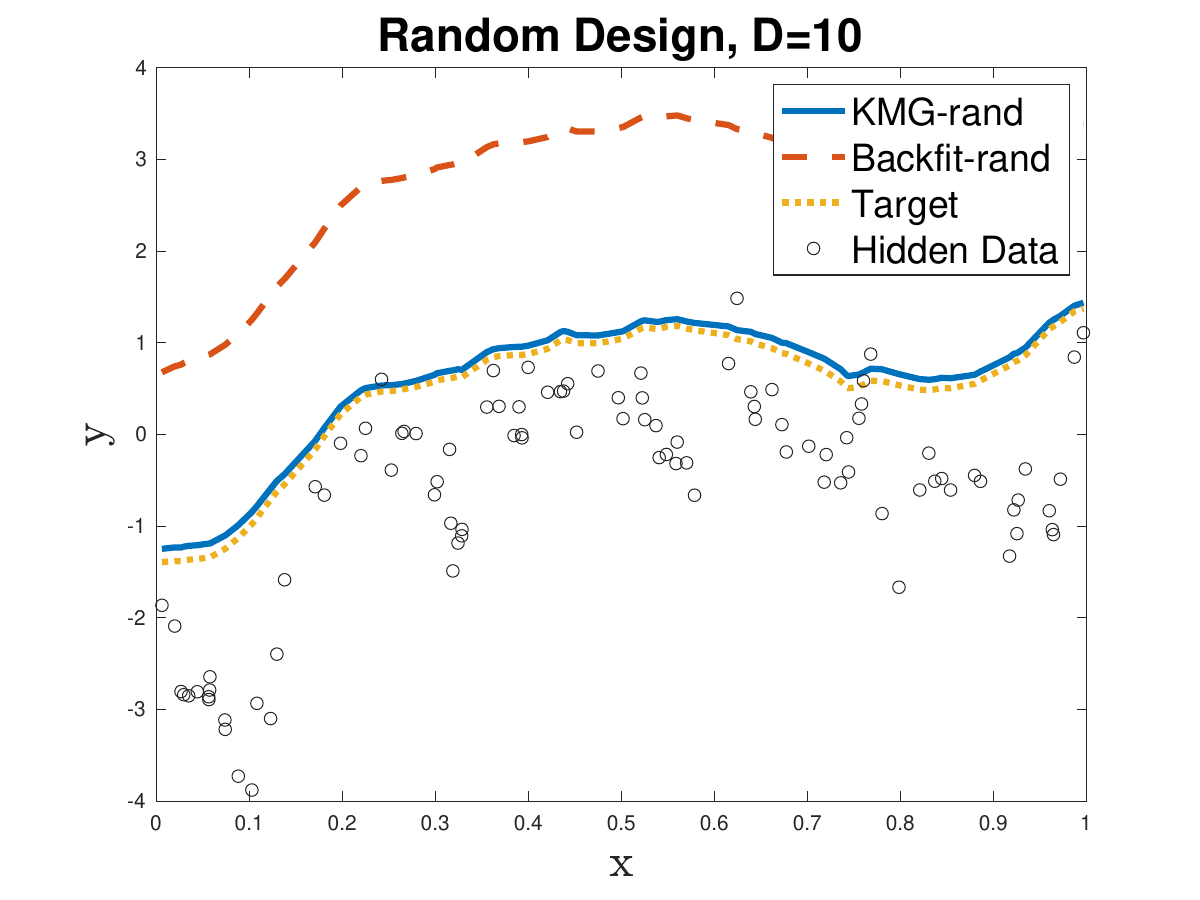}
    \includegraphics[width=.32\linewidth]{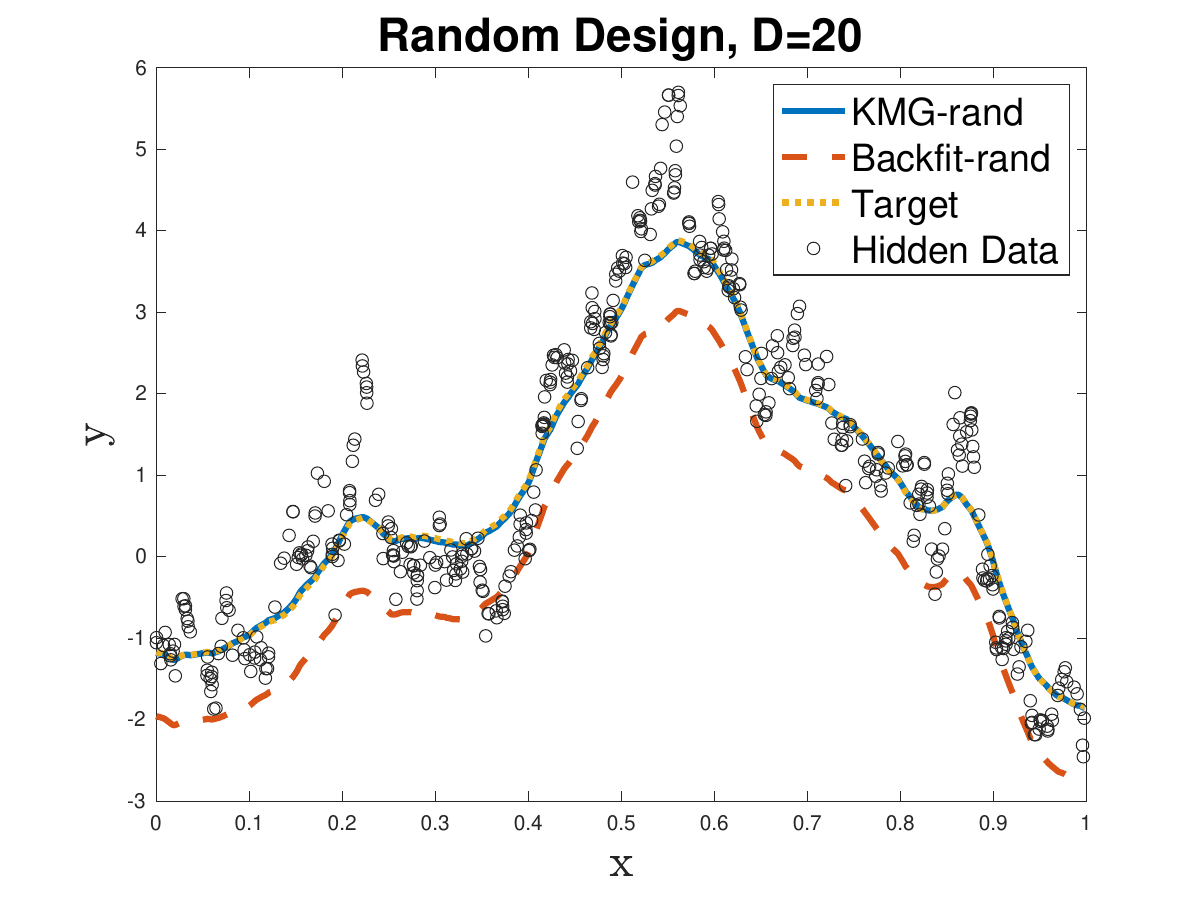}
    \includegraphics[width=.32\linewidth]{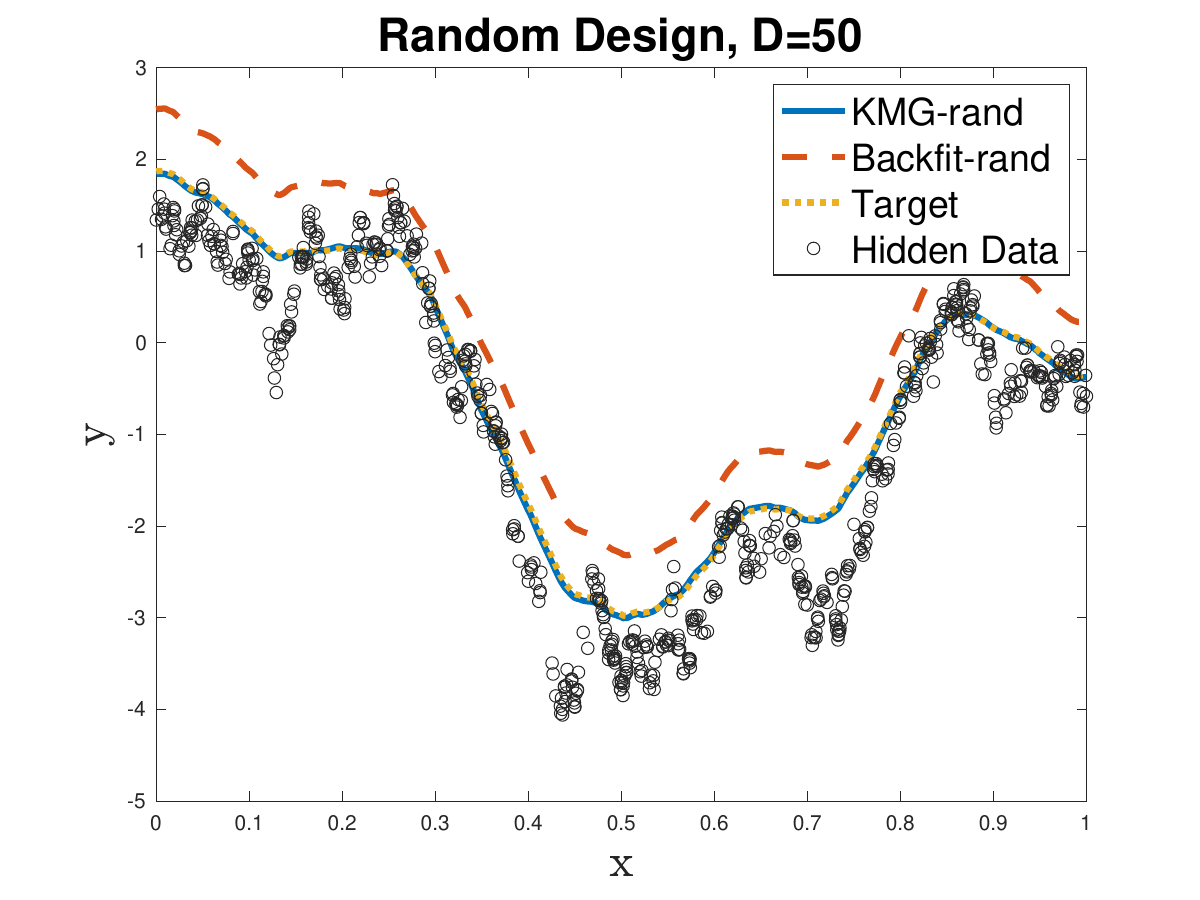}
    \caption{Experiments with Mat\'ern-${1}/{2}$. Upper row: logarithm of the error for the four competing algorithms.. Middle row: the resulting prediction curves for KMG and Back-fitting compared to the target function and the underlying hidden function $\CalG_d$, when $\BFX_n$ is  from a LHD. Lower row: the resulting prediction curves for KMG and Back-fitting compared to the target function and the underlying hidden function $\CalG_d$, when $\BFX_n$ is from a random design. }
    \label{fig:synthetic-Mat1/2}

\end{figure}

\begin{figure}[ht]
    \centering
    \includegraphics[width=.32\linewidth]{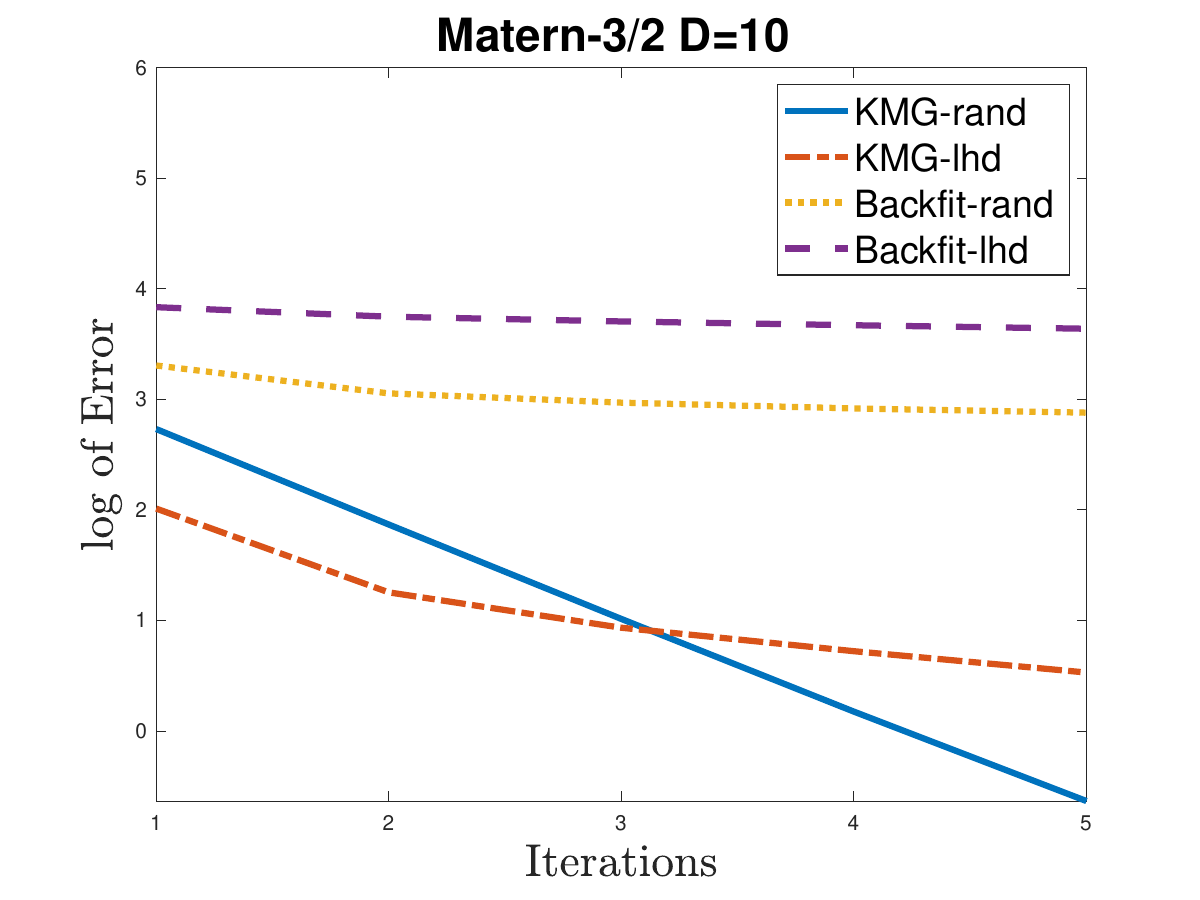}
   \includegraphics[width=.32\linewidth]{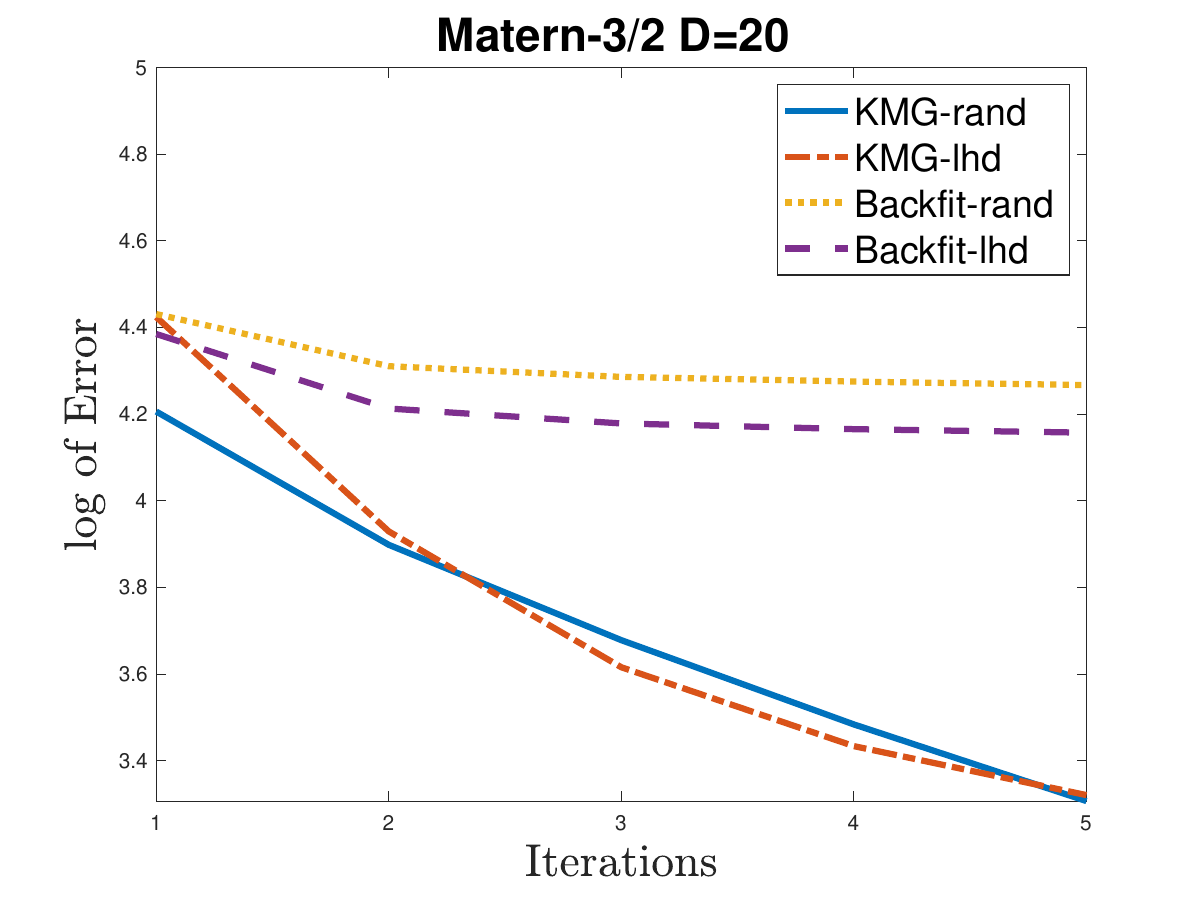}
    \includegraphics[width=.32\linewidth]{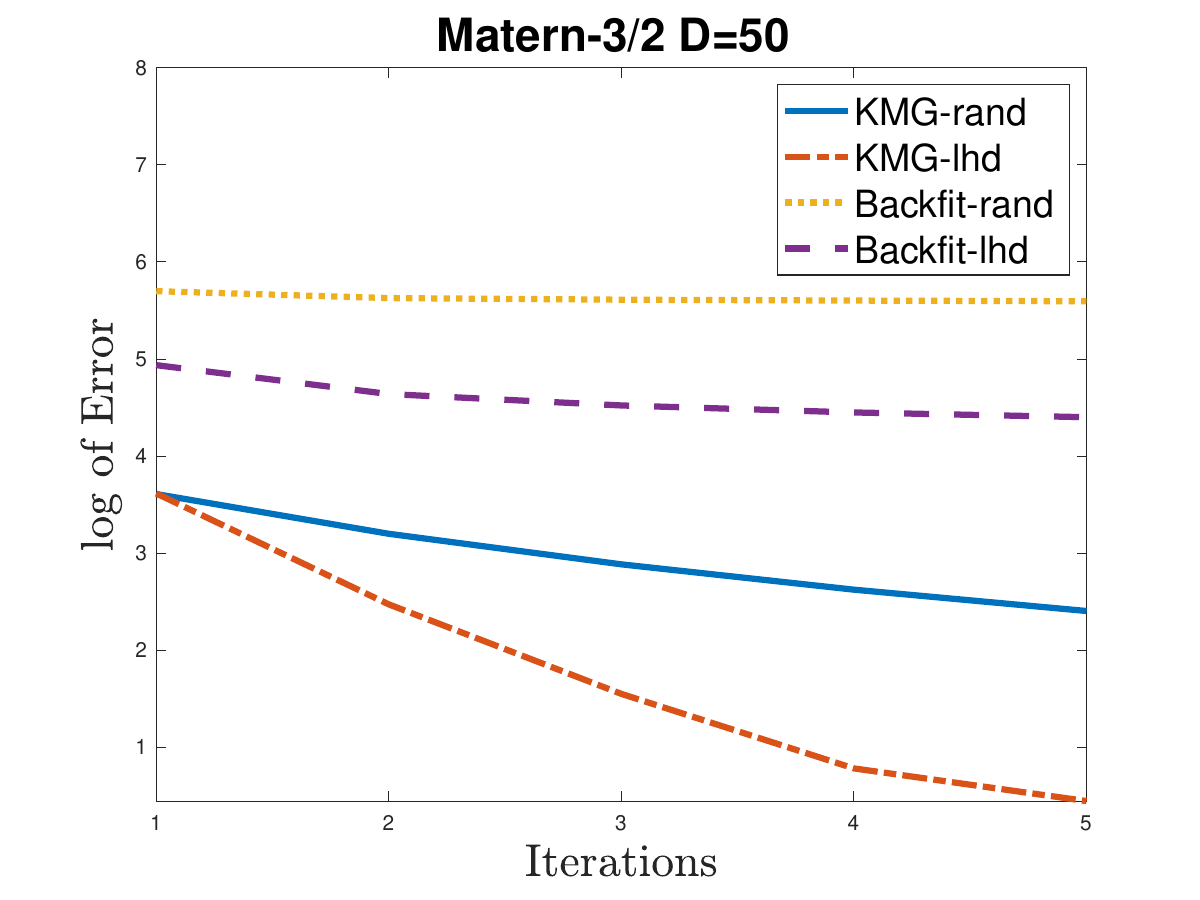}
     \includegraphics[width=.32\linewidth]{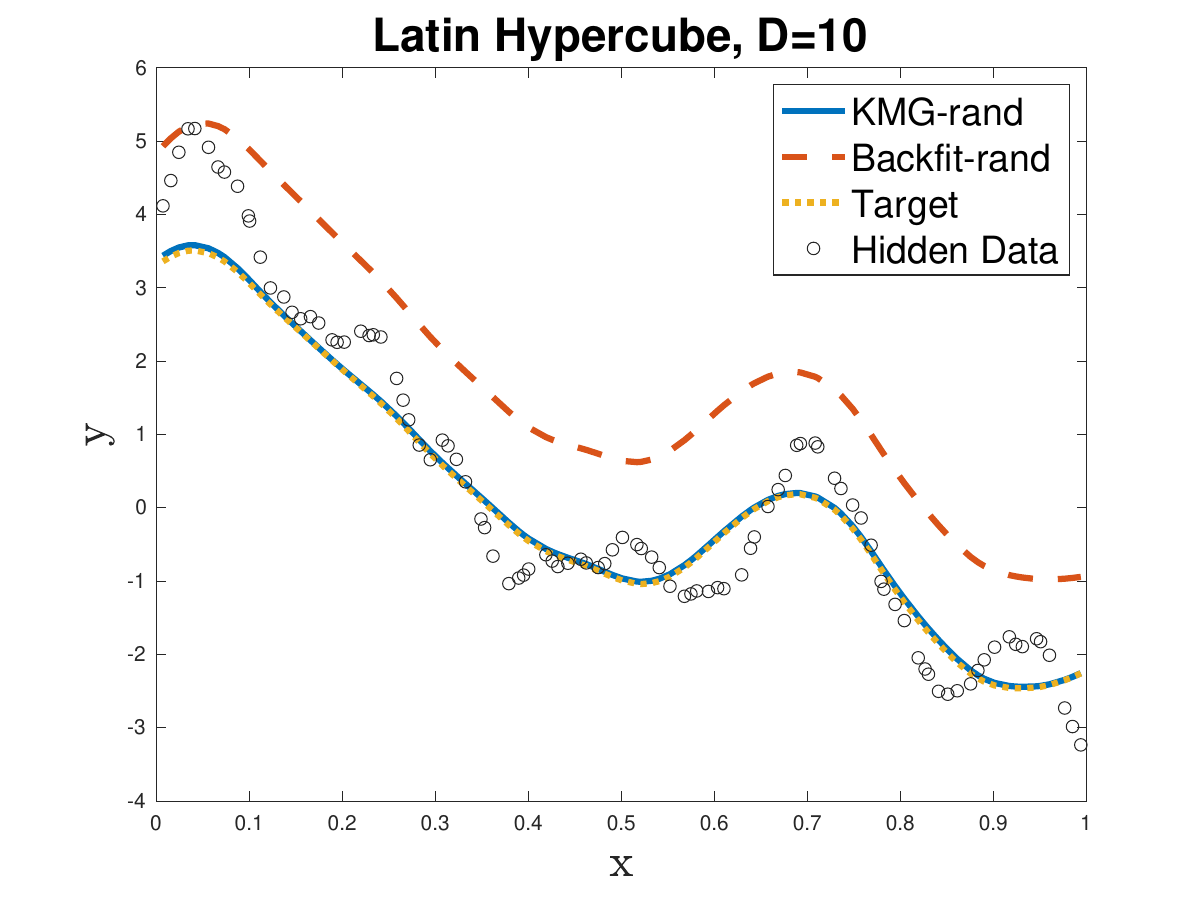}
    \includegraphics[width=.32\linewidth]{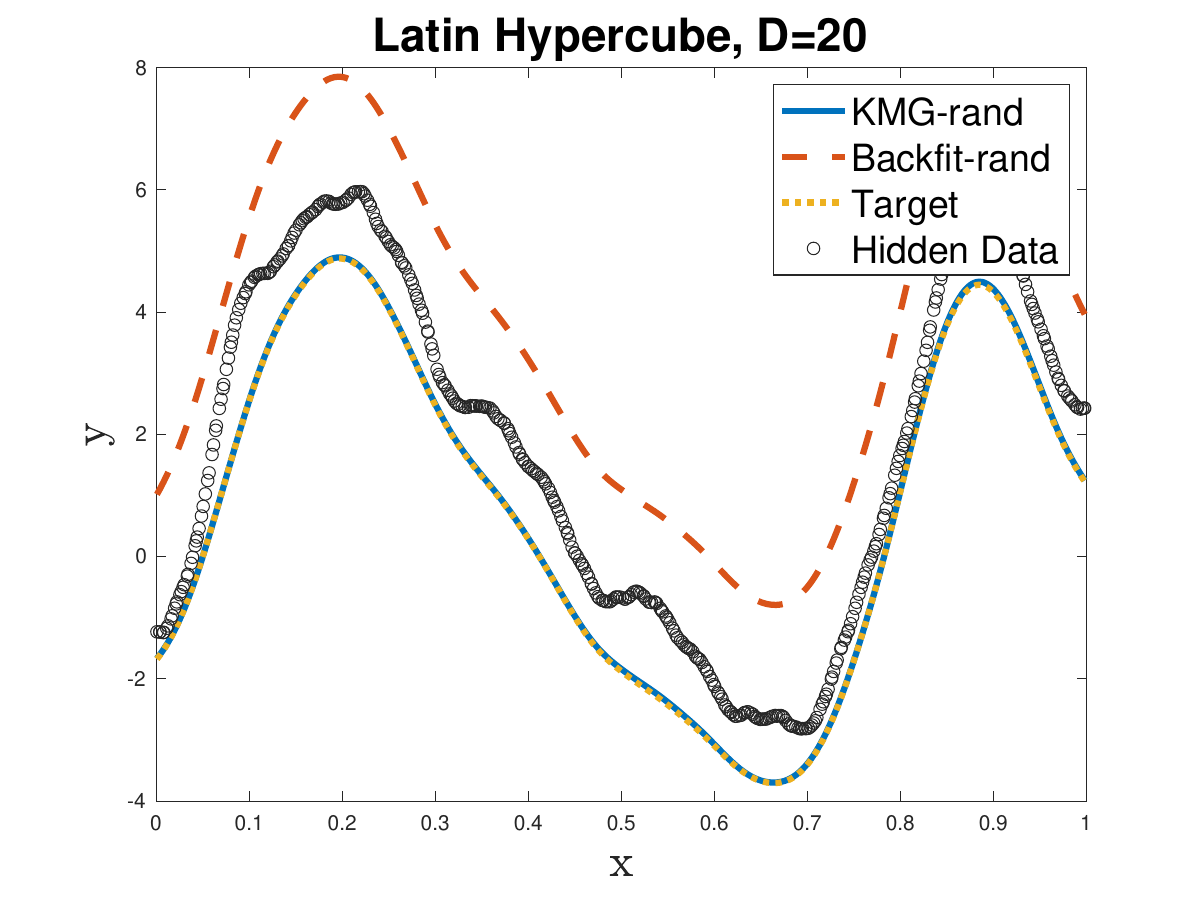}
    \includegraphics[width=.32\linewidth]{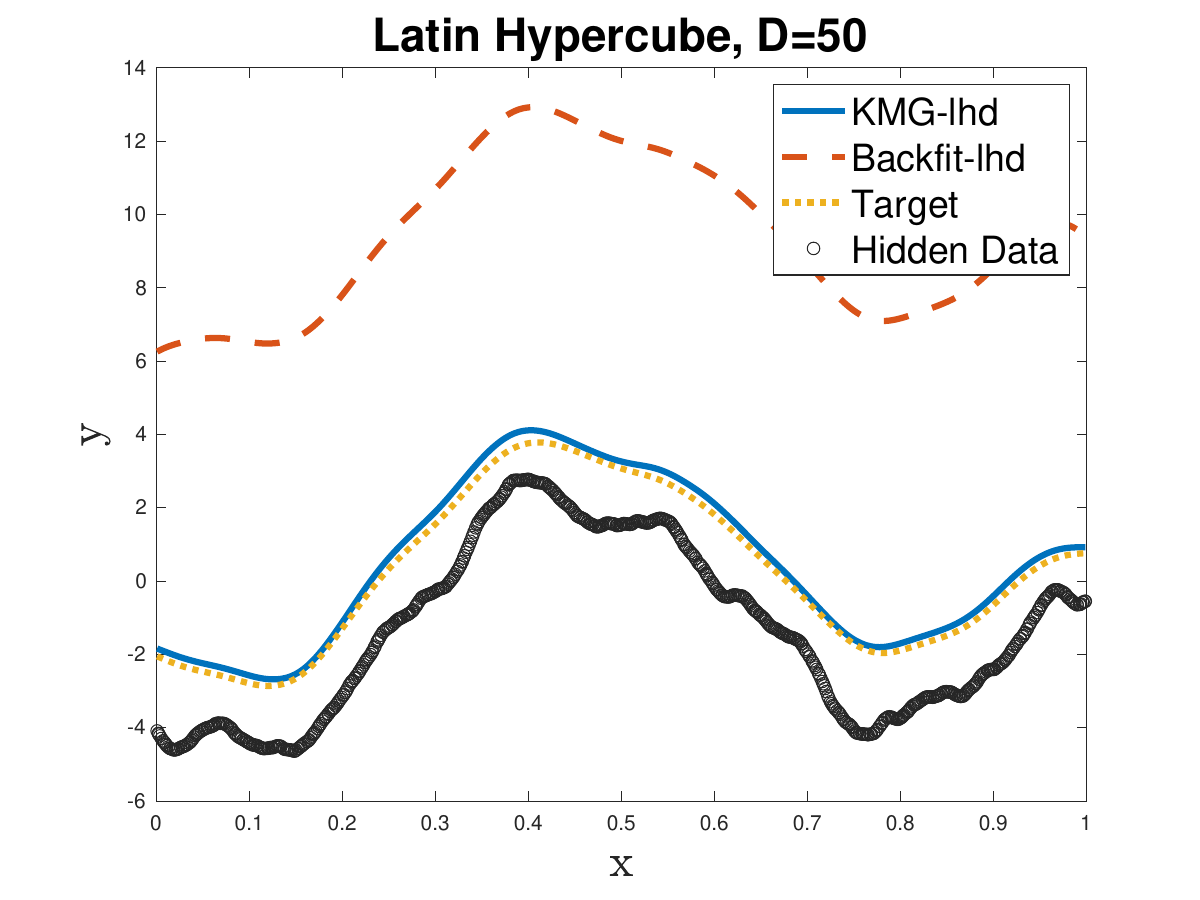}
    \includegraphics[width=.32\linewidth]{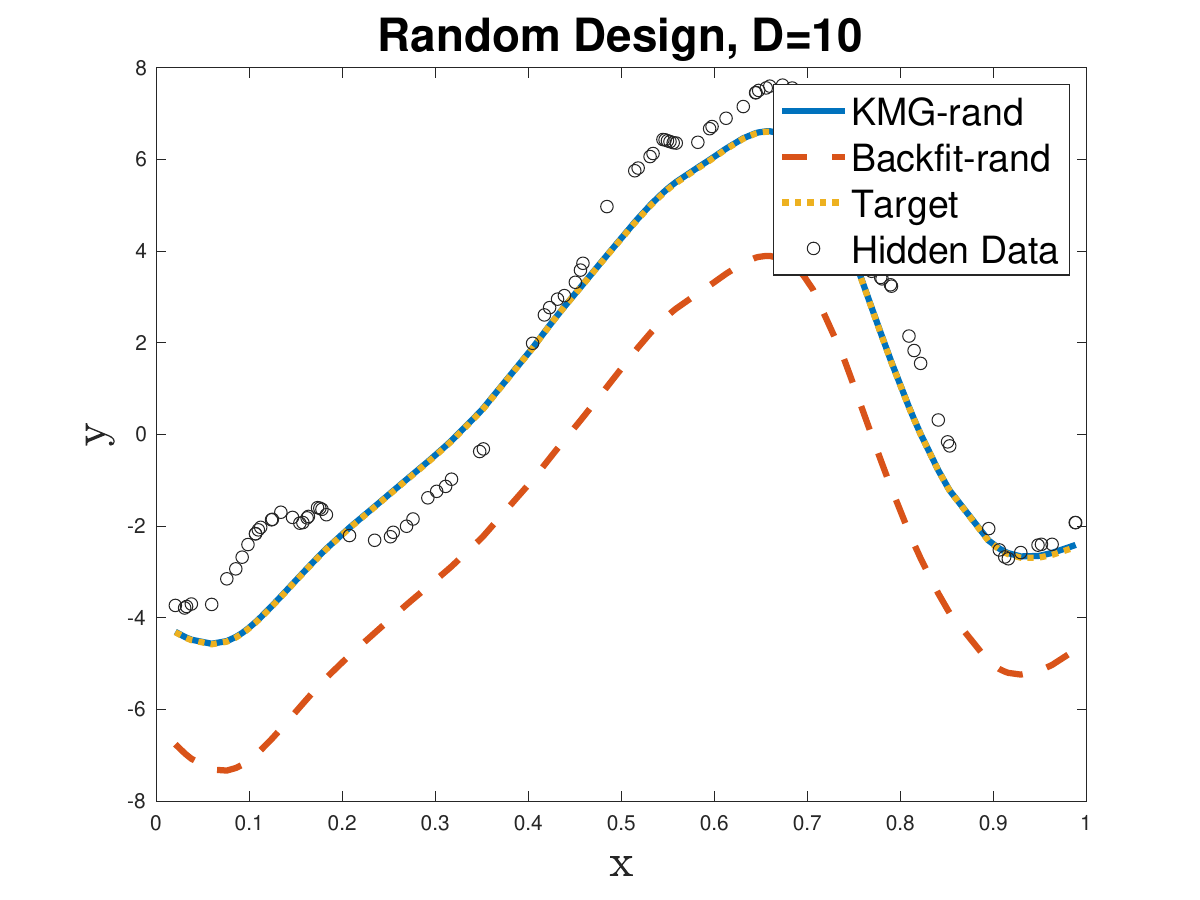}
    \includegraphics[width=.32\linewidth]{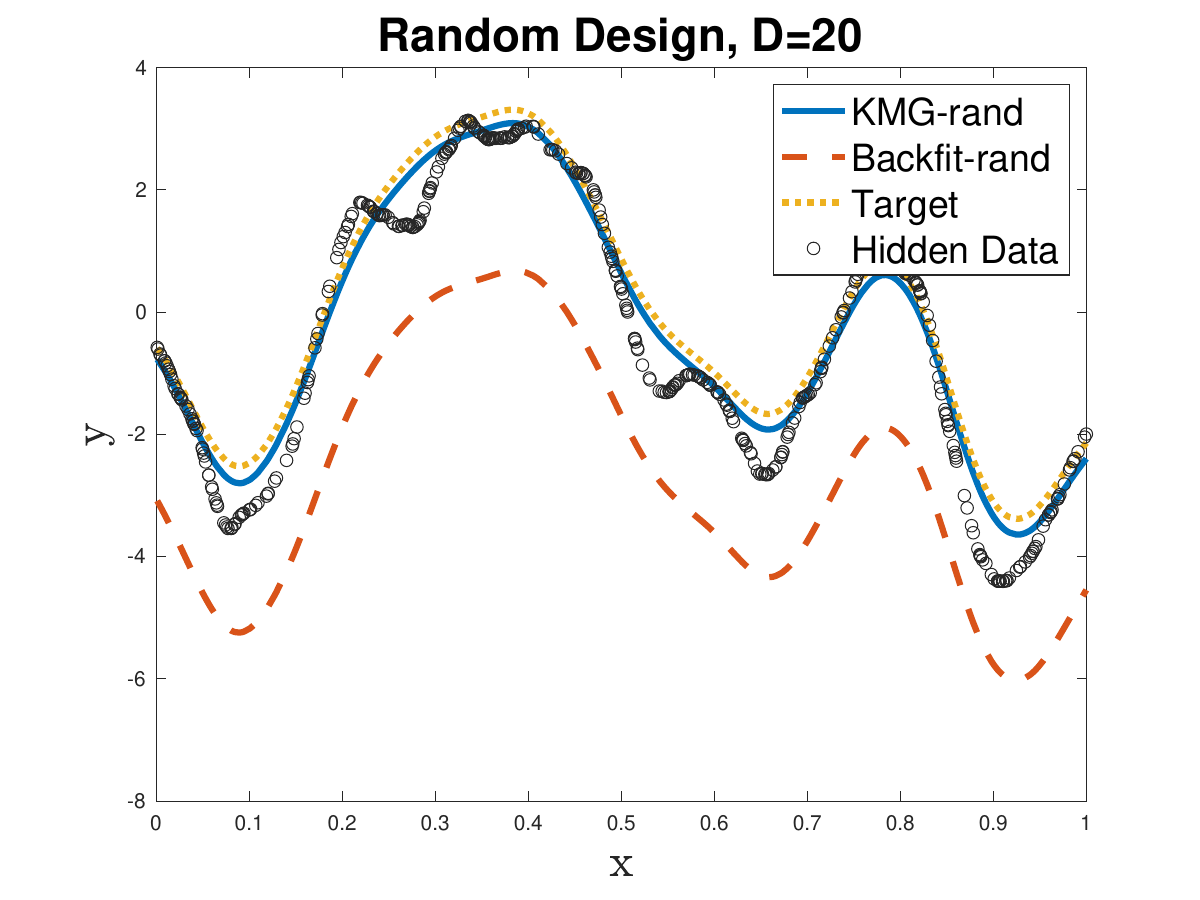}
    \includegraphics[width=.32\linewidth]{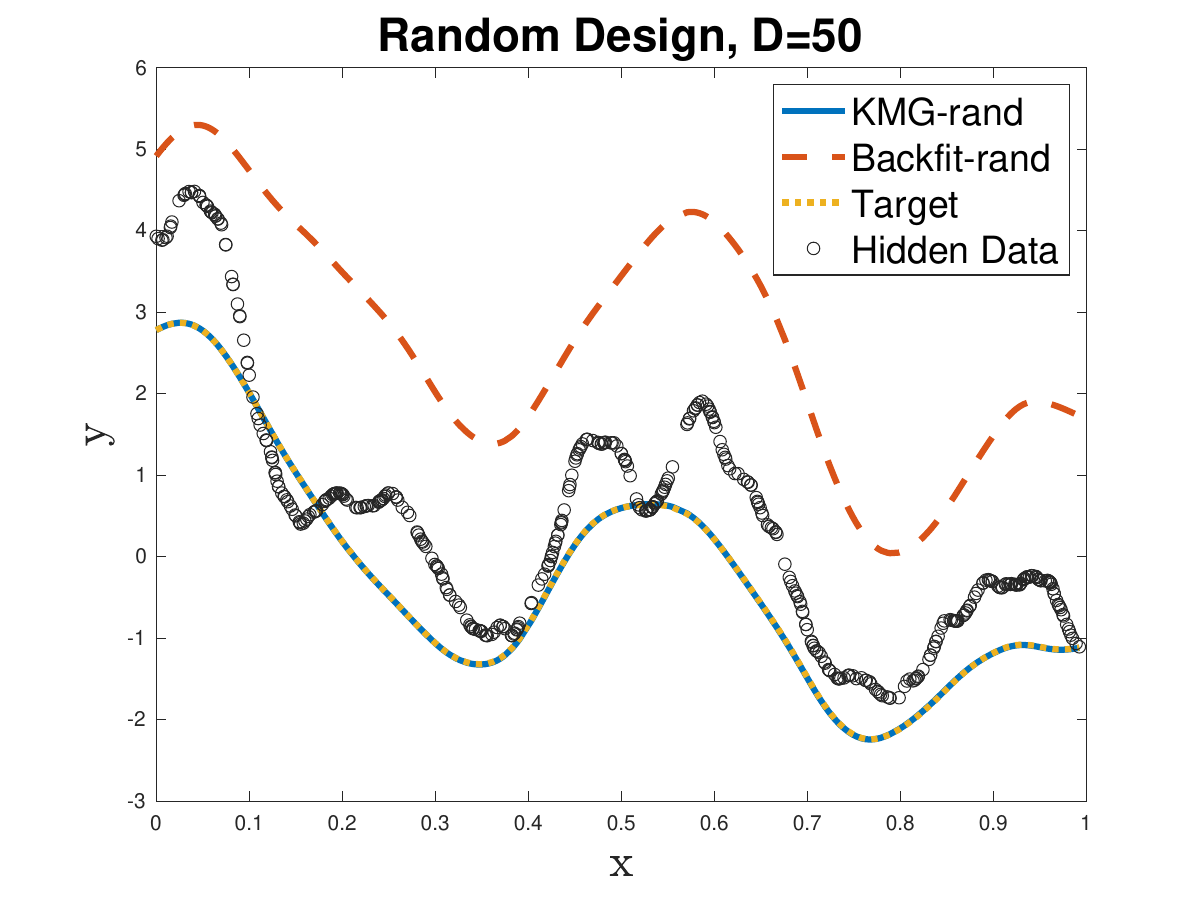}
    \caption{Experiments with Mat\'ern-${3}/{2}$. Upper row: logarithm of the error for the four competing algorithms. Middle row: the resulting prediction curves for KMG and Back-fitting compared to the target function and the underlying hidden function $\CalG_d$, when $\BFX_n$ is  from a LHD. Lower row: the resulting prediction curves for KMG and Back-fitting compared to the target function and the underlying hidden function $\CalG_d$, when $\BFX_n$ is from a random design.}
    \label{fig:synthetic-Mat3/2}
\end{figure}

We conduct experiments using the Mat\'ern-${1}/{2}$ kernel and the Mat\'ern-${3}/{2}$ kernel, limiting the number of iterations to $T=5$ before halting all competing algorithms. We evaluate performance by comparing $\sqrt{\sum_{d=1}^D\|\hat{f}_d(\BFX_d)-f^*(\BFX_d)\|_2^2}$, the $l^2$ vector norm between the outputs of the competing algorithms $\hat{f}_d$ and the target functions $f^*_d$. The results are presented in Figure \ref{fig:synthetic-Mat1/2} for the Mat\'ern-$1/2$ kernel and in Figure \ref{fig:synthetic-Mat3/2} for the Mat\'ern-${3}/{2}$ kernel, respectively.

From the upper rows of Figures \ref{fig:synthetic-Mat1/2} and \ref{fig:synthetic-Mat3/2}, it is evident that, even with as few as $m=10$ inducing points, the performance enhancement provided by KMG is significant. Merely 5 iterations suffice for KMG to converge, resulting in an error rate substantially lower than those observed with Back-fitting algorithms.

The middle and lower rows of Figures \ref{fig:synthetic-Mat1/2} and \ref{fig:synthetic-Mat3/2} underscore the limitations of Back-fitting in converging to the target function. As established in Section \ref{sec:lower_bound} and demonstrated through experiments in the preceding subsection, Back-fitting struggles with capturing global features, particularly the true magnitude of the target function. This deficiency is highlighted by the substantial margin of separation between the target function approximations generated by Back-fitting and the actual target function. In contrast, KMG adeptly overcomes this challenge, effectively capturing global features and aligning closely with the true target function, thus demonstrating a superior ability to learn and approximate global characteristics.

To summarize, the experiments presented in this subsection illustrate KMG's superior efficiency in capturing global features that Back-fitting overlooks. Consequently, KMG significantly enhances the performance of Back-fitting, elevating its efficiency to a new level.

\subsection{Real Case Examples}

We evaluate the performance of KMG and Back-fitting using the \textit{Breast Cancer} dataset \citep{misc_breast_cancer_wisconsin_(diagnostic)_17} and the \textit{Wine Quality} dataset \citep{misc_wine_quality_186}. The Breast Cancer dataset includes 569 samples, each with $D=30$ features, aimed at facilitating breast cancer diagnosis decisions. The Wine Quality dataset comprises 4898 samples, each with $D=11$ features, with the objective of predicting wine quality based on these 11 features.

Our experiments are designed with two primary objectives. The first objective is to assess the accuracy of KMG and Back-fitting in approximating the target function as detailed in \eqref{eq:experiment_goal}. This objective is distinct from conventional classification or prediction tasks, as it focuses on reconstructing how each feature contributes to the outcome. The second objective involves comparing the performance of KMG and Back-fitting in practical applications: we evaluate their classification error rate on the Breast Cancer dataset and their mean squared error (MSE) on the Wine Quality dataset. The classification error rate and MSE are defined as follows:
\begin{align*}
    &\text{Error rate}=\frac{\#(\text{incorrect classifications})}{\#(\text{test cases})},\\
    &\text{MSE}=\frac{1}{n_{\rm test}}\sum_{i=1}^{n_{\rm test}}|\hat{f}(\BFx_i)-y_i|^2.
\end{align*}

Details of all competing algorithms are outlined as follows:

\begin{enumerate}
    \item \textbf{KMG-Mat-1/2}: Kernel Multigrid for additive Mat\'en-1/2 kernel with $m=10$ inducing points $\BFU_m$; 
    \item \textbf{KMG-Mat-3/2}: Kernel Multigrid for additive Mat\'en-3/2 kernel with $m=10$ inducing points $\BFU_m$; 
   \item \textbf{Backfit-Mat-1/2}: Back-fitting for additive Mat\'en-1/2 kernel; 
   \item \textbf{Backfit-Mat-3/2}: Back-fitting for additive Mat\'en-3/2 kernel.
\end{enumerate}

\begin{figure}[ht]
    \centering
    \includegraphics[width=.42\linewidth]{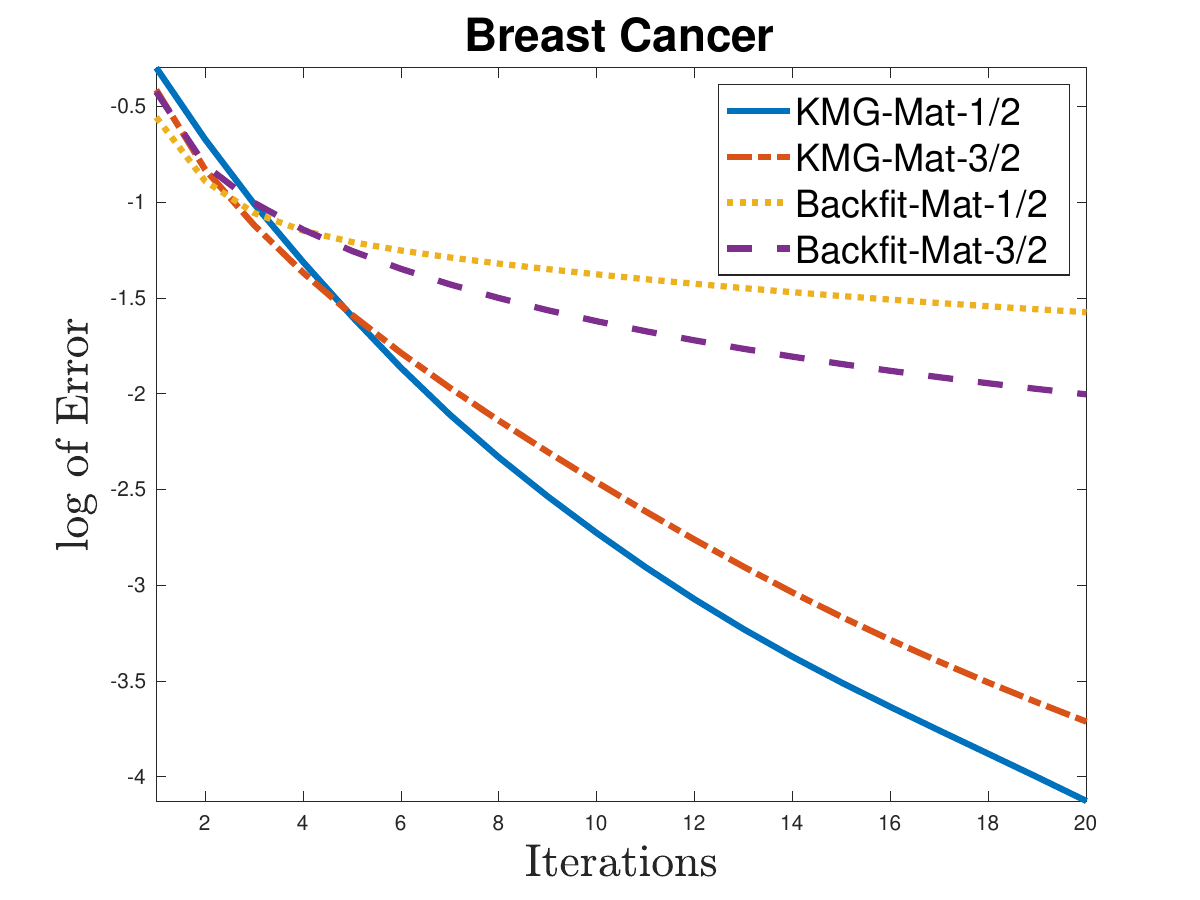}
   \includegraphics[width=.42\linewidth]{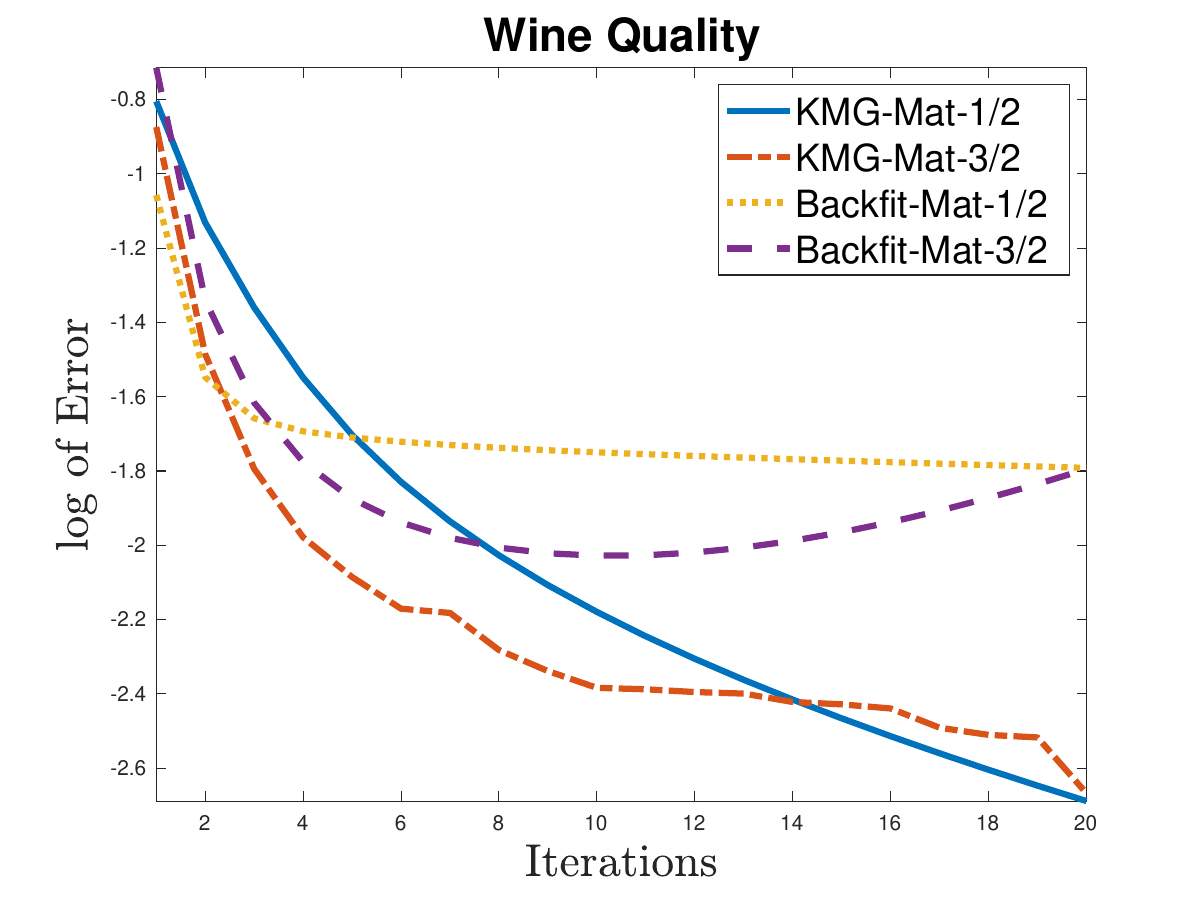}
   \caption{Left: Breast Cancer dataset; right: Wine Quality dataset.}
    \label{fig:real_data}

\end{figure}

To assess the performance of KMG and Back-fitting in reconstructing the target function outlined in \eqref{eq:experiment_goal}, we utilized all 569 samples from the Breast Cancer dataset and 2000 samples from the Wine Quality dataset. Initially, we computed the target function \eqref{eq:experiment_goal} utilizing both the Matérn-$1/2$ and Matérn-$3/2$ kernels. Subsequently, we executed both KMG and Back-fitting algorithms to approximate the components $f^*_d$ specified in \eqref{eq:experiment_goal}. The algorithms were allowed a maximum of 20 iterations. The outcomes of these experiments are shown in Figure \ref{fig:real_data}.

For the Breast Cancer classification, we randomly selected 500 samples for the training set and used the remaining 69 samples for the test set. Similarly, for Wine Quality prediction, we randomly chose 2000 samples as our training set and designated the subsequent 1000 samples for the test set. We then constructed additive predictors by summing the outputs from Back-fitting and KMG after 5, 10, and 20 iterations, respectively. Specifically, the predictors were formulated as
\[\hat{f}=\sum_{d=1}^D\hat{f}_d\]
where each $\hat{f}_d$ represents the estimators of $f^*_d$ obtained through Back-fitting or KMG. These predictors were then used for classification or prediction on the test sets. Each classification and prediction experiment was conducted 100 times to calculate the average error rate and mean squared error (MSE). The outcomes of  experiments for Breast Cancer are shown in Table \ref{tab:Breast_cancer} and those for Wine Quality are shown in Table \ref{tab:Wine_quality}.

Across all real-data experiments, it is evident that KMG significantly outperforms Back-fitting in approximating the target ${f}^*_d$, which denotes the contribution of each dimension to the overall data, across all datasets. These findings are consistent with those from the synthetic data experiments. Nevertheless, when employing the estimators derived from KMG and Back-fitting for classification or prediction tasks, the superiority of KMG over Back-fitting, though present, is not as pronounced as it is in estimating the contributions $f^*_d$ from each dimension. Estimating the overall model $f^* = \sum_{d=1}^D f^*_d$ and determining the contributions $f^*_d$ from each individual dimension are fundamentally different objectives. In estimating the overall model $f^*$, the precise distribution of global features across various dimensions is less crucial. Even if incorrect allocations occur within the Back-fitting algorithm, as demonstrated in Section \ref{sec:lower_bound}, the aggregate nature of the summation can neutralize errors from such misallocations. However, accurately identifying the contributions $f^*_d$ demands exact allocation of global features, where any misallocation can result in significant errors. 

\begin{table}[ht]
    
\centering
\begin{tabular}{ |p{3cm}||p{3cm}|p{3cm}|p{3cm}|  }
 \hline
 Iteration number & 5 & 10 &20\\
 \hline
 KMG-Mat-1/2   & 0.0922    &0.0699&   0.0689\\
  KMG-Mat-3/2&   0.0837  & 0.0664   &0.0652\\
 Backfit-Mat-1/2 &0.1266 & 0.1134&  0.1004\\
 Backfit-Mat-3/2    &0.1095 & 0.0932&  0.0876\\
 \hline
\end{tabular}
\caption{Classification error rate  of Breast Cancer}
\label{tab:Breast_cancer}
\end{table}

\begin{table}[ht]
    
\centering
\begin{tabular}{ |p{3cm}||p{3cm}|p{3cm}|p{3cm}|  }

 \hline
 Iteration number & 5 & 10 &20\\
 \hline
  KMG-Mat-1/2&   0.4327  & 0.3521   &0.3478\\
 KMG-Mat-3/2   & 0.3966    &0.3472&  0.3403\\
 Backfit-Mat-1/2 &0.4815 & 0.4331&  0.4122\\
 Backfit-Mat-3/2    &0.4387 & 0.4212&  0.4313\\
 \hline
\end{tabular}
\caption{Prediction MSE  of Wine Quality}
\label{tab:Wine_quality}
\end{table}

A simple example can demonstrate this phenomenon. Consider a model where \(f^* = f^*_1 + f^*_2\), with \(\|f^*_1\|_{L_2}^2 = 1\) and \(\|f^*_2\|_{L_2}^2 = 9\), leading to \(\|f^*\|_{L_2}^2 = 10\) if $f_1$ is orthogonal to $f_2$. When applying Back-fitting to this model, we have proved that it struggles to accurately discern the global features, which are \(L_2\) norms $\|f^*_d\|_{L_2}$ in this case. Essentially, Back-fitting can ascertain that the additive model has an \(L_2\) norm of \(\sqrt{10}\), but it faces challenges in deconstructing this aggregate norm back into the distinct contributions of each \(f^*_d\). This characteristic of Back-fitting hinders its ability to accurately reconstruct each individual \(f^*_d\), though it does not significantly impede the reconstruction of the overall model \(f^*\). Conversely, identifying the $L^2$ norm for each dimension can be effectively achieved with a modest number of inducing points, which serve to "isolate" the data by dimensions, followed by straightforward regression for precise estimations.

To summarize, our real-world experiments demonstrate that KMG notably excels over Back-fitting in reconstructing the individual contributions, \(f^*_d\), from each dimension. However, when it comes to reconstructing the overall model, the superiority of KMG is less pronounced.

\section{Conclusions and Discussion}\label{sec:Conclude}

In this study, we establish a lower bound for the convergence rate of Back-fitting, demonstrating that it struggles to effectively reconstruct global features, necessitating at least \(\mathcal{O}(n\log n)\) iterations to converge. To address this limitation, we propose an enhancement by applying a sparse GPR to the residuals generated by Back-fitting in each iteration. This approach enables the reconstruction of global features and significantly reduces the required number of iterations for convergence to \(\mathcal{O}(\log n)\), offering a more efficient  solution.

There are several directions that could be pursued in future research. Firstly, the KMG algorithm could be expanded into more advanced Multigrid techniques, such as F-cycles, W-cycles \citep{saad2003iterative}, or Multigrid Conjugate Gradients \citep{tatebe1993multigrid}. For complex real-world challenges, which often involve millions of data points across thousands of dimensions, these potential research directions could yield practical benefits. From a theoretical perspective, the KMG presented in our study, which employs only a small number of inducing points,  is sufficient to attain  the theoretically fastest exponential convergence rate, while maintaining time and space complexities at minimal order.

Secondly, our focus has been on training additive GPs to reconstruct contributions from each dimension. However, a variety of  kernel machines  also aim at achieving similar objectives, yet they operate with distinct kernel structure, loss functions, and update mechanisms. Extending our algorithms and analysis to enhance the training efficiency of these diverse models remains an unexplored area.

Thirdly, numerous iterative methods exist for training kernel machines, yet analyses focusing on the requisite number of iterations for these training algorithms are scarce.  We posit that there must exist simple modifications to many iterative kernel training methods that could significantly enhance their convergence rates. Similar to our study, it has been rigorously proved that training kernel ridge regression via gradient descent with data augmentation techniques as in \cite{ding2023random}, also requires $\mathcal{O}(\log n)$ iterations only. Thorough analysis regarding the required iteration number in  many other kernel machine training algorithms are needed, presenting an opportunity to broaden our analytical framework and methodology to improve these algorithms.

\bibliography{reference}





\newpage

\appendix
\section{Proof of Lower Bound}
In this section, we prove the the propositions and theorems in Section \ref{sec:lower_bound}. We will first prove Proposition \ref{prop:convolution_A} and Proposition \ref{prop:KP_estimation}, then Theorem \ref{thm:global_feature}. The final goal is to prove the lower bound in Theorem \ref{thm:back-fit_lower_bound}.

\proof[Proposition \ref{prop:convolution_A}]

Because the Fourier transform of Mat\'ern kernel satisfies
\[\CalF[k_{\nu}](\omega)\propto (1+\pi^2\omega^2)^{-\nu-1/2}=\prod_{j=1}^{\nu+1/2}(1+\pi^2\omega^2)^{-1}\]
so the Mat\'ern-$(\nu+1)$ is proportional to the convolution of the Mat\'ern-$(\nu)$ and Mat\'ern-$1/2$ kernels:
\begin{equation}
\label{eq:matern_convolution}
    k_{\nu+1}(x,x')=\frac{1}{z}\int k_{\nu}(x,s)k_{1/2}(s,x') d s
\end{equation}
where $z$ is the normalized  constant for Fourier transform and, without loss of generality, we can let $z=1$. 

\textbf{Central KP:} From Theorem 3 and Theorem 10 in \cite{chen2022kernel}, the constants $\{a_i\}$ for constructing KP are invariant under translation. For example, suppose $\{a_l\}_{l=1}^{2\nu+2}$ give the following central KP:
\[\phi_{0}=\sum_{l=1}^{2\nu+2}a_lk_{\nu}(\cdot,lh),\]
which means that the  $\phi_{\nu}$ is non-zero only on $(h,(2\nu+2)h)$. 

Recall that $\BFX^*=\{ih\}_{i=1}^n$
Then for any integer  $x^*_i=ih$, the following central KP
\[{\phi}_{i}=\sum_{i=1}^{2\nu+2}a_ik_{\nu}(\cdot,lh+ih)=\phi_0(\cdot-ih)\]
is non-zero only on $ih+(h,(2\nu+2)h)$. Therefore, the central KPs induced by all data points in $\BFX^*$ are translations of each other by multiples of $h$. To prove the proposition for central KPs, we only need to show that it holds for $\phi_0$ because 
\[[\BFA]_{i,(i-\nu-1/2):(i+\nu+1/2)}=[\BFA]_{j,(j-\nu-1/2):(j+\nu+1/2)}\]
for any $i,j=\nu+3/2,\cdots,n-\nu-1/2$ and half-integer $\nu$ due to the translation invariant property of KPs.

Let $A_{0}=[a_{1},\cdots,a_{2\nu+2},\ 0 \ 0]$, 
    $A_1=[0,\ a_{1},\cdots,a_{2\nu+2},\  0]$, 
    and $A_2=[0\ ,\ 0\ , a_{1},\cdots,a_{2\nu+2}]$.
So $\BFA=\beta_{0}A_{0}+\beta_1A_1+\beta_2A_2$ . If  $\{a_l\}_{l=1}^{2\nu+2}$ are the coefficients for constructing KP $\phi_0$ of Mat\'ern-$\nu$ kernel, we can construct KP $\psi$  of Mat\'ern-$(\nu+1)$ kernel by solving $\beta_i$ as follows:
\begin{align}
    \psi(x)&=\sum_{i=1}^{2\nu+4}\BFA_i k_{\nu+1}
    (x,ih)\nonumber\\
    &=\sum_{i=1}^{2\nu+4}\BFA_i \int_{\Real} k_{1/2}(x,s)
    k_{\nu}(s,ih)ds\nonumber\\
    &=\sum_{j=0}^{2}\beta_j \int_{\Real} e^{-\omega|x-s|}
    \phi_{0}(s-jh)ds\nonumber\\
    &=\sum_{j=0}^{2}\beta_j \int_{h+jh}^{(2\nu+2)h+jh} e^{-\omega|x-s|}
    \phi_{0}(s-jh)ds\label{eq:pf_convolution_A_1}
\end{align}
where the third line is from the translation invariance property of KP and the  definition of $\BFA$. Because $\psi$ is a KP of Mat\'ern $\nu+1$ kernel induced by points $\{ih\}_{i=1}^{2\nu+4}$, it should be non-zero only on $(h,(2\nu+4)h)$ according to definition. At any $x_-\leq h$ and $x_+\geq (2\nu+4)h$, we have
\begin{align}
   \psi(x_-)&= \sum_{j=0}^{2}\beta_j \int_{h+jh}^{(2\nu+2)h+jh} e^{\omega x_- -\omega s}
    \phi_{\nu}(s-jh)ds\nonumber\\
    &=\int_{h}^{(2\nu+2)h}e^{-\omega s}\phi_{\nu}(s)ds \sum_{j=0}^2\beta_j e^{\omega(h-jh)}=0 \label{eq:pf_prop3.3_KP_1}
    \end{align}
    \begin{align}
    \psi(x_+)&= \sum_{j=0}^{2}\beta_j \int_{h+jh}^{(2\nu+2)h+jh} e^{\omega s -\omega x_+}
    \phi_{\nu}(s-jh)ds\nonumber\\
    &=\int_{h}^{(2\nu+2)h}e^{\omega s}\phi_{\nu}(s)ds \sum_{j=0}^2\beta_j e^{\omega(jh-x_+)}=0\label{eq:pf_prop3.3_KP_2}.
\end{align}
We can see that the $\beta_j$ that satisfy \eqref{eq:pf_prop3.3_KP_1} and \eqref{eq:pf_prop3.3_KP_2} are
\begin{align*}
    &\sum_{j=0}^2\beta_j e^{\omega(jh)}=0, \quad\sum_{j=0}^2\beta_j e^{\omega(-jh)}=0
\end{align*}
and these are exactly the KP coefficients for Mat\'ern-$1/2$ kernel with three input points $0, h,2h$. According to the translation invariance property, $\beta_j$ are also the e KP coefficients for Mat\'ern-$1/2$ kernel with three input points $h,2h,3h$. This finishes the proof for central KPs.

\textbf{One-sided KP:}      Let $(a_1,\cdots,a_q)$ with $q=\nu+3/2$ be the coefficients for constructing left-sided Mat\'ern-$\nu$ KP  with input points $\{ih\}_{i=1}^{q}$, i.e.,
     \[\phi_0=\sum_{i=1}^{q}a_ik_\nu(\cdot,ih)\]
     is non-zero only on $(-\infty,qh)$. Let $(\beta_0,\beta_1)$  be the coefficients for constructing left-sided Mat\'ern-$1/2$ KP with input points $\{ih\}_{i=1}^{2}$. 
Here, we only prove the case for constructing left-sided Mat\'ern-$1/2$ KP. The  case for right-sided KPs can be proved by following the same reasoning.

Similar to central KP, let $A_{0}=[a_{1},\cdots,a_{\nu+3/2},\ 0 ]$ and  
    $A_1=[0,\ a_{1},\cdots,a_{\nu+3/2}]$, 
So $\BFA=\beta_{0}A_{0}+\beta_1A_1$  and  we need to solve for $\beta_0$ and $\beta_1$ for left KP $\psi$ as follows:
\begin{align}
    \psi(x)&=\sum_{i=1}^{q+1}\BFA_i k_{\nu+1}
    (x,ih)\nonumber\\
    &=\sum_{i=1}^{q+1}\BFA_i \int_{\Real} k_{1/2}(x,s)
    k_{\nu}(s,ih)ds\nonumber\\
    &=\sum_{j=0}^{1}\beta_j \int_{\Real} e^{-\omega|x-s|}
    \phi_{0}(s-jh)ds\nonumber\\
    &=\sum_{j=0}^{1}\beta_j \int_{-\infty }^{qh+jh} e^{-\omega|x-s|}
    \phi_{0}(s-jh)ds \label{eq:pf_convolution_A_2}
\end{align}
where $\phi_0$ now is a left KP for Mat\'ern-$\nu$ kernel, which is supported on $(-\infty,qh)$. If $\psi$ is a left KP for Mat\'ern-$(\nu+1)$ kernel, then for any point $x_+\geq qh+h$,  we should have
\begin{align}
     \psi(x_+)&=\sum_{j=0}^{1}\beta_j \int_{-\infty}^{qh+jh} e^{\omega s-\omega x_+}
    \phi_{0}(s-jh)ds \nonumber \\
    &=\int_{-\infty}^{qh}e^{\omega s}\phi_\nu(s)ds\  e^{-\omega x_+}\sum_{j=0}^1\beta_j e^{\omega jh}=0. \label{eq:pf_prop3.3_left_KP}
\end{align}
Similar to central KP, \eqref{eq:pf_prop3.3_left_KP} shows that $\beta_0$ and $\beta_1$ are the coefficients for Mat\'ern-$1/2$ left KP:
\[\sum_{j=0}^1\beta_j e^{\omega jh}=0.\]
 The statement for right KP can be proved in a symmetric way.
\endproof

\proof[ Proposition \ref{prop:convolution_Phi}:]
Our strategy is to first construct the following Mat\'ern $\nu+1$ KP factorization matrix $\BFA$ inductively by convolution as Proposition \ref{prop:convolution_A}:
  \begin{align*}
    &[\BFA_{\nu+1}]_{i,i:i+\nu+3/2}=\left[[\BFA_{\nu}]_{i,i:i+\nu+1/2}\ \ 0\right ]\BFA^*_{1,1}+\left[0\ \ [\BFA_{\nu}]_{i,i:i+\nu+1/2} \right]\BFA^*_{1,2},\ i\leq \nu+3/2\\
    &[\BFA_{\nu+1}]_{i,(i-\nu-3/2):i}=\left[[\BFA_{\nu}]_{i,(i-\nu-1/2):i}\ \ 0\right ]\BFA^*_{n,n-1}+\left[0\ \ [\BFA_{\nu}]_{i,(i-\nu-1/2):i} \right]\BFA^*_{n,n},\ i\geq n-\nu+1/2;
\end{align*}
and
\begin{align*}
    &[\BFA_{\nu+1}]_{i,(i-\nu-3/2):(i+(i+\nu+3/2)}=\left[[\BFA_{\nu}]_{i,(i-\nu-1/2):((i+\nu+1/2))}\ \ 0 \ \ 0 \right ]\BFA^*_{i,i-1}\nonumber \\
    &+\left[0\ \ [\BFA_{\nu}]_{i,(i-\nu-1/2):((i+\nu+1/2))} \ \ 0 \right]\BFA^*_{i,i}
    +\left[0\ \   0 \ \ [\BFA_{\nu}]_{i,(i-\nu-1/2):((i+\nu+1/2))}  \right]\BFA^*_{i,i+1}.
\end{align*}
for $i=\nu+5/2,\cdots,n-\nu-3/2$ with $\BFA^*$ the KP factorization matrix for Mat\'ern-$1/2$ kernel as stated in Proposition 1 in \cite{ding2022sample}:
\begin{align*}
    &\BFA^*_{i,i}=\left\{\begin{array}{lc}
            \frac{e^{\omega h}}{2\sinh(\omega h)} & \text{if} \quad i=1,n \\
             \frac{\sinh(2\omega h)}{2\sinh(\omega h)^2} & \text{otherwise}
        \end{array}\right. \ \ \BFA^*_{i,i+1}=\BFA^*_{i,i-1}=\frac{-1}{2\sinh(\omega h)}.
\end{align*}

Now we compute the entries on the Mat\'ern$\nu$ KP matrix $[\phi_i(x^*_j)]_{i,j}$ with $\nu>1/2$.  For the central KP of Mat\'ern-$1/2$ kernel, we can have the following KP induced by points $\{ih\}_{i=1}^3$ through direct calculation:
\begin{equation}
    \begin{aligned}
    \phi_0(x) =\left\{\begin{array}{lc}
           \frac{\sinh(\omega (x-h)}{\sinh(\omega h))}>0\  &\text{if} \quad x\in(h,2h)\\
             \frac{\sinh(\omega(3h-x))}{\sinh(\omega h)}>0 \  &\text{if}\quad x\in(2h,3h)\\
             0 \ &\text{otherwise}
        \end{array}\right.
\end{aligned}
\label{eq:central_KP_1-2}
\end{equation}
We now show that central KPs $\psi$ associated to matrix $\BFA$ in Proposition \ref{prop:convolution_A} is the $(\nu+1/2)$-time convolution of $\phi$:
\[\underbrace{\phi_0*\phi_0*\cdots*\phi_0}_{\nu+1/2\ \phi_0}(x).\]
This can be proved by induction. For the base case we already have $\phi$ for Mat\'ern-$1/2$. Now suppose we have the KP for Mat\'ern-$\nu$, then for Mat\'ern-$(\nu+1)$, remind that $[\beta_{-}\ \beta_0\ \beta_+]$ are the coefficients for Mat\'ern-$1/2$ KP, so
\begin{align*}
    \psi(x)=&\beta_-\int k_{1/2}(x,s)\phi(s+h)ds+\beta_0\int k_{1/2}(x,s)\phi(s)ds+\beta_+\int k_{1/2}(x,s)\phi(s-h)ds\\
    =&\beta_-\int k_{1/2}(x,s-h)\phi(s)ds+\beta_0\int k_{1/2}(x,s)\phi(s)ds+\beta_+\int k_{1/2}(x,s+h)\phi(s)ds\\
    =&\beta_-\int e^{-\omega|x-s+h|}\phi(s)ds+\beta_0\int e^{-\omega|x-s|}\phi(s)ds+\beta_+\int e^{-\omega|x-s-h|}\phi(s)ds\\
    =&\int \phi_0(x-s)\phi(s)ds.
\end{align*}
 Without loss of generality, let $\psi$ be the Mat\'ern-$\nu$ KP induced by points $\{ih\}_{i=-\nu-1/2}^{\nu+1/2}$, which is supported on $(-(\nu+1/2)h,(\nu+1/2)h)$. Using the translation invariant property of  KP,   we have the following identities for rows of matrix $\phi_i(x^*_j)$ associated to central KPs:
 \begin{align}
     \phi_i(x^*_j)=\psi(x^*_j-ih)=\psi\left((j-i)h\right)=\psi(|j-i|h),\ \text{for}\ \nu+3/2 \leq i\leq n-\nu-1/2 \label{eq:proof_prop3.4_1}
 \end{align}
 where the last equality is from the fact that $\phi_0$ is symmetric and $\psi$ is constructed of convolutions of $\phi_0$.  As a result, $\psi=\phi*\phi*\cdots*\phi(x)$ is also symmetric.

For the left-sided KP of Mat\'ern-$1/2$ kernel, we can have the following KP induced by points $\{ih\}_{i=0}^1$ through direct calculation:
\begin{align*}
    \phi_0(x) =\left\{\begin{array}{lc}
           \frac{e^{-\omega(|x|-h)}-e^{-\omega|x-h|}}{2\sinh(\omega h))}>0\  &\text{if} \quad x\leq h\\
             0 \  &\text{otherwise}
        \end{array}\right. .
\end{align*}
 Similar to central KPs, suppose we have the left-sided KP for Mat\'ern-$\nu$, then for Mat\'ern-$\nu+1$, remind that $[\beta_0,\beta_+]$ are the coefficients for Mat\'ern-$1/2$ KP, so
 \begin{align*}
     \psi(x)=&\beta_0\int k_{1/2}(x,s)\phi(s)ds+\beta_+\int k_{1/2}(x,s)\phi(s-h)ds\\
    =&\beta_0\int e^{-\omega|x-s|}\phi(s)ds+\beta_+\int e^{-\omega|x-s-h|}\phi(s)ds
    =\int \phi_0(x-s)\phi(s)ds.
 \end{align*}
Without loss of generality, let $\psi$ be the Mat\'ern-$\nu$ left-sided KP induced by points $\{ih\}_{i=0}^{\nu+1/2}$, which is supported on $(-\infty,(\nu+1/2)h)$.  So using the translation invariant property of KP, we also have
  \begin{align}
     \phi_i(x^*_j)=\psi(x^*_j-ih)=\psi\left((j+i)h\right). \label{eq:proof_prop3.4_2}
 \end{align}
The identity of right-sided KPs can be derived from a similar manner. Now let $\phi_0^{l}$ be the left-sided Mat\'ern-$1/2$ KP induced by points $[0,h]$, $\phi_0^{r}$ be the right-sided Mat\'ern-$1/2$ KP induced by points $[n,n-h]$, and $\phi_0^{c}$ be the Mat\'ern-$1/2$ central KP induced by points $[h,2h,3h]$. By combining \eqref{eq:proof_prop3.4_1} and \eqref{eq:proof_prop3.4_2}, we can have the final result regarding the entries on $[\phi_i(x^*_j)]_{i,j}$
\begin{align*}
     \phi_i(x^*_j) =\left\{\begin{array}{lc}
           \underbrace{\phi_0^{l}*\cdots*\phi_0^{l}}_{\nu+1/2\ \text{convolutions}}((j-i)h)>0\  &\text{if} \quad i\leq \nu+1/2\\
             \underbrace{\phi_0^{c}*\cdots*\phi_0^{c}}_{\nu+1/2\ \text{convolutions}}((j-i)h)>0\  &\text{if} \quad \nu+3/2\leq i\leq n-\nu-1/2\\
              \underbrace{\phi_0^{r}*\cdots*\phi_0^{r}}_{\nu+1/2\ \text{convolutions}}((j+n-i)h)>0\  &\text{if} \quad i\geq n-\nu+1/2
        \end{array}\right.
\end{align*}

\endproof

\proof[ Proposition \ref{prop:KP_estimation}]
We construct the KP matrix $\BFA$ that can satisfy the conditions of Proposition \ref{prop:KP_estimation}. We first define the following normalized constants:
\begin{align*}
    c_l&=\sum_{i=1}^{\nu+1/2} \underbrace{\phi_0^{l}*\cdots*\phi_0^{l}}_{\nu+1/2\ \text{convolutions}}\left((1-i)h\right)=\sum_{i=j}^{\nu+1/2+j-1} \underbrace{\phi_0^{l}*\cdots*\phi_0^{l}}_{\nu+1/2\ \text{convolutions}}\left((j-i)h\right)\\
    c_r&=\sum_{i=n-\nu+1/2}^{n} \underbrace{\phi_0^{r}*\cdots*\phi_0^{r}}_{\nu+1/2\ \text{convolutions}}\left((2n-i)h\right)=\sum_{i=n-\nu+1/2-j}^{n-j} \underbrace{\phi_0^{r}*\cdots*\phi_0^{r}}_{\nu+1/2\ \text{convolutions}}\left((2n-j-i)h\right)\\
    c_c&=\sum_{i=-\nu-1/2}^{\nu+1/2}\underbrace{\phi_0^{c}*\cdots*\phi_0^{c}}_{\nu+1/2\ \text{convolutions}}(ih)=\sum_{i=-\nu-1/2+j}^{\nu+1/2+j}\underbrace{\phi_0^{c}*\cdots*\phi_0^{c}}_{\nu+1/2\ \text{convolutions}}((i-j)h).
\end{align*}
where the second equality in each line is because KPs are translation invariant. We can also notice that if we sum over the rows of the KP matrix $[\phi_i(x^*_j)]_{i,j}$ (over all $i$) we have the following identity regarding $c_l$, $c_c$, and $c_r$:
\begin{equation}
    \label{eq:prop3.5_proof_1}\boldsymbol{1}^\transpose[\phi_i(x^*_j)]_{i,j}=[\underbrace{c_l,\cdots,c_l}_{\nu+1/2\ c_l's}, \underbrace{c_c,\cdots,c_c}_{n-2\nu-1\ c_c's}, \underbrace{c_r,\cdots,c_r}_{\nu+1/2\ c_r's}]   
\end{equation}
 Recall that the $\phi_i$ in Proposition \ref{prop:convolution_Phi} are constructed from the $\BFA_\nu$ in Proposition \ref{prop:convolution_Phi}, i.e. $\BFA k_\nu(\BFX^*,\BFX^*)=[\phi_i(x^*_j)]_{i,j}$. We now construct the desired $\BFA$ from $\BFA_{\nu}$ as follows:
 \begin{align}
     &\BFA_{i,:}=\frac{1}{c_l}[\BFA_{\nu}]_{i,:},\quad i\leq \nu+1/2\nonumber\\
     &\BFA_{i,:}=\frac{1}{c_c}[\BFA_{\nu}]_{i,:},\quad \nu+3/2\leq i\leq n-\nu-1/2\nonumber\\
     &\BFA_{i,:}=\frac{1}{c_r}[\BFA_{\nu}]_{i,:},\quad i\geq n-\nu+1/2.\label{eq:pf_prop_3.5_1}
 \end{align}
Let $[\psi_i]_i=\BFA k_{\nu}(\BFX^*,\cdot)$. Then, it is clear that $\psi_i$ are also KPs, as each row of $\BFA_\nu$ consists of coefficients for constructing a KP, and these coefficients, when multiplied by the scalars $c_l$, $c_c$, or $c_r$, remain valid for constructing KPs. Moreover, $c_l$, $c_c$, and $c_r$ normalize each summation over rows to $1$:
\[\boldsymbol{1}^\transpose[\psi_i(x^*_j)]_{i,j}=[\sum_{i}\psi_i(x^*_j)]_j=[1,\cdots,1]=\boldsymbol{1}^\transpose.\]
This finishes the proof for the identity of $[\psi_{i}(x^*_j)]_{i,j}$.

We now estimate the order of each entry of vector $\BFA\boldsymbol{1}$. Let $s=\nu+1/2$ as before. From the convolution identities in Proposition \ref{prop:convolution_A} and the identities of $\BFA^*$ provided in Proposition \ref{prop:ding_inverseK}, it is straightforward to derive that
\begin{align}
\label{eq:pf_prop_3.5_2}
    [\BFA_\nu\boldsymbol{1}]_i=\left\{\begin{array}{lc}
           (\frac{e^{\omega h}-1}{2\sinh(\omega h)})^{s} =\CalO(1) &\text{if} \quad i\leq \nu+1/2\ \text{or}\ i\geq n-\nu+1/2\\
             (\frac{\sinh(2\omega h)/\sinh(\omega h)-2}{2\sinh(\omega h)})^s=\CalO(h^s)  &\text{if} \quad \nu+3/2\leq i\leq n-\nu-1/2
        \end{array}\right.
\end{align}
where the big O for each line is derived directly from Taylor's expansion. 

The last part of the proof is to estimate the order of $c_l$, $c_c$, and $c_r$. For constants $c_l$ and $c_r$ associated to one-sided KPs, because Mat\'ern-$1/2$ one-sided KPs $\phi_0^l$ and $\phi_0^r$ are supported on $(-\infty,h)$ and $(1-h,\infty)$, respectively. Therefore, self-convolutions of $\phi^l_0$ and $\phi^r_0$ preserve the order of their magnitudes: 
\begin{align}
    \phi^l_0*\cdots *\phi^l_0(jh-ih)=\CalO(1)=\phi^r_0*\cdots*\phi^l_0(jh+1-ih),\quad \forall i,j=1,\cdots,n.\label{eq:c_r_and_c_l}
\end{align}

For the constant $c_c$, given that the Matérn-$\frac{1}{2}$ central KP as in \eqref{eq:central_KP_1-2}, $\phi_0^c(0)=1$ and it has support on $(-h, h)$. So it is straightforward to check by induction that the $(s-1)$-time self-convolution of $\phi_0^c$, which is a KP for the Mat\'ern-$\nu$ kernel, has magnitude as follows:
\begin{equation}
\label{eq:c_c}
    \phi^c_0*\cdots *\phi^c_0(0)\geq C h^{s-1}
\end{equation}
for some universal constant $C$.

Finally, substitute \eqref{eq:pf_prop_3.5_2}, \eqref{eq:c_r_and_c_l}, and $\eqref{eq:c_c}$ into \eqref{eq:pf_prop_3.5_1}, we can have the desired result:
\begin{align*}
     &\sum_j \BFA_{i,j}=\frac{1}{c_l}\sum_j[\BFA_{\nu}]_{i,j}=\CalO(1),\quad i\leq \nu+1/2\\
     &\sum_j \BFA_{i,j}=\frac{1}{c_c}\sum_j[\BFA_{\nu}]_{i,j}=\CalO(h),\quad \nu+3/2\leq i\leq n-\nu-1/2\\
     &\sum_j \BFA_{i,j}=\frac{1}{c_r}\sum_j[\BFA_{\nu}]_{i,j}=\CalO(1),\quad i\geq n-\nu+1/2.
\end{align*}
\endproof

\proof[ Theorem \ref{thm:global_feature}]
We first prove the theorem when kernel $k$ is Mat\'ern-$\nu$ kernel. From Proposition \ref{prop:convolution_Phi}, we have 
\begin{align}
    &\boldsymbol{1}^\transpose\BFA k(\BFX^*,\BFX^*)=\boldsymbol{1}^\transpose[\psi_i(x^*_j)]_{i,j}=\boldsymbol{1}^\transpose. \label{eq:pf_thm_global_feature_1}
\end{align}
For notation simplicity, let $\BFK=k(\BFX^*,\BFX^*)$. From Woodbury Matrix identity, we have
\begin{align*}
    \frac{\boldsymbol{1}^\transpose}{\sqrt{n}} [\rmI_n+\BFK^{-1}]^{-1}\frac{\boldsymbol{1}}{\sqrt{n}}=&\frac{\boldsymbol{1}^\transpose}{\sqrt{n}} [\rmI_n-\BFK^{-1}+\BFK^{-1}[\rmI_n+\BFK^{-1}]^{-1}\BFK^{-1} ]\frac{\boldsymbol{1} }{\sqrt{n}}\\
    \geq & \frac{\boldsymbol{1}^\transpose}{\sqrt{n}} [\rmI_n-\BFK^{-1}]\frac{\boldsymbol{1}}{\sqrt{n}}\\
    =&1-\frac{1}{n}\boldsymbol{1}^\transpose [\psi_{i}(x^*_j)]^{-1}_{i,j}\BFA\boldsymbol{1}\\
    =&1-\frac{1}{n}\boldsymbol{1}^\transpose[\psi_i(x^*_j)]_{i,j}[\psi_{i}(x^*_j)]^{-1}_{i,j}\BFA\boldsymbol{1}\\
    =& 1-\frac{1}{n}\sum_{i,j}\BFA_{i,j}=1-\CalO(\frac{1}{n})
\end{align*}
where the second line is because $\BFK$ is positive definite, the third line is from \eqref{eq:pf_thm_global_feature_1}, and the last line is from the estimation in Proposition \ref{prop:KP_estimation} for the entries of $\BFA$. This finishes the prove for Mat\'ern kernel.

For other kernel $k$ satisfying Assumption \ref{assump:kernel}, i.e., $\mathcal{F}[k](\omega) =\Psi(\omega) \geq C(1+\omega^2)^{-s} $ where $s=\nu+1/2$,  we can show that the spectrum of $k(\BFX^*, \BFX^*)$ are lower bounded by those of the Mat\'ern-$\nu$ kernel matrix $k_\nu(\BFX^*, \BFX^*)$:
    \begin{align*}
        \sum_{i,j}v_iv_jk(x_i,x_j)= &\int {\sum_{i,j}v_ie^{\omega(x_i-x_j)}}v_j\Psi(\omega)d\omega =\int \|[v_ie^{\omega x_i}]_i\|^2_2\Psi(\omega)d\omega\\
        \geq & C_1\int  \|[v_ie^{\omega x_i}]_i\|^2_2 (1+\omega^2)^{-\nu-1/2}d\omega\\
        \geq & C_2  \sum_{i,j}v_iv_jk_{\nu}(x_i,x_j)
    \end{align*}
    for any $[v_1,\cdots,v_n]\in\Real^n$ where $C_1$ and $C_2$ are some universal constant independent of kernel $k$ and $k_\nu$. So $[\rmI_n+[k(\BFX^*,\BFX^*)]^{-1}]^{-1}\geq [\rmI_n+[k_\nu(\BFX^*,\BFX^*)]^{-1}]^{-1}$.
\endproof

\proof[Theorem \ref{thm:back-fit_lower_bound}:]
We only need to show that after $t$ iterations, error reduced by the error form is lower bounded as follows:
\[\|\CalS_n^t\BFS\BFY\|_2\geq (1-\CalO(\frac{1}{\lambda n}))^t.\]

We first define the following unit vector in $\Real^{Dn}$:
\[\BFV_*^\transpose=\frac{1}{\sqrt{n}
}[\boldsymbol{0},\boldsymbol{1}^\transpose,\boldsymbol{0},\cdots,\boldsymbol{0}],\quad\text{where}\ \boldsymbol{0},\boldsymbol{1}\in\Real^n.\]
Recall that the Back-fit operator $\CalS_n=[\BFK^{-1}+\lambda\BFL+\lambda\rmI_{Dn}]^{-1}\lambda\BFL^\transpose$. Let $\BFu_d,\BFw_d\in\Real^n$ be  any $n$-dimensional vector for $d=1,\cdots,D$ such that
\[[\BFu_1;\cdots;\BFu_D]^\transpose=\lambda [\BFw_1;\cdots;\BFw_D]^\transpose[\BFK^{-1}+\lambda\BFL+\lambda\rmI_{Dn}]^{-1}.\]
Then, we can notice that the block Gauss-Seidel iteration gives
\[\BFu_d^\transpose=\lambda\left(\BFw_d^\transpose-\sum_{d'>d}\BFu_{d'}^\transpose\right)[\BFK_d^{-1}+\lambda\rmI_n]^{-1}.\]
Substitute $\BFw=\BFV_*=[\boldsymbol{0};\boldsymbol{1};\boldsymbol{0};\cdots;\boldsymbol{0}]$, we can immediately get
\begin{align*}
    &\BFV_*^\transpose\lambda[\BFK^{-1}+\lambda\BFL+\lambda\rmI_{Dn}]^{-1}
    =\left[\BFV_1,\BFV_2,\boldsymbol{0},\cdots,\boldsymbol{0}\right]
\end{align*}
where
\begin{align*}
    &\BFV_1=-\frac{1}{\sqrt{n}}\boldsymbol{1}^\transpose\lambda^2[\BFK_2^{-1}+\lambda\rmI_n]^{-1}[\BFK_1^{-1}+\lambda\rmI_n]^{-1}\\
    &\BFV_2=\frac{1}{\sqrt{n}}\boldsymbol{1}^\transpose\lambda[\BFK_2^{-1}+\lambda\rmI_n]^{-1}.
\end{align*}
On the other hand $\BFL^\transpose\BFV_*=\frac{1}{\sqrt{n}}[\boldsymbol{1};\boldsymbol{0};\cdots;\boldsymbol{0}]$. So we have
\begin{align*}
    \left|\BFV_*^\transpose\CalS_n\BFV_*\right|=&\left|\BFV_*^\transpose\lambda[\BFK^{-1}+\lambda\BFL+\lambda\rmI_{Dn}]^{-1}\BFL^\transpose\BFV_*\right|\\
    =&\left|\BFV_1\frac{\boldsymbol{1}}{\sqrt{n}}\right|\\
    =&\frac{1}{n}\boldsymbol{1}^\transpose\lambda^2[\BFK_2^{-1}+\lambda\rmI_n]^{-1}[\BFK_1^{-1}+\lambda\rmI_n]^{-1}\boldsymbol{1}\\
    \geq&1-\CalO(\frac{1}{\lambda n})
\end{align*}
where the last line is from the fact that the third line is exactly the error for the two-dimensional counterexample  in Section \ref{sec:counter_example_2d}.

Because $\BFV_*$ is a normalized vector, we can conclude that the largest eigenvalue of the Back-fit operator $\CalS_n$ is greater than $1-\CalO(\frac{1}{n})$:
\[\max_{\|\BFv\|_2^2\leq 1}\BFv^\transpose\CalS_n\BFv\geq 1-\CalO(\frac{1}{\lambda n}).\]

Let $\BFE_*$ be the eigenvector associated to the largest eigenvalue $\lambda_*\geq 1-\CalO(\frac{1}{n})$. Then the projection of initial error $\BFY=\sum_{d=1}^2\CalG_d(\BFX_d)+\varepsilon$ onto $\BFE_*$ can be lower bounded simply as follows:
\begin{align*}
    \E\|\BFE_*^\transpose\BFS\left(\sum_{d=1}^2\CalG_d(\BFX_d)+\varepsilon\right)\|_2^2\geq \E|\sum_{i=1}^n[\BFE_*]_i\varepsilon_i|^2=\sigma_y^2.
\end{align*}
It can be directly deduced that in each iteration, at most a $1-\CalO(\frac{1}{n})$ fraction of the error projected onto $\BFE_*$ is eliminated. By induction, the fraction of error eliminated after $t$ iterations is at most $(1-\CalO(\frac{1}{n}))^t$:
\[\E\|\CalS_n^t\BFS\left(\sum_{d=1}^2\CalG_d(\BFX_d)+\varepsilon\right)\|_2\geq \left(\BFE_*^\transpose\CalS_n\BFE_*\right)^t\sigma_y\geq (1-\CalO(\frac{1}{\lambda n}))^t.\]
\endproof

\section{Approximation Property of Sparse Additive GPR }
We first introduce lemmas that will be used in our later proof. 
\subsection{Useful Lemmas}
\begin{lemma}[Corollary 10.13 in \cite{wendland2004scattered}]
    \label{lem:pf_thm4-1_RKHS_Sobolev}
    Let $\CalH_k$ be RKHS induced by one-dimensional kernel $k$ satisfying
    Assumption \eqref{assump:kernel}.
     Then $\CalH_k$ is equivalent to the $s$-th order Sobolev space:
    \[\CalW^{s}=\{f:\ \frac{\partial ^{j}f}{\partial x^{j}}\in L_2, \forall j\leq s\}.\]
\end{lemma}

\begin{lemma}[Theorem 3.3 \& 3.4 in \cite{utreras1988convergence}]
\label{lem:pf_thm4-1_sampling_inequality}
    Let $\CalW^s$ be the $s$-th order Sobolev space. Let $\BFX=\{x_i\}_{i=1}^n$ be any point set satisfying the following fill distance condition:
    \begin{equation}
    \label{eq:ph_thm4-1_sample_fill_distance}
        \max_{x\in[0,1]}\min_{i=1,\cdots,n}|x_i-x|\leq \CalO(h)
    \end{equation}
    for some value $h$. Then for  any function $f\in\CalW^s$, we have
    \begin{align*}
        &\frac{1}{n}\|f(\BFX)\|_{2}^2\leq C_1\left(\|f\|_{L_2}^2+{h^{2\nu+1}}\|f\|_{\CalW^s}^2\right),\\
    &\|f\|_{L_2}^2\leq C_2\left(\frac{1}{n}\|f(\BFX)\|_{2}^2+{h^{2\nu+1}}\|f\|_{\CalW^s}^2\right)
    \end{align*}
    where $\|\cdot\|_{\CalW^s}$ denote the Sobolev norm associated to $\CalW^s$, $C_1$ and $C_2$ are some constants independent of $f$ and $h$.
\end{lemma}
From Lemma \ref{lem:pf_thm4-1_sampling_inequality}, we can derive the following error estimate for one-dimensional interpolation.
\begin{lemma}
    \label{lem:pf_thm4-1_markov_convergece}
    Let $k$ be a one-dimensional kernel satisfying Assumption \eqref{assump:kernel}. For any one-dimensional point set $\{\BFX\}$ satisfying the fill distance condition \eqref{eq:ph_thm4-1_sample_fill_distance} and function $f\in\CalH_k$, we have
    \[\|f-k(\cdot,\BFX)[k(\BFX,\BFX)]^{-1}f(\BFX)\|_{L_2}\leq \CalO(h^{s}\|f\|_{\CalH_k}).\]
\end{lemma}
\proof
Let $\hat{f}$ denote the interpolator $k(\cdot,\BFX)[k(\BFX,\BFX)]^{-1}f(\BFX)$. From the representation theorem, we have
\[\hat{f}=\min_{g\in\CalH_k}\frac{1}{n}\|f(\BFX)-g(\BFX)\|_2^2.\]
Obviously, $\frac{1}{n}\|f(\BFX)-\hat{f}(\BFX)\|_{2}^2=0$. From Lemma \ref{lem:pf_thm4-1_RKHS_Sobolev}, we know that $\CalH_k$ is equivalent to the $s$-th order Sobolev space $\CalW^s$.  We the can directly have the result from Lemma \ref{lem:pf_thm4-1_sampling_inequality}.
\endproof

The following Lemma highlights key geometric characteristics commonly employed in the theory of sparse recovery, notably the concept of "restricted isometry constants" as introduced in \cite{Dantzig_Selector}. We adopt its kernel version as proposed by \cite{Koltchinskii2010}.

\begin{lemma}[Proposition 1 in \cite{Koltchinskii2010}]
\label{lem:pf_thm4-1_kolchinskii}
    Let $\pi$ be any distribution on $[0,1]^D$. Let $k_d$ be one-dimensional kernel satisfying Assumption \ref{assump:kernel}. For any linear independent functions $h_d\in\CalH_{k_d}$, let $\BFG_{d,t}=\int h_d h_t d\pi$ be the gram matrix induced by $\{h_d\}_{d=1}^D$. Let $\kappa(\{h_d\}_{d=1}^D)$ be the minimum eigenvalue of $\BFG_{d,t}$. Define
    \begin{equation}
    \label{eq:pf_thm4-1_kolchinskii_kappa}
        \kappa^*=\inf\{\kappa(\{h_d\}_{d=1}^D): h_d\in\CalH_d,\int h_d^2d\pi=1\}
    \end{equation}
    Then for any $f_d\in\CalH_{k_d}$, we have
    \[\sum_{d=1}^D\int f_d^2 d\pi \leq \frac{1}{\kappa^*}\int \left|\sum_{d=1}^Df_d\right|^2 d\pi.\]
\end{lemma}

From Lemma \ref{lem:pf_thm4-1_kolchinskii}, we can derive the following sampling inequality for function of additive form $f=\sum_{d=1}^Df_d$.

\begin{lemma}
    \label{lem:pf_thm4-1_interpolation_inequality}
    Let $k=\sum_{d=1}^Dk_d$ be an additive kernel with $k_d$ satisfying Assumpyion \eqref{assump:kernel}. Let $\BFX=\{\BFx_i\}_{i=1}^n\subset[0,1]^D$ be any point set satisfying \eqref{eq:ph_thm4-1_sample_fill_distance} . Then for $f\in\text{span}\{k(\BFx,\cdot):\BFx\in\BFX\}$, we have the following interpolation inequality
    \[\max_{\BFx\in[0,1]^D}|f(\BFx)|^2\leq \CalO\left(\|f\|_{\CalH_k}^{\frac{1}{s}}\left|\frac{1}{{\kappa^*_n n}}\|f(\BFX)\|^2_2\right|^{1-\frac{1}{2s}}\right)\]
    where $s=\nu+1/2$ and $\kappa^*_n$ is the $\kappa^*$ in \eqref{eq:pf_thm4-1_kolchinskii_kappa} induced by empirical density $\frac{1}{n}\sum_{i=1}^n\delta_{\BFx_i}$. 
\end{lemma}
\proof{}
For any $f\in \text{span}\{k(\BFx,\cdot):\BFx\in\BFX\}$, it must be in the form
\[f=\BFalpha^\transpose k(\BFX,\cdot)=\sum_{d=1}^D\BFalpha^\transpose k_d(\BFX_d,\cdot),\quad \|f\|_{\CalH_k}^2=\BFalpha^\transpose\BFK\BFalpha=\sum_{d=1}^D\BFalpha^\transpose\BFK_d\BFalpha,,\quad \BFalpha\in\Real^n. \]
Then, the sup norm of $f$ can be bounded as follows:
\begin{align*}
    \max_{\BFx\in[0,1]^D}|f(\BFx)|^2&= \max_{\BFx\in[0,1]^D}|\sum_{d=1}^D\BFalpha^\transpose k_d(\BFX_d,x_d)|^2\\
    &\leq D \sum_{d=1}^D\max_{x_d\in[0,1]}|\BFalpha^\transpose k_d(\BFX_d,x_d)|^2\\
    &\leq  C_1 \sum_{d=1}^D \|\BFalpha^\transpose k_d(\BFX,\cdot)\|_{\CalH_{k_d}}^{\frac{1}{s}}\|\BFalpha^\transpose k_d(\BFX,\cdot)\|_{L_2
    }^{2-\frac{1}{s}}\\
    &\leq C_2 \sum_{d=1}^D\left({\BFalpha^\transpose\BFK_d\BFalpha}\right)^{\frac{1}{2s}} \left(\frac{1}{n}\sum_{i=1}^n\big|\BFalpha^\transpose k_d(\BFX_d,\BFx_i)\big|^2\right)^{1-\frac{1}{2s}}\\
    &\leq C_2 \left({\BFalpha^\transpose\BFK \BFalpha}\right)^{\frac{1}{2s}}\left({\sum_{d=1}^D\frac{1}{n}\sum_{i=1}^n\big|\BFalpha^\transpose k_d(\BFX_d,\BFx_i)\big|^2}\right)^{1-\frac{1}{2s}}\\
    &\leq C_2 \|f\|_{\CalH_k}^{\frac{1}{s}}\left({\frac{1}{n\kappa^*_n}\sum_{i=1}^n\left|\sum_{d=1}^D\BFalpha^\transpose k_d(\BFX_d,x_{i,d})\right|^2}\right)^{1-\frac{1}{2s}} 
\end{align*}
where the third line is from Lemma \ref{lem:pf_thm4-1_RKHS_Sobolev} and Gagliardo–Nirenberg interpolation inequality, the fourth line is from Lemma \ref{lem:pf_thm4-1_sampling_inequality}, the fifth line is from H\"oler inequality, and the last line is from Lemma \ref{lem:pf_thm4-1_kolchinskii}.

\endproof
\subsection{Main Proof of Theorem \ref{thm:sparse_GPR}}
Equipped with Lemmas introduced in the previous subsection, we now can start our proof for Theorem \ref{thm:sparse_GPR}.

\proof[ Theorem \ref{thm:sparse_GPR}]
 For any function  $g_d\in\text{span}\{k_d(\cdot,x_{i,d}):i=1,\cdots,n\}$, 
we define the following vector as the values of $g_d$ on the data points $\BFX_d$:
\[\BFE_n=\begin{bmatrix}
    g_1(\BFX_1)& g_2(\BFX_2)&\cdots&g_D(\BFX_D)
\end{bmatrix}^\transpose\in\Real^{Dn}.\]
So the RKHS norm of $g_d$ is 
\[\|g\|^2_{\CalH_{k_d}}=g_d(\BFX_d)^\transpose\BFK_d^{-1}g_d(\BFX_d).\]
and we can also define the following vector as the values of $g_d$ on the inducing points $\BFU_d\subset\BFX_d$:
\[\BFE_m=\begin{bmatrix}
    g_1(\BFU_1)& g_2(\BFU_2)&\cdots&g_D(\BFU_D)
\end{bmatrix}^\transpose\in\Real^{Dn}.\]

The equation can be written as
\begin{align}
     &\quad \BFK_{n,m}\BFK_{m,m}^{-1}[\lambda^{-1}\BFK_{m,m}^{-1}+\BFSigma_{m,m}]^{-1}\BFK_{m,m}^{-1}\BFK_{m,n}[\lambda^{-1}\BFK_{n,n}^{-1}+\BFS\BFS^\transpose]\BFE_n\nonumber\\
    &=\lambda^{-1}\BFK_{n,m}\BFK_{m,m}^{-1}[\lambda^{-1}\BFK_{m,m}^{-1}+\BFK_{m,m}^{-1}\BFK_{m,n}\BFS\BFS^\transpose\BFK_{n,m}\BFK_{m,m}^{-1}]^{-1}\BFK_{m,m}^{-1}\BFE_m\nonumber\\
    &\ +\BFK_{n,m}\BFK_{m,m}^{-1} [\lambda^{-1}\BFK_{m,m}^{-1}+\BFK_{m,m}^{-1}\BFK_{m,n}\BFS\BFS^\transpose\BFK_{n,m}\BFK_{m,m}^{-1}]^{-1}\BFK_{m,m}^{-1}\BFK_{m,n}\BFS\BFS^\transpose\BFE_n\label{eq:pf_thm4-1_2_eq1}.
\end{align}
We define $\BFD_n=\BFE_n-\BFK_{n,m}\BFK_{m,m}^{-1}\BFE_m$,
and substitute $\BFD_n$ into \eqref{eq:pf_thm4-1_2_eq1}:
\begin{align}
        &\quad \BFK_{n,m}\BFK_{m,m}^{-1}[\lambda^{-1}\BFK_{m,m}^{-1}+\BFSigma_{m,m}]^{-1}\BFK_{m,m}^{-1}\BFK_{m,n}[\lambda^{-1}\BFK_{n,n}^{-1}+\BFS\BFS^\transpose]\BFE_n\nonumber\\
        &=\BFK_{n,m}\BFK_{m,m}^{-1}[\lambda^{-1}\BFK_{m,m}^{-1}+\BFK_{m,m}^{-1}\BFK_{m,n}\BFS\BFS^\transpose\BFK_{n,m}\BFK_{m,m}^{-1}]^{-1}\bigg(\big[\lambda^{-1}\BFK^{-1}_{m,m}\nonumber\\
        &\quad\quad +\BFK_{m,m}^{-1}\BFK_{m,n}\BFS\BFS^\transpose\BFK_{n,m}\BFK_{m,m}^{-1}\big]\BFE_m+\BFK_{m,m}^{-1}\BFK_{m,n}\BFS\BFS^\transpose\BFD_n\bigg)\nonumber\\
        &=\BFK_{n,m}\BFK_{m,m}^{-1}\BFE_m\label{eq:pf_thm4-1_2_eq2}\\
        &\quad +\BFK_{n,m}\BFK_{m,m}^{-1} \underbrace{[\lambda^{-1}\BFK_{m,m}^{-1}+\BFK_{m,m}^{-1}\BFK_{m,n}\BFS\BFS^\transpose\BFK_{n,m}\BFK_{m,m}^{-1}]^{-1}\BFK_{m,m}^{-1}\BFK_{m,n}\BFS\BFS^\transpose\BFD_n}_{\tilde{D}_m}. \label{eq:pf_thm4-1_2_eq3}
\end{align}
For $\BFD_n$ and $\BFK_{n,m}\BFK_{m,m}^{-1}\BFE_m$, we can use Lemma \ref{lem:pf_thm4-1_sampling_inequality} and Lemma \ref{lem:pf_thm4-1_markov_convergece} to show that
\begin{align}
    &\frac{1}{\sqrt{n}}\|\BFD_n\|_2=\frac{1}{\sqrt{n}}\|\BFE_n-\BFK_{n,m}\BFK_{m,m}^{-1}\BFE_m\|_2\leq\CalO(h_m^s{\sqrt{\BFE_n^\transpose\BFK_{n,n}^{-1}\BFE_n}})\label{eq:pf_thm4-1_2_eq4}
\end{align}
We now estimate the $\tilde{D}_m$ in \eqref{eq:pf_thm4-1_2_eq3}. Multiply  it by $\BFS^\transpose_m$:
\begin{align}
     &\quad \BFS^\transpose_m[\lambda^{-1}\BFK_{m,m}^{-1}+\BFK_{m,m}^{-1}\BFK_{m,n}\BFS\BFS^\transpose\BFK_{n,m}\BFK_{m,m}^{-1}]^{-1}\BFK_{m,m}^{-1}\BFK_{m,n}\BFS\BFS^\transpose\BFD_n\nonumber\\
     &=\lambda \left(\underbrace{\BFS_m^\transpose\tilde{\BFK}_{m,n}\BFS- \BFS^\transpose_m\tilde{\BFK}_{m,n}\BFS[\BFS^\transpose\tilde{\BFK}_{n,n}\BFS+\lambda^{-1}\rmI_{Dn}]^{-1}\BFS^\transpose\tilde{\BFK}_{n,n}\BFS}_{\BFM}\right)\BFS^\transpose\BFD_n\label{eq:pf_thm4-1_2_eq6}
\end{align}
where $\tilde{\BFK}_{m,n}$ and $\tilde{\BFK}_{n,n}$ are low-rank kernel matrices:
\begin{align*}
    &\tilde{\BFK}_{m,n}=\text{diag}[\tilde{k}_1(\BFU_1,\BFX_1),\cdots,\tilde{k}_D(\BFU_D,\BFX_D)],\nonumber\\
    &\tilde{\BFK}_{n,n}=\text{diag}[\tilde{k}_1(\BFX_1,\BFX_1),\cdots,\tilde{k}_D(\BFX_D,\BFX_D)],\\
    &\tilde{k}_d(x,x')=k(x,\BFU_d)[k_d(\BFU_d,\BFU_d)]^{-1}k_d(\BFU_d,x').
\end{align*}
Let $\tilde{k}=\sum_{d=1}^D\tilde{k}_d$ and
\[\hat{k}(\BFx,\BFx')= \tilde{k}(\BFx,\BFX)[\tilde{k}(\BFX,
\BFX)+\lambda^{-1}\rmI_{Dn}]^{-1}\tilde{k}(\BFX,\BFx').\]
We can notice that the matrix $\BFM$ in \eqref{eq:pf_thm4-1_2_eq6} has the following identities for all its entries
\[\tilde{k}(\BFu_i,\BFx_j)-\hat{k}(\BFu_i,\BFx_j)=\left[\BFS_m^\transpose\tilde{\BFK}_{m,n}\BFS-\BFS^\transpose_m\tilde{\BFK}_{m,n}\BFS[\BFS^\transpose\tilde{\BFK}_{n,n}\BFS+\lambda^{-1}\rmI_{Dn}]^{-1}\BFS^\transpose\tilde{\BFK}_{n,n}\BFS\right]_{i,j}\]
and for any $\BFu_i$, we have
\[\hat{k}(\BFu_i,\cdot)=\argmin_{h\in\CalH_{\tilde{k}}}\frac{1}{ n}\|h(\BFX)-\tilde{k}(\BFX,\BFu_i)\|^2_2+\frac{1}{\lambda n}\|h\|_{\CalH_{\tilde{k}}}^2\]
so we can bound the empirical norm of each $\hat{k}(\BFu_i)$ as follows:
\begin{equation}
\label{eq:pf_thm4-1_2_eq7}
    \frac{1}{n}\|\hat{k}(\BFX,\BFu_i)-\tilde{k}(\BFX,\BFu_i)\|^2_2\leq \frac{1}{\lambda n}\|\tilde{k}(\BFu_i),\cdot)\|_{\CalH_{\tilde{k}}}^2=\frac{1}{\lambda n}\tilde{k}(\BFu_i,\BFu_i).
\end{equation}
For any function $f\in\CalH_{\tilde{k}}$, we define the following linear operator for notation convenient:
\[\CalL[f](\BFx)=\tilde{k}(\BFx,\BFX)[\BFS^\transpose\tilde{\BFK}_{n,n}\BFS+\lambda^{-1}\rmI_{Dn}]^{-1}f(\BFX)\]
Obviously, $f$ is also in $\CalH_k$ and
\begin{align*}
    \|f\|_{\CalH_{\tilde{k}}}^2=&\|\BFalpha_f^\transpose\tilde{k}(\BFU,\cdot)\|_{\CalH_{\tilde{k}}}^2
    =\BFalpha_f^\transpose \tilde{k}(\BFU,\BFU)\BFalpha_f
    = \BFalpha_f^\transpose {k}(\BFU,\BFU)\BFalpha_f =\|f\|_{\CalH_k}.
\end{align*}
Then using Lemma \ref{lem:pf_thm4-1_interpolation_inequality}, we have for any $\BFx_j\in\BFX$:
\begin{align}
    &\left|\tilde{k}(\BFx,\BFx_j)-\hat{k}(\BFx,\BFx_j)\right|^2=\left|\tilde{k}(\BFx,\BFx_j)-\CalL[\tilde{k}(\BFx_j,\cdot)](\BFx_j)\right|^2\nonumber\\=
    &\left|\left(\rmI-\CalL\right)[\tilde{k}(\BFx_j,\cdot)](\BFx_j)\right|^2\nonumber\\
    \leq& \| \tilde{k}(\BFx_j,\cdot)-\CalL[\tilde{k}(\BFx_j,\cdot)]\|_{\CalH_{\tilde{k}}}^{{1}/{s}}\left(\frac{1}{{n\kappa^*_n}}\|\hat{k}(\BFX,\BFx_j)-\tilde{k}(\BFX,\BFx_j)\|^2_2\right)^{1-\frac{1}{2s}}\nonumber\\
    \leq& \left([\rmI-\CalL]^2[\tilde{k}(\BFx_j,\cdot)](\BFx_j) \right)^{\frac{1}{2s}}\left(\frac{k(\BFx_j,\BFx_j)}{\lambda n\kappa^*_n}\right)^{1-\frac{1}{2s}}\label{eq:pf_thm4-1_2_eq8}
\end{align}
where the last line is from \eqref{eq:pf_thm4-1_2_eq7} and direct calculations. 

It is straightforward to check that $\rmI-\CalL$ is positive definite. So for any $\BFx$, we have
\begin{align*}
    \max_{\|f\|_{\CalH_{\tilde{k}}}\leq \tilde{k}(\BFx,\BFx)}(\rmI-\CalL)[f](\BFx)&=\max_{\|f\|_{\CalH_{\tilde{k}}}\leq \tilde{k}(\BFx,\BFx)}\langle (\rmI-\CalL)[f],\tilde{k}(\BFx,\cdot)\rangle_{\CalH_{\tilde{k}}}\\
    &\leq \langle(\rmI-\CalL)[\tilde{k}(\BFx,\cdot)],\tilde{k}(\BFx,\cdot)\rangle_{\CalH_{\tilde{k}}}\\
    &=(\rmI-\CalL)[\tilde{k}(\BFx,\cdot)](\BFx).
\end{align*}
 Also,  $\|(\rmI-\CalL)[\tilde{k}(\BFx,\cdot)]\|_{\CalH_{\tilde{k}}}\leq \|\tilde{k}(\BFx,\cdot)\|_{\CalH_{\tilde{k}}}$ because, obviously, spectrum of  $\rmI-\CalL$ is less than 1. So
\begin{align*}
    (\rmI-\CalL)^2[\tilde{k}(\BFx,\cdot)]&\leq  \left[\max_{\|f\|_{\CalH_{\tilde{k}}}\leq \tilde{k}(\BFx,\BFx)}(\rmI-\CalL)[f]\right]\|(\rmI-\CalL)[\tilde{k}(\BFx,\cdot)]\|_{\CalH_{\tilde{k}}}.
\end{align*}
 From previous calculations, we have already known $\|(\rmI-\CalL)[\tilde{k}(\BFx,\cdot)]\|_{\CalH_{\tilde{k}}}^2=(\rmI-\CalL)^2[\tilde{k}(\BFx,\cdot)](\BFx)$. So
\begin{equation}
    \| \tilde{k}(\BFx_j,\cdot)-\CalL[\tilde{k}(\BFx_j,\cdot)]\|_{\CalH_{\tilde{k}}}\leq (\rmI-\CalL)[\tilde{k}(\BFx,\cdot)](\BFx) \label{eq:pf_thm4-1_2_eq9}
\end{equation}
Substitute \eqref{eq:pf_thm4-1_2_eq9} into \eqref{eq:pf_thm4-1_2_eq8}, and remind that $ \tilde{k}(\BFx,\BFx_j)-\hat{k}(\BFx,\cdot)=(\rmI-\CalL)[\tilde{k}(\BFx,\cdot)]$, we can have
\begin{equation*}
  \tilde{k}(\BFx,\BFx_j)-\hat{k}(\BFx,\BFx_j)\leq \sqrt{\frac{k(\BFx,\BFx)}{\lambda n\kappa^*_n}}.
\end{equation*}
It is straightforward to check that for any $\BFx,\BFx'$, $\tilde{k}(\BFx,\BFx')-\hat{k}(\BFx,\BFx')$ is a positive definite kernel. As a result, for any $\BFu_i,\BFx_j$, we can derive the following upper bound:
\begin{align}
    \tilde{k}(\BFu_i,\BFx_j)-\hat{k}(\BFu_i,\BFx_j)&\leq \sqrt{\left(\tilde{k}(\BFu_i,\BFu_i)-\hat{k}(\BFu_i,\BFu_i)\right)\left(\tilde{k}(\BFx_j,\BFx_j)-\hat{k}(\BFx_j,\BFx_j)\right)}\nonumber\\
    &\leq \sqrt{\frac{\max_{\BFx}k(\BFx,\BFx)}{\lambda n\kappa^*_n}}.\label{eq:pf_thm4-1_2_eq10}
\end{align}
Substitute\eqref{eq:pf_thm4-1_2_eq10} into \eqref{eq:pf_thm4-1_2_eq6},  we can have
\begin{align}
    &\quad \frac{1}{m}\|\left(\BFS_m^\transpose\tilde{\BFK}_{m,n}\BFS-\BFS^\transpose_m\tilde{\BFK}_{m,n}\BFS[\BFS^\transpose\tilde{\BFK}_{n,n}\BFS+\lambda^{-1}\rmI_{Dn}]^{-1}\BFS^\transpose\tilde{\BFK}_{n,n}\BFS\right)\BFS^\transpose\BFD_n\|_2^2\nonumber\\
    &=\frac{1}{m}\sum_{i=1}^m\left|\sum_{j=1}^n\left(\tilde{k}(\BFu_i,\BFx_j)-\hat{k}(\BFu_i,\BFx_j)\right)[\BFS^\transpose\BFD_n]_j\right|^2\nonumber\\
    &\leq \frac{1}{m}\sum_{i=1}^m\max_{i,j}\left|\tilde{k}(\BFu_i,\BFx_j)-\hat{k}(\BFu_i,\BFx_j)\right|^2\sum_{j=1}^n\left|[\BFS^\transpose\BFD_n]_j\right|^2\leq \CalO(\frac{h_m^{2s}{{\BFE_n^\transpose\BFK_{n,n}^{-1}\BFE_n}}}{\lambda \kappa^*_n}) \label{eq:pf_thm4-1_2_eq11}
\end{align}
where the last line is from \eqref{eq:pf_thm4-1_2_eq4}. Then we can apply Lemma \ref{lem:pf_thm4-1_kolchinskii} on \eqref{eq:pf_thm4-1_2_eq11} to have the following upper bound for \eqref{eq:pf_thm4-1_2_eq3}: 
\begin{align}
    \frac{1}{m}\|\tilde{D}_m\|_2^2=& \frac{1}{m}\|[\BFK_{m,m}^{-1}+\BFK_{m,m}^{-1}\BFK_{m,n}\BFS\BFS^\transpose\BFK_{n,m}\BFK_{m,m}^{-1}]^{-1}\BFK_{m,m}^{-1}\BFK_{m,n}\BFS\BFS^\transpose\BFD_n\|_2^2 \nonumber\\
    \leq &\frac{1}{\kappa^*_m m}\|\left(\BFS_m^\transpose\tilde{\BFK}_{m,n}\BFS-\BFS^\transpose_m\tilde{\BFK}_{m,n}\BFS[\BFS^\transpose\tilde{\BFK}_{n,n}\BFS+\rmI_{Dn}]^{-1}\BFS^\transpose\tilde{\BFK}_{n,n}\BFS\right)\BFS^\transpose\BFD_n\|_2^2 \nonumber\\
    \leq& \CalO(\frac{h_m^{2s}{{\BFE_n^\transpose\BFK_{n,n}^{-1}\BFE_n}}}{\lambda \kappa^*_n\kappa_m^*}). \label{eq:pf_thm4-1_2_eq12}
\end{align}

The final part of our proof is to estimate  $\frac{1}{\sqrt{n}}\|\BFK_{n,n}\BFK_{m,m}^{-1}\tilde{D}_m\|_2$ in \eqref{eq:pf_thm4-1_2_eq3}. Let $$\CalD(\BFx)=\BFK_{\BFx,m}\BFK_{m,m}^{-1}\tilde{D}_m=[\CalD_1(x_1),\cdots,\CalD_D(x_D)]^\transpose$$
So $\frac{1}{\sqrt{n}}\|\BFK_{n,n}\BFK_{m,m}^{-1}\tilde{D}_m\|_2=\frac{1}{n}\|\CalD(\BFX)\|_2$. From Theorem \ref{lem:pf_thm4-1_sampling_inequality}, it is straightforward to derive the following sampling inequality
\begin{equation}
    \label{eq:pf_thm4-1_2_eq13}
    \frac{1}{n}\|\CalD(\BFX)\|_2^2\leq \|\CalD\|_{L_2}^2+\sum_{d=1}^Dh_n^{2s}\|\CalD_d\|^2_{\CalH_{k_d}}\leq \frac{1}{m}\|\CalD(\BFU)\|_2^2+\sum_{d=1}^D2h_m^{2s}\|\CalD_d\|^2_{\CalH_{k_d}}
\end{equation}

Obviously, $\frac{1}{m}\|\CalD(\BFU)\|_2^2=\frac{1}{m}\|\tilde{D}_m\|_2^2\leq \CalO(\frac{h_m^{2s}{{\BFE_n^\transpose\BFK_{n,n}^{-1}\BFE_n}}}{\lambda \kappa^*_n\kappa_m^*})$ so we only need to estimate the RKHS norm of $\CalD$. From long but straightforward calculations, we can have
\begin{align}
    \sum_{d=1}^D\|\CalD_d\|^2_{\CalH_{k_d}}=&\lambda^2\BFD_n^\transpose\BFS \bigg[\Bar{\BFK}_{n,n}-\Bar{\BFK}_{n,n}[\Bar{\BFK}_{n,n}+\lambda^{-1}\rmI]^{-1}\Bar{\BFK}_{n,n}\nonumber\\
    \quad&-\lambda^{-1}\Bar{\BFK}_{n,n}[\Bar{\BFK}_{n,n}+\lambda^{-1}\rmI]^{-2}\Bar{\BFK}_{n,n}\bigg]\BFS^\transpose\BFD_n\nonumber\\
    \leq & \lambda^2\BFD_n^\transpose\BFS [{\Bar{\BFK}_{n,n}-\Bar{\BFK}_{n,n}[\Bar{\BFK}_{n,n}+\lambda^{-1}\rmI]^{-1}\Bar{\BFK}_{n,n}}]\BFS^\transpose\BFD_n\nonumber\\
    =&\lambda \BFD_n^\transpose\BFS [\lambda {\Bar{\BFK}_{n,n}-\sum_{l=2}^{\infty}\left(-\lambda \Bar{\BFK}_{n,n}\right)^l}]\BFS^\transpose\BFD_n\nonumber \\
    \leq &\lambda \|\BFD_n\|_2^2\label{eq:pf_thm4-1_2_eq14}
\end{align}
where $\Bar{\BFK}_{n,n}=\sum_{d=1}^D\tilde{k}_d(\BFX_d,\BFX_d)$ and the fourth line is from the Neumann series expansion for $[\rmI+\lambda\Bar{\BFK}_{n,n}]^{-1}$.

Substitute \eqref{eq:pf_thm4-1_2_eq14} and \eqref{eq:pf_thm4-1_2_eq4} into \eqref{eq:pf_thm4-1_2_eq13}, we can have the following upper bound 
\begin{equation}
    \sum_{d=1}^D\|\CalD_d\|^2_{\CalH_{k_d}}\leq  \lambda \|\BFD_n\|_2^2 \leq \CalO\left( \lambda h_m^{2s} n \BFE_n^\transpose\BFK_{n,n}^{-1}\BFE_n  \right) \label{eq:pf_thm4-1_2_eq15}
\end{equation}
and substitute \eqref{eq:pf_thm4-1_2_eq15} into \eqref{eq:pf_thm4-1_2_eq13}, we can have the following estimate for \eqref{eq:pf_thm4-1_2_eq3}:
\begin{equation}
    \frac{1}{n}\|\BFK_{n,m}\BFK_{m,m}^{-1}\tilde{D}_m\|_2^2=\frac{1}{n}\|\CalD(\BFX)\|_2^2\leq \CalO(\frac{h_m^{2s}\BFE_n^\transpose\BFK^{-1}_{n,n}\BFE_n}{\lambda \kappa_n^*\kappa_m^*}+\lambda h_m^{4s}n\BFE_n^\transpose\BFK^{-1}_{n,n}\BFE_n)\label{eq:pf_thm4-1_2_eq16}
\end{equation}

Finally, substitute \eqref{eq:pf_thm4-1_2_eq16} and \eqref{eq:pf_thm4-1_2_eq4} into \eqref{eq:pf_thm4-1_2_eq1},we can have the desired result:
\begin{align*}
    &\frac{1}{\sqrt{n}}\|\BFK_{n,m}\BFK_{m,m}^{-1}[\BFK_{m,m}^{-1}+\BFSigma_{m,m}]^{-1}\BFK_{m,m}^{-1}\BFK_{m,n}[\BFK_{n,n}^{-1}+\BFS\BFS^\transpose]\BFE_n\|_2\\
    \leq&\CalO\left([\frac{h_m^{s}}{\sqrt{\lambda \kappa^*_n\kappa_m^*}}+\sqrt{\lambda n}h_m^{2s}]\sqrt{\BFE_n^\transpose\BFK^{-1}_{n,n}\BFE_n}\right).
\end{align*}
\endproof

\section{Convergence Rate of Kernel Multigird}
\subsection{Smoothing Property of \texorpdfstring{$\CalS_n$}{Lg}}

\proof[Lemma \ref{lem:smoothing_property}]
We first write the Back-fit operator $\CalS_n$ in error form $\BFvarepsilon^{(t)}=\CalS_n\BFvarepsilon^{(t-1)}$. Denote the error in the $t$-th iteration as $\BFvarepsilon^{(t)}=[\BFvarepsilon^{(t)}_1,\cdots,\BFvarepsilon^{(t)}_D]$ where $\BFvarepsilon^{(t)}_d$ corresponds to the error of dimension $d$. Then each Back-fitting iteration can be written a
\begin{align}
    -\BFvarepsilon_d^{(t)}=&[\lambda^{-1}\BFK^{-1}_d+\rmI_n]^{-1}\left(\sum_{d'<d}\BFvarepsilon_{d'}^{(t)}+\sum_{d'>d}\BFvarepsilon_{d'}^{(t-1)}\right)\nonumber\\
    =& \BFK_d[\BFK_d+\lambda^{-1}\rmI_n]^{-1}\left(\sum_{d'<d}\BFvarepsilon_{d'}^{(t)}+\sum_{d'>d}\BFvarepsilon_{d'}^{(t-1)}\right) .\label{eq:pf_lemma_4_2_1}
\end{align}
From \eqref{eq:pf_lemma_4_2_1}, we can notice that $-\BFvarepsilon_d^{(t)}$ can be viewed as the values of a kernel ridge regression on $\BFX_d$: $h_d^{(t)}(\BFX_d)=-\BFvarepsilon^{(t)}_d$ such that
\begin{equation}
    h_d^{(t)}=\min_{h\in\CalH_{k_d}}\|h(\BFX_d)-\sum_{d'<d}\BFvarepsilon_{d'}^{(t)}-\sum_{d'>d}\BFvarepsilon_{d'}^{(t-1)}\|_2^2+\frac{1}{\lambda}\|h\|_{\CalH_{k_d}}^2 .\label{eq:pf_lemma_4_2_2}
\end{equation}
Moreover, from the representation theorem, we must have the following equality regarding the RKHS norm of $h^{(t)}_d$:
\[\|h_d^{(t)}\|_{\CalH_{k_d}}^2=[\BFvarepsilon_d^{(t)}]^\transpose\BFK_d^{-1}\BFvarepsilon_d^{(t)}.\]
Obviously $h=0$ is a feasible solution of \eqref{eq:pf_lemma_4_2_2}. Because $h^{(t)}_d$ is the minimum of \eqref{eq:pf_lemma_4_2_2}, we must have
\begin{align}
    \frac{1}{\lambda}\|h_d^{(t)}\|_{\CalH_{k_d}}^2\leq & \|0-\sum_{d'<d}\BFvarepsilon_{d'}^{(t)}-\sum_{d'>d}\BFvarepsilon_{d'}^{(t-1)}\|^2_2\nonumber\\
    \leq & D \left(\sum_{d'< d}\|\BFvarepsilon_{d'}^{(t)}\|_2^2+\sum_{d'> d}\|\BFvarepsilon_{d'}^{(t-1)}\|_2^2\right)\nonumber\\
    \leq & D\sum_{d'=1}^D\|\BFvarepsilon_{d'}^{(t-1)}\|_2^2=D\|\BFvarepsilon^{(t-1)}\|_2^2\label{eq:pf_lemma_4_2_3}
\end{align}
where the second line is from Jensen's inequality and the convexity of $l_2$ norm and the last line is because Back-fitting converges so we must have $\|\BFvarepsilon_{d'}^{(t)}\|_2\leq \|\BFvarepsilon_{d'}^{(t-1)}\|_2$ for any $d'$ and $t$. As a result, the RKHS norm for the error of the $t$-th iteration is
\begin{align*}
    [\CalS_n\BFvarepsilon^{(t-1)}]^\transpose\BFK^{-1}\CalS_n\BFvarepsilon^{(t-1)}=&[\BFvarepsilon^{(t)}]^\transpose\BFK^{-1}\BFvarepsilon^{(t)}\\
    =&\sum_{d=1}^D [\BFvarepsilon_d^{(t)}]^\transpose\BFK_d^{-1}\BFvarepsilon_d^{(t)}\\
    \leq & \lambda D\sum_{d=1}^D\|\BFvarepsilon^{(t-1)}\|_2^2=\lambda D^2\|\BFvarepsilon^{(t-1)}\|_2^2
\end{align*}
where the last line is from \eqref{eq:pf_lemma_4_2_3}. 

According to definition, $\|\CalS_n\BFvarepsilon^{(t-1)}\|_{\CalH_n}^2=[\CalS_n\BFvarepsilon^{(t-1)}]^\transpose\BFK^{-1}\CalS_n\BFvarepsilon^{(t-1)}$, we can finish the proof.
\endproof

\subsection{Proof of Theorem \ref{thm:KMG_convergence}}

\proof[Theorem \ref{thm:KMG_convergence}]
From Theorem \ref{thm:sparse_GPR} and Lemma \ref{lem:smoothing_property}, we can conclude that the error form has the following induction property for any $\BFvarepsilon_n^{(t)}$:
\begin{align*}
    \|\BFvarepsilon^{(t)}\|_2=&\|\CalT_n^m\CalS_n\BFvarepsilon^{(t-1)}\|_2\\
    \leq &  C^*\sqrt{n}\left|[\frac{h_m^{s}}{\sqrt{\lambda \kappa^*_n\kappa_m^*}}+\sqrt{\lambda n}h_m^{2s}]\|\CalS_n\BFvarepsilon^{(t-1)}\|_{\CalH_n}\right|\\
    \leq & C^*\sqrt{\lambda n}D[\frac{h_m^{s}}{\sqrt{\lambda \kappa^*_n\kappa_m^*}}+\sqrt{\lambda n}h_m^{2s}]\|\BFvarepsilon^{(t-1)}\|_2\\
    \leq &(1-\delta) \|\BFvarepsilon^{(t-1)}\|_2
\end{align*}
where the last line is from the condition \eqref{eq:KMG_condition} imposed on $h_m$. Then by induction, we can immediately have the result:
\[\|\BFvarepsilon^{(t)}\|_2\leq (1-\delta) \|\BFvarepsilon^{(t-1)}\|_2\leq (1-\delta)^2 \|\BFvarepsilon^{(t-2)}\|_2\leq\cdots\leq (1-\delta)^t \|\BFvarepsilon^{(0)}\|_2. 
\]\endproof





\end{document}